\documentclass[runningheads]{llncs}
\usepackage{graphicx}

\usepackage{tikz}
\usepackage{comment}
\usepackage{amsmath,amssymb} 
\usepackage{color}
\usepackage{subcaption}
\usepackage{booktabs} 
\usepackage{csquotes}
\usepackage{hyperref}
\newcommand{\vect}[1]{\boldsymbol{#1}}

\usepackage[accsupp]{axessibility}  

\usepackage{array,multirow}

\begin{document}
\pagestyle{headings}
\mainmatter
\def\ECCVSubNumber{2589}  

\title{ Learning Continuous Implicit Representation for Near-Periodic Patterns}
\newcommand{\etal}{\emph{et al.}}
\newcommand{\etals}{\emph{et al.} }

\newcommand{\eg}{\emph{e.g}., }
\newcommand{\ie}{\emph{i.e}., }

\titlerunning{NPP-Net}
%
\author{Bowei Chen\inst{1}\index{Chen, Bowei} \and
Tiancheng Zhi\inst{1} \and
Martial Hebert\inst{1} \and 
Srinivasa G. Narasimhan\inst{1}
}
\authorrunning{B. Chen et al.}
%
\institute{ Carnegie Mellon University, Pittsburgh PA 15213, USA 
\email{\{boweiche,tzhi,mhebert,srinivas\}@andrew.cmu.edu}\\
}
\maketitle

\begin{abstract}
Near-Periodic Patterns (NPP) are ubiquitous in man-made scenes and are composed of tiled motifs with appearance differences caused by lighting, defects, or design elements. A good NPP representation is useful for many applications including image completion, segmentation, and geometric remapping. But representing NPP is challenging because it needs to maintain global consistency (tiled motifs layout) while preserving local variations (appearance differences). Methods trained on general scenes using a large dataset or single-image optimization struggle to satisfy these constraints, while methods that explicitly model periodicity are not robust to periodicity detection errors.
To address these challenges, we learn a neural implicit representation using a coordinate-based MLP with single image optimization. We design an input feature warping module and a periodicity-guided patch loss to handle both global consistency and local variations. To further improve the robustness, we introduce a periodicity proposal module to search and use multiple candidate periodicities in our pipeline. We demonstrate the effectiveness of our method on more than 500 images of building facades, friezes, wallpapers, ground, and Mondrian patterns in single and multi-planar scenes.
Code and demo are available in our project website: \url{https://armastuschen.github.io/projects/NPP_Net/}
\end{abstract}

\section{Introduction}
\label{sec:intro}

\begin{figure}[!tbh]
\centering
    \captionsetup[subfigure]{labelformat=empty}
        \begin{tabular}{cccc}
 \begin{subfigure}[c]{0.24\linewidth}
  {\includegraphics[width=2.96cm]{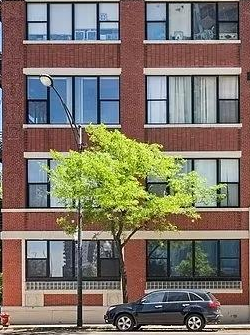}}
  \caption{(a) A NPP Scene}
\end{subfigure}
 \begin{subfigure}[c]{0.24\linewidth}
  {\includegraphics[width=2.96cm]{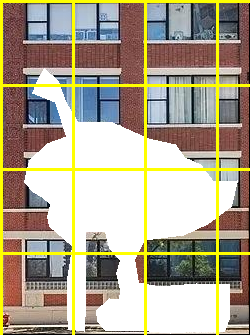}}
  \caption{(b) Input}
\end{subfigure}
 \begin{subfigure}[c]{0.24\linewidth}
  {\includegraphics[width=2.96cm]{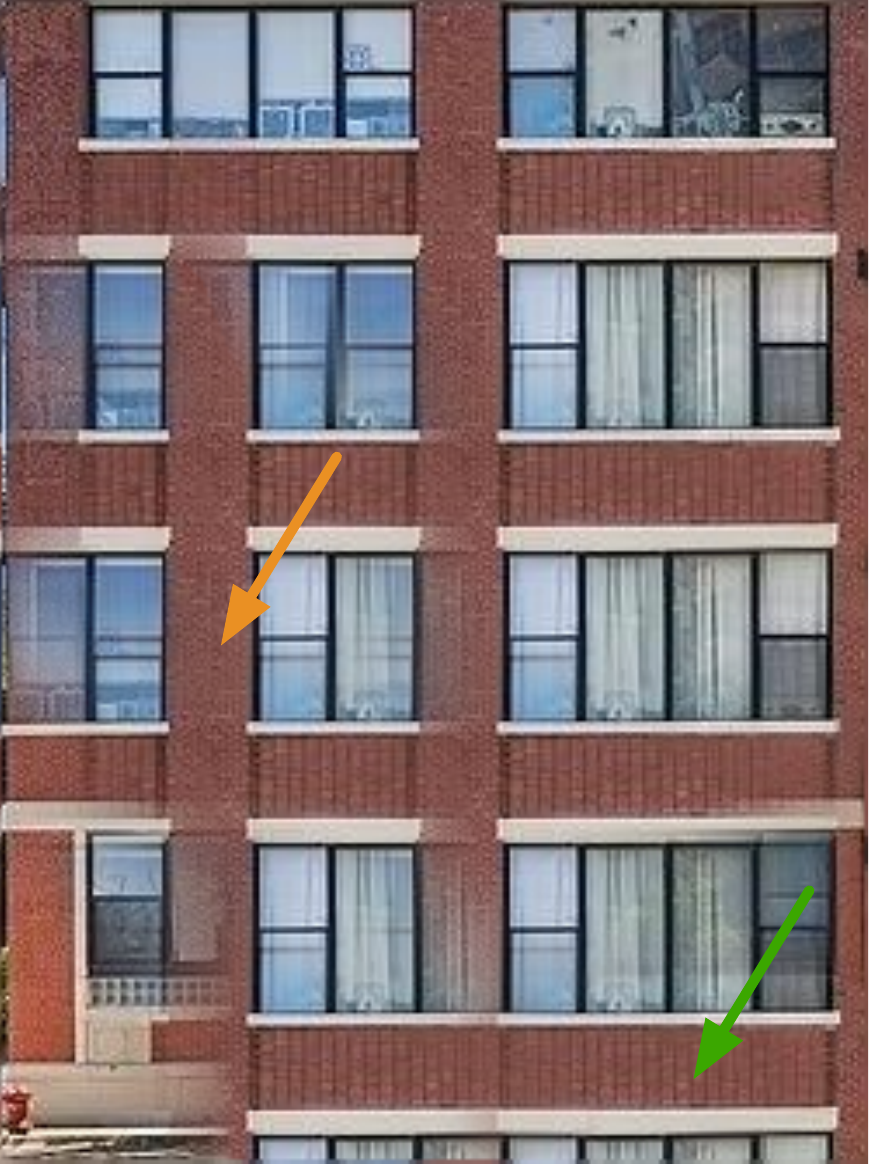}}
  \caption{(c) BPI~\cite{li2020multi}}
\end{subfigure}
 \begin{subfigure}[c]{0.24\linewidth}
  {\includegraphics[width=2.96cm]{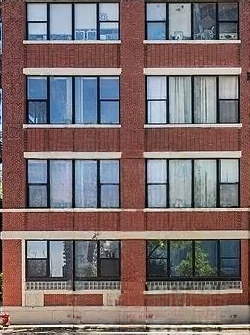}}
  \caption{(d) Ours}
\end{subfigure} 
\end{tabular}
\caption{Inpainting to remove the tree, street lamp, and car from a near-periodic patterned (NPP) scene in (a). Input image (b) visualizes the mask (white unknown region) and detected (but inaccurate) NPP representation (yellow lattice). Guided by this inaccurate representation, the state of the art method BPI~\cite{li2020multi} (c) fails to generate windows occluded by the tree (orange arrow) and the white strip across the bottom (green arrow). 
Our NPP-Net (d) maintains global consistency and local variations, while preserving the known regions. 
}
\label{fig:intro}
\end{figure}

Patterns are all around us and help us understand our visual world. In the 1990s, a human preattentive vision experiment~\cite{RAO1993218} showed that periodicity is a crucial factor in high-level pattern perception. But most real patterns are not composed of perfectly periodic (tiled) motifs. Consider the commonly occurring  real-world building facade scene in Figure \ref{fig:intro} (a). While the windows are laid out periodically, they vary in their individual appearances. There are several design elements (borders, texture), shading variations, or obstructions (tree, car, street lamp) that are not periodic. These factors make it challenging to create a good computational representation for such ``Near-Periodic Patterns'' (NPP). 

A good NPP representation must preserve both \textit{global consistency} (similar motifs layout) and \textit{local variations} (different appearances). For global consistency, the distance (periods) and orientations between adjacent motifs should be accurate (\eg window layout). At the same time, the local details in the scene should be fully encoded (\eg appearance variations in windows or the horizontal design strips). 
In this paper, we present a novel method to learn such a representation that can be used for  applications such as image completion (our  main focus), segmentation of periodic parts, and resolution enhanced scene remapping, \eg transforming to a fronto-parallel view (see supplementary). 

Existing image completion works applicable for NPP can be classified into two categories. 
The first category does not explicitly consider knowledge of periodicity. They complete images by training on large datasets~\cite{Nazeri_2019_ICCV,zeng2020high,Li_2020_CVPR,suvorov2022resolution,yan2019PENnet} or by exploring single image  statistics~\cite{sitzmann2020implicit,ulyanov2018deep,barnes2009patchmatch}. 
However, these methods fail to generate good global consistency, especially with a large unknown mask inside (interpolation) or outside (extrapolation) the image border, or severe perspective effect.
The other category~\cite{huang2014image,li2020perspective,li2020multi,mao2019program} models periodicity as prior for image completion.
These works extract explicit NPP representations (\eg displacement vectors) and use them to guide image completion. 
These methods can generate good global periodic structure \emph{if} the estimated periodicity is accurate. However, this is hard to achieve in the presence of strong local variations.

Our work is inspired by the recent progress on implicit neural representations~\cite{mildenhall2020nerf} that map image coordinates to RGB values using coordinate-based multi-layer perceptrons (MLP).
But, naively using this method fails on our task due to the lack of a good periodicity prior. Thus, we present a periodicity-aware coordinate-based MLP to learn a continuous implicit neural representation, which we call NPP-Net for short.
The key idea is to extract periodicity information from a partially observed NPP scene and inject it into both the MLP input and the loss function to help optimize NPP representation. 

Three novel steps are proposed for the above idea: (1) The \emph{Periodicity Proposal} step extracts periodicity in the form of a set of candidate periods and orientations that are used together to handle inaccurate detections; (2) The \emph{Periodicity-Aware Input Warping}  injects periodicity into the MLP input by warping input coordinates according to the proposed periodicities. This step preserves global consistency and the MLP converges to a good periodic pattern easily; (3) Finally, the \emph{Periodicity-Guided Patch Loss} samples observable patches according to periodicity to optimize the representation. This step preserves local variations, improves extrapolation ability, and removes high-frequency artifacts. 

Our approach only requires a single image for optimization. This is important since there are no large dedicated NPP datasets.
Thus, we evaluate our approach on a total of 532 NPP subclasses chosen from three  datasets~\cite{NRTDatabase,cimpoi14describing,teboul2010segmentation}. The scenes include building facades, friezes, ground patterns, wallpapers, and Mondrian patterns that are tiled on one or more geometric planes and perspectively warped.
Our dataset is larger than those (157 at most) used in previous works~\cite{huang2014image,li2020multi,li2020perspective,mao2019program} that are designed for NPP.
We mainly apply NPP-Net for the image completion task, but extend it to resolution enhanced NPP remapping and NPP segmentation in the supplementary. 
We compare NPP-Net with four traditional \cite{image_quilting,barnes2009patchmatch,huang2014image,li2020multi} and five deep learning-based methods \cite{ulyanov2018deep,sitzmann2020implicit,zeng2020high,yan2019PENnet,suvorov2022resolution}, and eight variants of NPP-Net. 
Experiments show that NPP-Net can interpolate and extrapolate images, in-paint large and arbitrarily shaped regions, recover blurry regions when images are remapped, segment periodic and non-periodic regions, in planar and multi-planar scenes.
Figure \ref{fig:intro} shows the effectiveness of NPP-Net, inpainting a complex NPP scene, compared to the state of the art BPI \cite{li2020multi}. While our method is not designed for general scenes, it is a useful tool to understand a large class of man-made scenes with near-periodic patterns.

\section{Related Work}
\label{sec:formatting}

\noindent\textbf{Near-Periodic Patterns Completion:}
There are two types of image completion methods that can be applied to NPP. The first type of methods do not explicitly consider periodicity as prior for completion~\cite{Nazeri_2019_ICCV,zeng2020high,Li_2020_CVPR,suvorov2022resolution,yan2019PENnet,wang2021image,cao2021learning,zhou2021transfill}. The second stream of methods takes advantage of periodicity to guide the completion. 
We focus on reviewing the second type of methods.

The first stage for these methods is to obtain an NPP representation to guide image completion.
Existing methods aim to represent NPP by detecting the global periodicity despite local variations.
The types of NPP arrangements vary~\cite{pritts2014detection,pritts2017coplanar,liu2013grasp,huang2014image,wu2010detecting,park2010translation} but commonly, periodic patterns are assumed to form a 2D lattice~\cite{liu2004computational,Liu2015patchmatch,hays2006discovering,lettry2017repeated,Park2008deformed,park2009deformed,li2020perspective}. The first lattice-based work~\cite{hays2006discovering} for periodicity detection without human interaction finds correspondences using visual similarity and geometric consistency. Liu \etal~\cite{Liu2015patchmatch} improve this process by incorporating generalized PatchMatch~\cite{barnes2010generalizedpatchmatch} and Markov Random Field. Furthermore, Lettry \etal~\cite{lettry2017repeated} detect a repeated pattern model by searching in the feature space of a pre-trained CNN. Recently, Li \etal~\cite{li2020perspective} design a compact strategy by searching on deep feature space without any implicit models. But it requires hyperparameter tuning to achieve competitive results. All existing methods describe periodicity using an explicit representation such as keypoints~\cite{hays2006discovering,pritts2017coplanar,torii2013visual,Liu2015patchmatch}, feature-based motifs~\cite{pritts2014detection} or displacement vectors~\cite{lettry2017repeated,li2020perspective}. But they do not preserve both global consistency and local variations well.

The second stage is to generate or inpaint an NPP image guided by the NPP representation~\cite{liu2003deformable,liu2004near,liu2005promise,huang2014image,mao2019program,li2020perspective,li2020multi,Halperin2021animation}. 
One common assumption is that the NPP lie on a single plane. Liu \etal~\cite{liu2004near} synthesize an NPP image through multi-model deformation fields given an input NPP patch and its representation. Mao \etal~\cite{mao2019program} propose GAN-based NPP generation. 
Huang \etal~\cite{huang2014image} and BPI~\cite{li2020multi} extend image completion to the multi-plane case. They detect periodicities in these planes~\cite{he2012statistics,lowe1999objectsift,li2020perspective} and use them to guide image completion, based on \cite{wexler2007space} and \cite{barnes2009patchmatch}.
Unlike earlier work Huang \etal, BPI uses the periodicity detection method based on feature maps extracted from a pre-trained network.
Also, the state of the art BPI's image completion step does not use their prior GAN-based method~\cite{mao2019program}.

In summary, the above methods assume that NPP representation is good enough for guidance, which is not guaranteed.
By contrast, we merge the two stages by optimizing the implicit representation using image reconstruction error.

\noindent\textbf{Implicit Neural Representations:}
Recently, coordinate-based multi-layer perceptron (MLP) has been used to obtain implicit neural representation (INR). It maps coordinates to various signals such as shapes~\cite{genova2019learning,chen2019learning,sitzmann2020implicit}, scenes~\cite{mildenhall2020nerf,martin2021nerf} and images~\cite{chen2021learning,bemana2020x,sitzmann2020implicit}. Mildenhall \etal~\cite{mildenhall2020nerf} represent a 3D scene from a sparse set of views for novel view synthesis. 
Siren~\cite{sitzmann2020implicit} 
replaces ReLU by a periodic activation function and designs an initialization scheme for modeling finer details. 
Chen \etal~\cite{chen2021learning} present a Local Implicit Image Function for the generation of arbitrary resolutions.  Skorokhodov \etal~\cite{skorokhodov2021adversarial} design a decoder based on INR with GAN training, for high-quality image generation. 

NPP-Net differs from previous methods in two ways:
(1) Directly using MLP~\cite{mildenhall2020nerf,sitzmann2020implicit} fails to learn accurate NPP representation without high-level structural understanding. We propose a periodicity-aware MLP.
(2) Many works require a large dataset for training, while we optimize on a single image.

\section{NPP-Net}

\begin{figure*}[!t]
    \centering
    \begin{tabular}{cc}
        \includegraphics[width=0.9\textwidth]{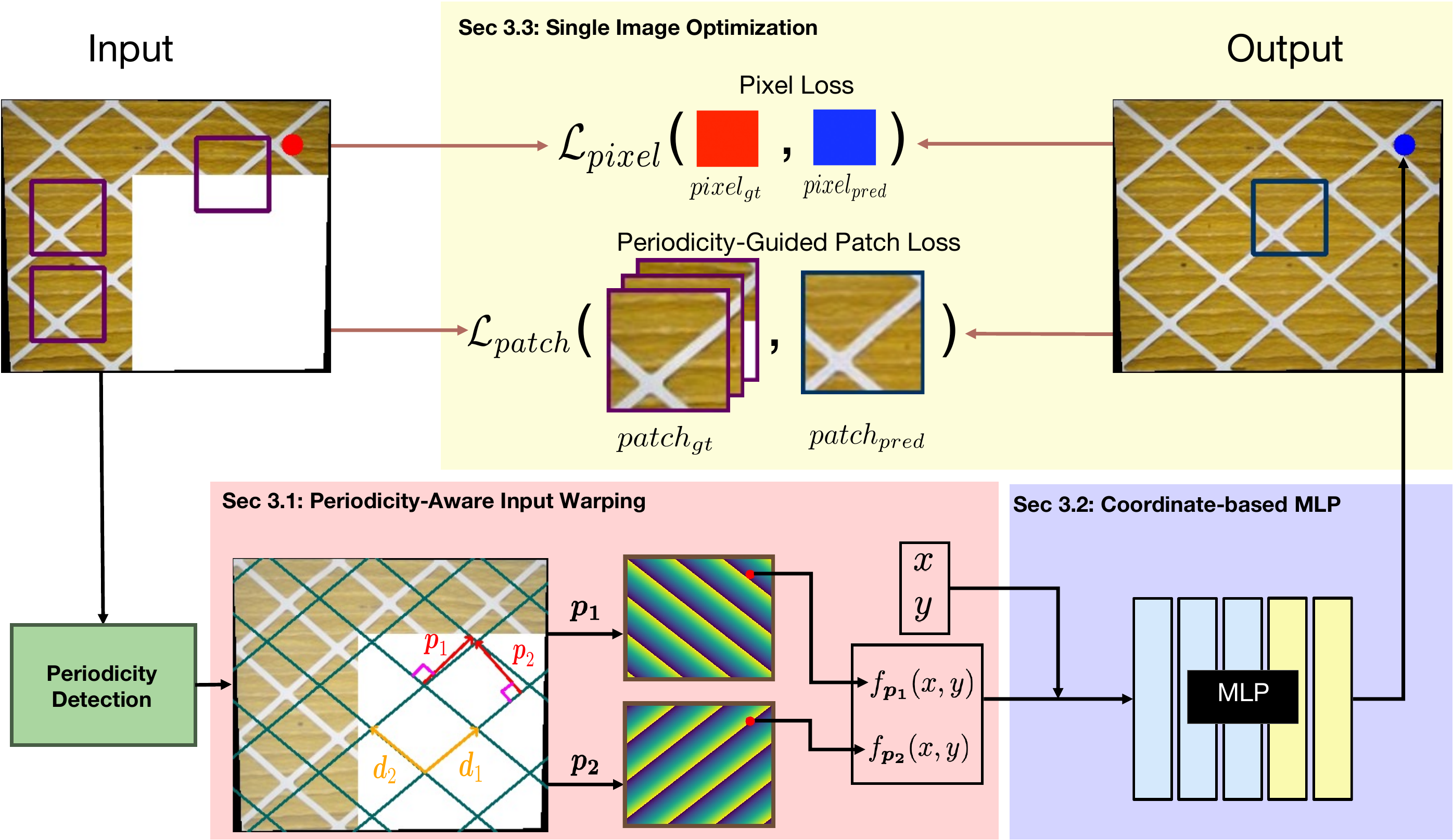}\\
    \end{tabular}
    \caption{ Initial pipeline of NPP-Net consists of three modules. 
    (1)  \textit{Periodicity-Aware Input Warping} (pink) warps input coordinates using detected periodicity. 
    (2) \textit{Coordinate-based MLP} (blue) maps warped and input coordinate features to an RGB value. 
    (3) \textit{ Single Image Optimization} (yellow) uses pixel loss and periodicity-guided patch loss on a single NPP image. 
    Final pipeline in section \ref{sec:pp} shows  how multiple periodicities are automatically detected and utilized. 
     }
    \label{NPP-Net_overview}
 \end{figure*}

We aim to build an MLP that maps image coordinates to pixel values, given a partial observation of an NPP image. We will describe NPP-Net using the image completion task.  The unknown (masked) region is completed (or inpainted) by training on the remainder of the NPP image.
For clarity, we first describe the method for single planar NPP scene and pre-warp the image to be fronto-parallel \cite{zhang2012tilt}. Then we will extend NPP-Net to handle multi-planar scenes.

Our key idea is to extract periodicity information from the known NPP region and inject it into the MLP input and the loss function.
The initial pipeline of NPP-Net (Figure~\ref{NPP-Net_overview}) consists of three modules: (1) Periodicity-Aware Input Warping transforms image coordinates using the detected periodicity. 
(2) Coordinate-based MLP maps the transformed coordinates to the corresponding RGB value. 
(3) Single Image Optimization provides a periodicity-guided loss function for optimizing the MLP on a single image.

\subsection{Periodicity-Aware Input Warping}
\label{sec: PAIW}

A traditional MLP is not good at capturing global periodic structure without additional priors. 
In fact, previous works~\cite{xu2021how,NEURIPS2020_11604531} have shown that a traditional MLP is unable to extrapolate a 1D periodic signal even with many training samples. 
The Periodicity-Aware Input Warping module thus explicitly injects periodicity information into the MLP by warping image coordinates $(x,y)$.

Assuming a 2D lattice arrangement, the periodicity is represented as two displacement vectors $\vect{d}_1$ and $\vect{d}_2$ (orange arrows in Figure \ref{NPP-Net_overview}). A perfect infinite periodic pattern is invariant if shifted by $\alpha \vect{d}_1 + \beta\vect{d}_2 (\alpha, \beta \in \mathbb{Z})$. This representation can be transformed into periods and orientations, visualized as the magnitudes and orientations of the red arrows ($\vect{p}_1$ and $\vect{p}_2$, called \textbf{periodicity vectors}). 
Mathematically, the transformation is obtained by solving $\vect{p}_1 \cdot \vect{d}_2 = \vect{d}_1 \cdot \vect{p}_2 = \vect{0}$ and $\vect{p}_1 \times \vect{d}_2 = \vect{d}_1 \times \vect{p}_2 = \vect{d}_1 \times \vect{d}_2$, where the cross product is defined using the corresponding 3D vectors on the plane $z=0$. A \textbf{periodicity} is then defined as a vector pair $(\vect{p}_1, \vect{p}_2)$. Extension to circular patterns is available in supplementary.

One way to obtain the periodicity for an NPP image is to treat it as a learnable parameter and jointly optimize it with NPP-Net~\cite{chen2022exemplar,jetchev2016texture}. However, this is hard for two reasons. 
(1) Good periodicity is not unique (any multiple works). 
(2) Many real-world NPP scenes contain strong local variations, leading to a complicated cost function. Thus we adopt an existing periodicity detection method~\cite{li2020perspective} for input warping, which extracts feature maps from a pretrained CNN and performs brute force search to obtain a periodicity. 
Then, for each periodicity vector $\vect{p} = (p \cos \theta, p \sin \theta) \in \{\vect{p}_1, \vect{p}_2\}$,
we define a warp as a bivariate function:
\begin{equation}
    f_{\vect{p}}(x,y) = (x \cos\theta + y\sin\theta) \bmod p.
\end{equation}
This function generates a warped coordinate value sampled from a periodic pattern with period $|\vect{p}|$ along direction $\frac{\vect{p}}{|\vect{p}|}$, as shown in Figure \ref{NPP-Net_overview}. 
Through this feature engineering, the warped coordinates explicitly encode the periodicity information.
The warped coordinates $f_{\vect{p}_1}(x,y)$ and $f_{\vect{p}_2}(x,y)$, together with the original coordinates $x$ and $y$, are further normalized to $[-1,1]$ and passed through positional encoding~\cite{mildenhall2020nerf} to allow the network to model high frequency signals~\cite{tancik2020fourier}. The encoded coordinates are then input to the MLP. 
We keep the notations of coordinates before and after positional encoding the same for simplicity.
The dimension of the features is $4d$, including $d$  frequencies in the positional encoding, and a set of four values for each frequency: $x, y, f_{\vect{p}_1}(x,y),$ and $f_{\vect{p}_2}(x,y)$.

\subsection{Coordinate-Based MLP}
\label{ssec:mlp}
We adopt coordinate-based MLP to represent NPP images. It is more effective and compact than a CNN to model periodic signals since coordinates are naturally suited for encoding positional (periodic) information.
Specifically, we input the warped coordinate features to enforce global consistency, and also input the original coordinate features without warping to help preserve local variations. The output of the MLP is an RGB value corresponding to the input image coordinate.
Since ReLU activation function has been proven to be ineffective to extrapolate periodic signals~\cite{xu2021how}, we use the more suitable SNAKE function~\cite{NEURIPS2020_11604531}.

\subsection{Single Image Optimization}\label{single_image_optim}

\subsubsection{Pixel Loss:}
Pixel loss is the most intuitive way to optimize coordinate-based MLP~\cite{mildenhall2020nerf}, which compares predicted and ground truth pixel values. 
For image coordinate $\vect{x}=(x, y)$, we adopt the robust loss function $\mathcal{L}_{rob}$~\cite{barron2019general}, given by:
\begin{equation}
    \mathcal{L}_{pixel}(\vect{x}) = \mathcal{L}_{rob} (\hat{C} 
    (\vect{x}), {C}
    (\vect{x})),
\end{equation}
where $\hat{C} (\vect{x})$ and ${C} (\vect{x})$ are the output RGB values of the MLP and the ground truth RGB values from the input image at position $\vect{x}$, respectively. This loss is applied only to the known regions.

But merely adopting pixel loss like NeRF~\cite{mildenhall2020nerf} fails to generate a good NPP for two reasons:
(1) The high-dimensional input features result in the generation of high-frequency artifacts (Figure \ref{method:loss_abaltion} (b)). See \cite{tancik2020fourier,zheng2021rethinking} for details about this problem. 
(2) Pixel loss does not enforce explicit constraints to model the correlation between a coordinate's features and its neighbors. This constraint is critical for preserving local variations since it helps capture local patch statistics.
Thus, pixel loss fails to preserve local variations. For example, in Figure \ref{method:loss_abaltion} (b), pixel loss
generates some periodic artifacts in the top non-periodic region{\footnote{ \footnotesize Figure \ref{method:loss_abaltion} is generated based on the final pipeline explained in Section \ref{sec:pp}.}.
\normalsize

\subsubsection{Periodicity-Guided Patch Loss:}
\label{Patch_loss}
To address the limitations of pixel loss, we force the network to learn patch internal statistics by incorporating patch loss, which compares predicted and ground truth patches. The ground truth patches can be sampled at the same position as the predicted patch (for known regions), or sampled according to periodicity (for any region).

\noindent\textbf{GT Patch at the Same Position:} For a predicted patch in the known region, the ground truth patch at the same position is available. Specifically, for a square patch with size $s$ centered at position $\vect{x}$, we input all the pixel coordinates in the patch into MLP to obtain a predicted RGB patch $\hat{I}_s(\vect{x})$. Let $I_s( \vect{x})$ be the corresponding ground truth at the same position and $M_s(\vect{x})$ be the mask of known pixels. We apply perceptual loss~\cite{zhang2018perceptual} on masked patches:
\begin{equation}    
\mathcal{L}_{p}( \vect{x}) =\mathcal{L}_{pct}(
    \hat{I}_s( \vect{x}) \odot M_s( \vect{x}),  I_s( \vect{x}) \odot M_s( \vect{x})),
\end{equation}
where $\odot$ is the element-wise product. 

\noindent\textbf{GT Patches Sampled Based on Periodicity:}
To train on unknown regions, we propose to sample ground truth patches based on periodicity. This is an effective way to handle the MLP extrapolation problem, which cannot be solved by merely using input warping and SNAKE activation function~\cite{NEURIPS2020_11604531}. The input and output images in Figure \ref{NPP-Net_overview} illustrate this sampling strategy.

Specifically, we sample multiple nearby ground truth patches for supervision by shifting position $\vect{x}$ based on the  estimated periodicity. 
The shifted patch center is defined as $\vect{x}_{\alpha\beta} = \vect{x} + \alpha \vect{d}_1 + \beta \vect{d}_2 (\alpha, \beta \in \mathbb{Z})$, where $\vect{d}_1$ and $\vect{d}_2$  are the displacement vectors.
Because the predicted and ground truth patches are not necessarily aligned, we adopt contextual loss $\mathcal{L}_{ctx}$~\cite{mechrez2018contextual}:
\begin{equation}
\begin{split}        
\mathcal{L}_{c}( \vect{x}) = 
    \frac{1}{N}\sum_{(\alpha, \beta) \in T_N}\mathcal{L}_{ctx}(&
    \hat{I}_s( \vect{x}) \odot M_s( \vect{x}_{\alpha\beta}),  I_s( \vect{x}_{\alpha\beta}) \odot M_s( \vect{x}_{\alpha\beta})),
\end{split}
\end{equation}
where $T_N$ is a set of $(\alpha, \beta)$ pairs corresponding to the $N$ nearest ground truth patches, since local variations are preserved using nearby patches for supervision.

\noindent\textbf{Patch Loss:} 
Our patch loss combines the two sampling strategies: $\mathcal{L}_{patch}( \vect{x}) = \lambda_p \gamma(\vect{x}) \mathcal{L}_{p}( \vect{x}) + \lambda_c \mathcal{L}_{c}( \vect{x})$, where $\lambda_p$ and $ \lambda_c$ are constant weights. $\gamma(\vect{x})$ is a binary function: 1 for $\vect{x}$ in known regions and 0 for $\vect{x}$ in unknown regions.

\begin{figure}[!t]
    \captionsetup[subfigure]{labelformat=empty}
\begin{subfigure}{.16\linewidth}
  \centerline{\includegraphics[width=\textwidth]{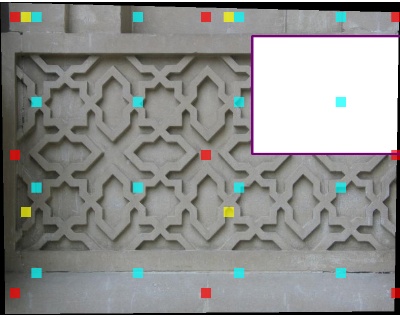}}
  \caption{\scriptsize (a) Input \normalsize}
\end{subfigure}
\hfill
\begin{subfigure}{.16\linewidth}
  \centerline{\includegraphics[width=\textwidth]{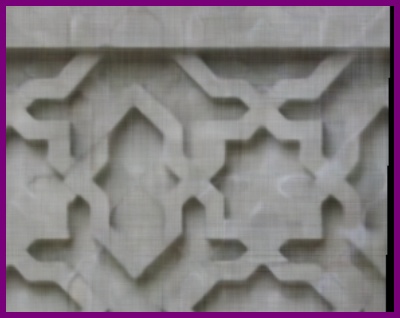}}
  \caption{\scriptsize (b) Pixel Only \normalsize}
\end{subfigure}
\hfill
\begin{subfigure}{.16\linewidth}
  \centerline{\includegraphics[width=\textwidth]{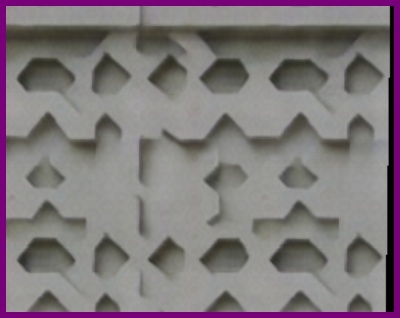}}
  \caption{\scriptsize (c) Patch Only \normalsize}
\end{subfigure}
\hfill
\begin{subfigure}{.16\linewidth}
  \centerline{\includegraphics[width=\textwidth]{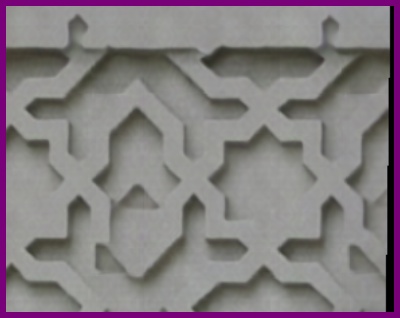}}
  \caption{\scriptsize (d) Pixel+Rand   \normalsize}
\end{subfigure}
\hfill
\begin{subfigure}{.16\linewidth}
  \centerline{\includegraphics[width=\textwidth]{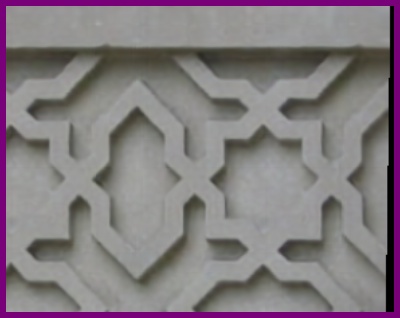}}
  \caption{\scriptsize (e) NPP-Net \normalsize}
\end{subfigure} 
\hfill
\begin{subfigure}{.16\linewidth}
  \centerline{\includegraphics[width=\textwidth]{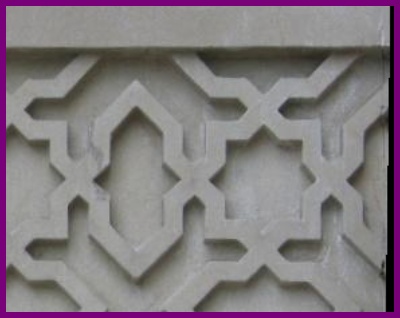}}
  \caption{\scriptsize (f) GT \normalsize}
\end{subfigure}
\caption{
Comparing different losses based on the final pipeline. The red, yellow and cyan dots in (a) visualize the Top-3 periodicities. Zoom-ins of the unknown area (white rectangle) are in (b)-(f). Merely using pixel loss (b) generates high-frequency artifacts across the image and periodic artifacts in the top part. 
Adopting only patch loss (c) removes the artifacts but has poor global structure. Using pixel loss and patch loss with random sampling (d) cannot preserve global consistency and local variations well since the ground truth patches are not sampled according to periodicity and might be far from the predicted patch. With pixel loss and periodicity-guided patch loss, NPP-Net (e) solves these issues. 
}
\label{method:loss_abaltion}
\end{figure}

\noindent\textbf{Total Loss: }
Our final loss is the combination of patch loss and pixel loss:
\small
\begin{equation}
    \mathcal{L} = \frac{\lambda_1}{|B_1|} \sum_{ \vect{x} \in B_1} \mathcal{L}_{pixel}(\vect{x}) + \frac{\lambda_2}{|B_2|}  \sum_{\vect{x} \in B_2} \mathcal{L}_{patch}(\vect{x}), 
\end{equation}
\normalsize
where $\lambda_1$ and $\lambda_2$ are constant weights. $B_1$ contains pixel coordinates that are randomly sampled in known areas, and $B_2$ contains the center coordinates sampled in both known and unknown areas in proportion.

Training with this loss preserves both global consistency and local variations, as shown in Figure \ref{method:loss_abaltion} (e). In fact, only using patch loss cannot ensure global consistency if the detected periodicity is not accurate enough. In Figure \ref{method:loss_abaltion} (c), the pattern structure is poorly reconstructed because it only focuses on the local structure. We also show the result for the combination of pixel loss and patch loss with patches that are randomly sampled in the known regions (we call it random sampling strategy) in Figure \ref{method:loss_abaltion} (d). This fails to generate correct periodic patterns and good local details because the output and sampled patches have a large misalignment and are far away from each other.

\subsection{Periodicity Proposal}
\label{sec:pp}

Although the above initial pipeline shows good performance, it still fails to handle very inaccurate periodicity detection. To improve the robustness of NPP-Net, we design a Periodicity Proposal module to provide additional periodicity information. As shown in Figure \ref{NPP-Net_Aug}, we first search multiple candidate periodicities and then augment the input to MLP to handle inaccurate periodicity detection.

\noindent\textbf{Periodicity Searching:} 
Our searching strategy is based on the same periodicity detection method~\cite{li2020perspective} we adopt in the initial pipeline.
But the authors' original implementation requires manual hyperparameter tuning. Instead, we design an automatic tuning method, which evaluates each candidate periodicity (obtained from various hyperparameters) in the context of image completion. 
Specifically, we first generate $M$ pseudo masks in the known regions and treat them as unknown masks for image completion.
Then we execute the initial pipeline for each candidate periodicity, and compute its reconstruction error in pseudo mask regions for periodicity ranking. 
Since we focus on reconstructing a coarse global structure, we use a lightweight initial NPP-Net without patch loss for efficiency, which takes around 10 seconds for each periodicity in a Titan Xp GPU.

\begin{figure*}[t]
    \centering
    \begin{tabular}{cc}
\includegraphics[width=12cm]{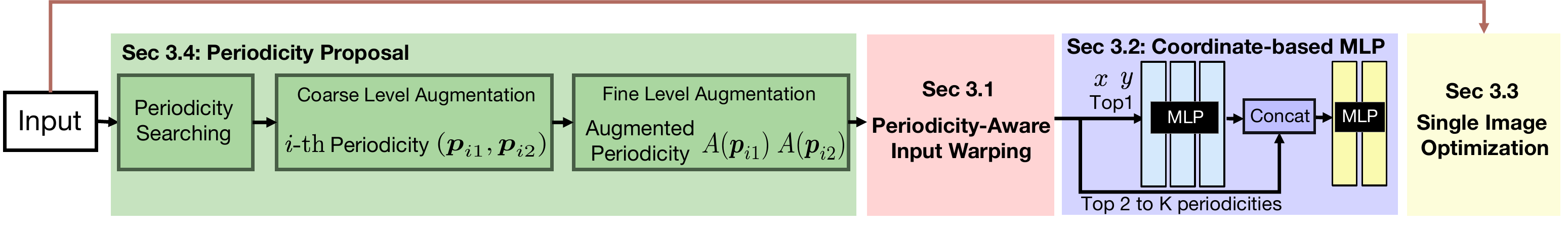}
    \end{tabular}
    \caption{Final pipeline of NPP-Net modifies two modules of the initial pipeline. (1) \textit{Periodicity Proposal} (green) automatically searches and augments the input periodicity to handle inaccurate periodicity detection and encourage the global consistency.
    (2) \textit{Coordinate-based MLP} (lavender blue) has two branches: (a) for Top-1 periodicity and original coordinates, and (b) for the rest. 
    }
    \label{NPP-Net_Aug}
 \end{figure*}

\noindent\textbf{Periodicity Augmentation:} 
Prior methods~\cite{huang2014image,mao2019program,li2020perspective,li2020multi} also use one periodicity to guide completion as in our initial pipeline. This cannot guarantee global consistency if the periodicity is  inaccurate, especially when the unknown mask is large (see experiments). So, we augment the pattern periodicity at two levels to improve robustness. At the coarse level, instead of searching the best periodicity, we keep Top-K candidates $\{(\vect{p}_{i1}, \vect{p}_{i2}) \mid i \in \mathbb{Z}^+, i \leq K \}$, to cover multiple possible solutions. 
This coarse-level augmentation encourages NPP-Net to move towards the most reasonable candidate periodicity. 
At the fine level, we augment periodicities with small offsets to better handle smaller errors.
Specifically, a periodicity vector $\vect{p}$ is augmented to be 
$A(\vect{p}) = \{ \vect{p} + \delta\frac{\vect{p}}{|\vect{p}|}    \mid \delta\in\Delta\}$. 
We empirically define $\Delta=\{0, \pm0.5, \pm1\}$ (in pixels). Finally, we merge all the augmented periodicity vectors as $P = \bigcup_{i \in \mathbb{Z}^+, i \leq K } A(\vect{p}_{i1}) \cup A(\vect{p}_{i2})$.
Note that $|P| = 2K|\Delta|$. 

In our final pipeline, $P$ contains $K|\Delta|$ periodicities. We perform input warping (Section \ref{sec: PAIW}) for each periodicity and input the transformed coordinate features into the MLP. We add an additional branch to the MLP, as shown in Figure \ref{NPP-Net_Aug}. Since the Top-1 periodicity is likely the most accurate (see experiments), we input the coordinate features warped using the Top-1 periodicity (including fine-level augmentation) and original coordinate features, to the first branch. The coordinate features warped using the Top-2 to K periodicities are sent to the second branch. For optimization, we sample patches according to the Top-1 periodicity. All other parts remain the same in our final pipeline. We evaluate these changes in our ablation study. 
See supplementary for implementation details including hyperparameters, network architecture, and runtime.
\subsection{Extensions}
\textbf{Non-NPP region segmentation}: Parts of a scene may not be near-periodic (\eg trees in front of a building facade). We thus segment the non-periodic regions in an NPP image in an unsupervised manner. We use a traditional segmentation method~\cite{Segmentation2017} to provide an initial guess for the non-periodic regions, treated as the unknown mask in image completion. After training NPP-Net, we relabel the initial non-periodic regions with low reconstruction error as periodic regions. 
\textit{Similar strategy is adopted to serve as a pre-filtering step before applying our method to any arbitrary scene (see supplementary).} 

\noindent \textbf{NPP remapping}: 
NPP scenes captured from a tilted angle can result in blurry motifs after rectification. 
To enhance resolution, 
we detect blurry regions and treat them as the unknown mask in image completion. The difference is that we compute the pixel loss in the blurry regions with a smaller weight.

\noindent \textbf{Multi-Plane NPP completion}: 
Given an image with different NPPs on different planes, we first adopt a pre-trained plane segmentation network~\cite{YuZLZG19} to obtain a coarse plane segmentation. Then we select a bounding box in each plane as a reference to rectify the plane using TILT~\cite{zhang2012tilt}. Note that, we do not require accurate segmentation since it is only used for bounding box selection.
For each rectified plane, we first use our NPP segmentation method to segment the non-periodic regions (mainly from other planes) and treat them as invalid pixels. Then we perform NPP completion on each plane, transform it back to the original image coordinate system, and recompose the image. Figure \ref{multi_plane_example} shows qualitative results for this extension.
\textit{Detailed implementation and experiments for these extensions are in supplementary.}

\section{Experiments}

\label{exp: completion}
\noindent\textbf{Dataset:}
We evaluate NPP-Net on 532 images selected from three relevant datasets for NPP completion: PSU Near-Regular Texture Database (NRTDB)~\cite{NRTDatabase}, Describable Textures Dataset (DTD)~\cite{cimpoi14describing}, and Facade Dataset~\cite{teboul2010segmentation}. 
There are 165 NPP images in the NRTDB dataset including facades, friezes, bricks, fences, grounds, Mondrian images, wallpapers, and carpets.
Similarly, there are 258 NPP images in the DTD Dataset including honeycombs, grids, meshes and dots.
The Facade Dataset has 109 rectified images of facades. Some of these facades are strictly \textit{not} NPP because often the windows are not arranged periodically. But nonetheless we include these to evaluate our approach when the NPP assumption is not strictly satisfied. Finally, we also collect a small dataset with 11 NPP images for real-world applications (\eg removing trees in the scene). 
In general, scenes in NRTDB are more challenging than DTD since they contain more non-periodic regions (boundaries, trees, sky, etc.), complex illuminations and backgrounds, and multiple periodicities across an image (Figure \ref{comparison_baseline_completion} row 3).
We use TILT~\cite{zhang2012tilt} to rectify all the images to be fronto-parallel if needed. 
See supplementary for more details including sampled images and mask generation.

\noindent\textbf{Metrics:} No single metric can evaluate NPP image completion comprehensively. So we adopt three metrics to cover different scales, including LPIPS (perceptual distance)~\cite{zhang2018perceptual}, SSIM~\cite{wang2004image},  and PSNR. Lower LPIPS, higher SSIM, and higher PSNR mean better performance. A known limitation for SSIM and PSNR is that blurry images also tend to receive high scores in these metrics~\cite{ledig2017photo}, while LPIPS handles this issue better. See supplementary for FID~\cite{parmar2021cleanfid} and RMSE metrics.

\begin{table}[!tb]
	\begin{center}
		\scalebox{0.78}{
	 \begin{tabular}{ccccccccccc}	\toprule
	 \multirow{ 2}{*}{Category} &
	  \multirow{ 2}{*}{Method}  & \multicolumn{3}{c}{NRTDB~\cite{NRTDatabase}} &  \multicolumn{3}{c}{DTD \cite{cimpoi14describing}}  &  \multicolumn{3}{c}{Facade \cite{teboul2010segmentation}} \\
\cmidrule(lr){3-5} \cmidrule(lr){6-8} \cmidrule(lr){9-11} & 
	 	   &  LPIPS $\downarrow$  & SSIM $\uparrow$  & PSNR $\uparrow$ & LPIPS $\downarrow$  & SSIM $\uparrow$  & PSNR $\uparrow$ & LPIPS $\downarrow$  & SSIM $\uparrow$  & PSNR $\uparrow$\\
	 	\midrule 
  \multicolumn{1}{c|}{\multirow{3}{*}{\shortstack[l]{Large\\ Datasets}}}	 &	PEN-Net~\cite{yan2019PENnet}   & 0.497  & 0.452 & 17.97 & 0.473  & 0.365 & 15.81 & 0.426 & 0.444 & \underline{15.78}\\
\multicolumn{1}{c|}{}	 &	ProFill~\cite{zeng2020high}   & 0.401  & 0.300 & 16.35 & 0.443  & 0.249 & 14.30 &  0.374 & 0.391 & 14.73 \\
\multicolumn{1}{c|}{}	&    Lama~\cite{suvorov2022resolution}   & {\underline{0.196}}  & {\underline{0.551}} & {\underline{18.64}} &{\underline{0.274}}  & {\underline{0.479}} & \underline{16.39}  & \textbf{0.207} & \underline{0.468} & 15.24 \\  
	    	 	\midrule 
\multicolumn{1}{c|}{ \multirow{6}{*}{\shortstack[l]{Single\\ Image}} }	 &	 	Image Quilting~\cite{image_quilting} &0.428  & 0.074 & 13.25 & 0.415  & 0.077 & 12.18 & 0.550 & 0.002 & 10.28 \\ 
\multicolumn{1}{c|}{}	 &	PatchMatch~\cite{barnes2009patchmatch}  &   0.263  & 0.542 & 18.14 & 0.361 & 0.383 & 15.47 & 0.369 & 0.341 & 14.22 \\
\multicolumn{1}{c|}{}	& 	DIP~\cite{ulyanov2018deep}  &  0.554  & 0.292 & 16.46  &0.659  & 0.181 & 13.15  & 0.582 & 0.258 & 15.22\\
\multicolumn{1}{c|}{}&	 	Siren~\cite{sitzmann2020implicit} & 0.636  & 0.084 & 14.38 & 0.762  & 0.080 & 13.11 & 0.780 & 0.052 & 12.00 \\
\multicolumn{1}{c|}{}	& 	Huang \etal~\cite{huang2014image}   & 0.287  & 0.410 & 16.99  & 0.302  & 0.320 & 14.88 & 0.387  & 0.279 & 13.75 \\  
\multicolumn{1}{c|}{}	& 	BPI~\cite{li2020multi}    & 0.254  & 0.442 & 16.86  & 0.303  & 0.305 & 14.82 & 0.458 & 0.173 & 12.20 \\  
	 	\midrule
\multicolumn{1}{c|}{ \multirow{8}{*}{\shortstack[l]{NPP-Net\\ Variants}}}	 	 &	No Periodicity   &0.429  & 0.449 & 18.22  &  0.468  & 0.350 & 16.38  & 0.379 & 0.443 & 15.54 \\  
\multicolumn{1}{c|}{}&	 	Pixel Only   & 0.308  & 0.618 & 20.20  & 0.397  & 0.473 & 17.86 & 0.427 & 0.458 & 15.54 \\  
\multicolumn{1}{c|}{}&	 	Patch Only   & 0.322  & 0.340 & 15.83  &  0.395  & 0.275 & 13.22 & 0.412 & 0.164 & 11.99\\  
\multicolumn{1}{c|}{}&	 	Pixel + Random   & 0.216  & 0.670 & 20.74  & 0.264  & 0.501 & 18.03 & 0.316 & 0.426 & 15.43 \\  
\multicolumn{1}{c|}{}&	 	Initial Pipeline   & 0.213  & 0.647 & 20.38  & 0.293  & 0.462 & 17.26 & 0.277 & 0.438 & 15.11 \\  
\multicolumn{1}{c|}{}&	 	Top1 + Offsets   &0.205  & 0.656 & 20.60  & 0.285  & 0.477 & 17.70 & 0.289 & 0.412 & 14.80\\  
\multicolumn{1}{c|}{}&	 	Top5 + Offsets   & 0.197  & 0.661 & 20.73  &  0.259  & 0.492 & 18.03 & 0.265 & 0.483 & 15.51 \\  
\multicolumn{1}{c|}{}&	 	Top3 w/o Offsets   & 0.210  & 0.648 & 20.44  & 0.275  & 0.474 & 17.36  & 0.269 & 0.460 & 15.33 \\  
	 		 	
	 	\midrule
\multicolumn{1}{c|}{NPP-Net}	& 	Top3 + Offsets   &  \textbf{0.188}    & \textbf{0.679}  &\textbf{21.01}       & \textbf{0.249} & \textbf{0.504} & \textbf{18.32} & \underline{0.263} & \textbf{0.485} & \textbf{15.93} \\   
	 	\bottomrule
	 	\vspace{2mm}
	 \end{tabular}}
	 	 		\caption{
	 	 		Comparison with baselines and NPP-Net variants for NPP completion and the metrics are evaluated only in unknown regions. The best and second-best results (excluding variants) are highlighted in bold and underline respectively. NPP-Net outperforms all the baselines on NRTDB and DTD. While Facade has some non-NPP images, NPP-Net can still outperform all other baselines except for Lama. 
	 	 		See the supplementary for the results tested on the full images.}
	 	\label{tab1:completion_baseline}
	\end{center}
	\vspace{-13mm}
	\end{table}

\noindent\textbf{Ablation Study:}
We perform three studies. First, we compare to a ``No Periodicity'' variant, which uses a standard coordinated-based MLP without a periodicity prior. Results in Table \ref{tab1:completion_baseline} show that it fails to understand the arrangement of tiled motifs. Note that the Facade dataset may have different performances because it has some non-NPP images.  
Second, to study the loss functions, we design three variants: (1) Only pixel loss, (2) Only patch loss, (3) Pixel loss and patch loss with random patch sampling. As discussed in Section \ref{single_image_optim}, the results in Table \ref{tab1:completion_baseline} and Figure \ref{method:loss_abaltion} show that NPP-Net outperforms other variants. 

Third, we study the effect of the periodicity augmentation (coarse level Top-K candidate periodicities and fine level offsets) by testing four variants: (1) Initial pipeline (No augmentation), (2) Top-1 with offsets, (3) Top-5 with offsets, (4) Top-3 without offsets. Table \ref{tab1:completion_baseline} shows that the initial pipeline performs the worst.
Larger K (Top-5) hurts the performance as more inaccurate periodicities may be included. But, smaller K (Top-1) also performs badly because the correct periodicity may not be included.  
A suitable K (Top-3) without offsets performs worse since offsets can help better handle smaller errors. 
With the appropriate K and offsets, NPP-Net generates the best results. See more studies in supplementary.

\noindent\textbf{Baselines:}
We compare against non-periodicity-guided and periodicity-guided baselines. 
For the former, we select two traditional methods, Image Quilting~\cite{image_quilting} and PatchMatch~\cite{barnes2009patchmatch}, that can handle pattern structure locally for some scenes with properly selected patch size. 
We then consider two learning-based methods DIP~\cite{ulyanov2018deep} (CNN-based) and Siren~\cite{sitzmann2020implicit} (MLP-based) trained on a single image for inpainting. 
We also choose several learning-based methods: PEN-Net~\cite{yan2019PENnet}, ProFill~\cite{zeng2020high} and Lama~\cite{suvorov2022resolution} that are trained on large real-world datasets~\cite{zhou2017places,cimpoi14describing,Segmentation2017,karras2017progressive} since they show competitive NPP completion examples in their work. 

\begin{figure*}
    \centering
    \begin{tabular}{cc}
        \includegraphics[width=0.99\textwidth]{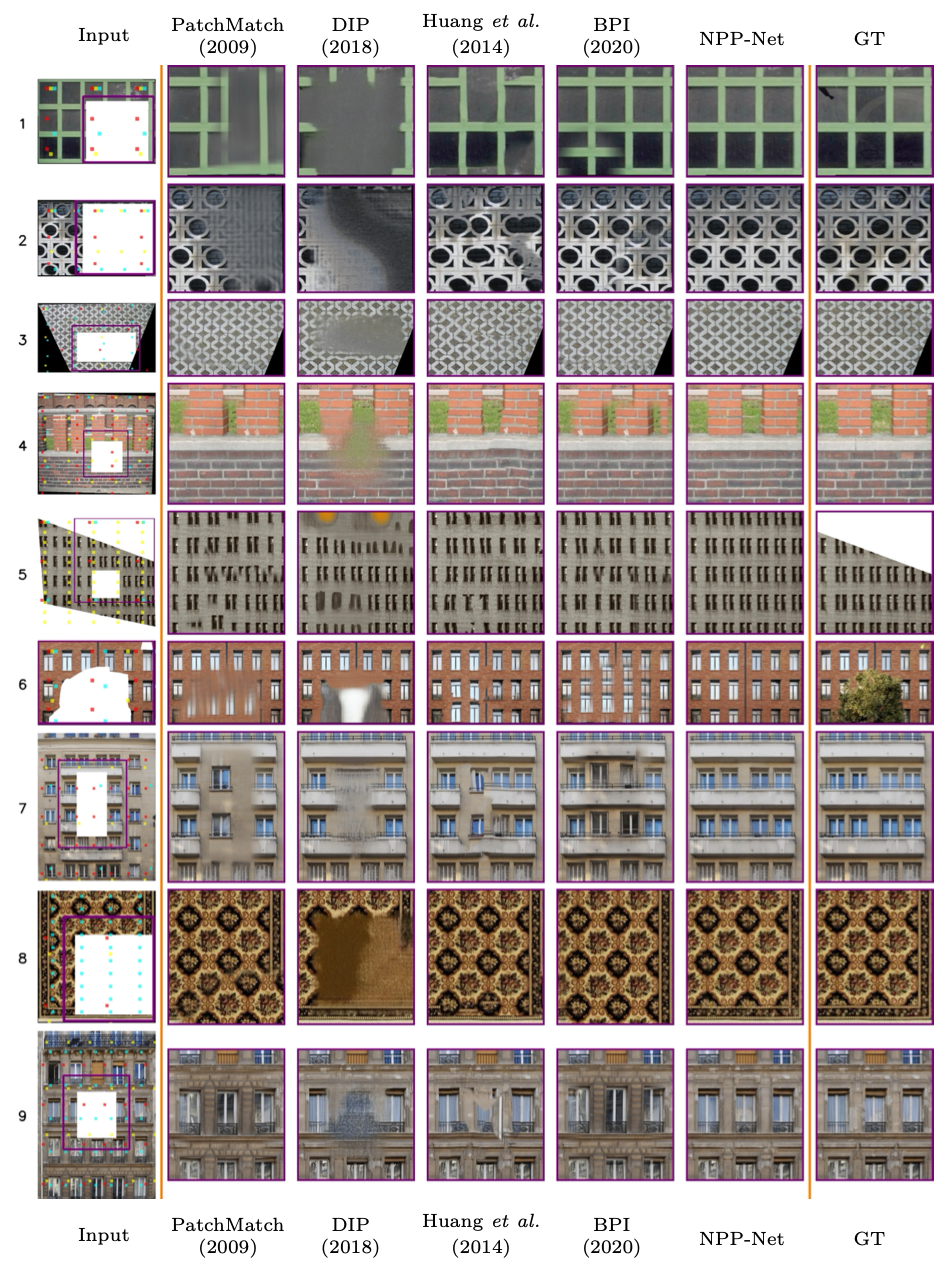}\\
    \end{tabular}
    \caption{Qualitative results for NPP completion. We show four baselines that operate on a single image. The red, yellow, and cyan dots in input images show the first, second, and third periodicity from periodicity searching module, respectively. For visualization, all periods are scaled by 2. 
    Our NPP-Net outperforms all baselines for global consistency (rows 1-6) and local variations (rows 6-9).
    }
    \label{comparison_baseline_completion}
 \end{figure*}

For periodicity-guided methods, we choose two baselines - Huang \etal~\cite{huang2014image} and BPI~\cite{li2020multi}. 
Both works were designed for multi-planar scenes, but can be used for single-plane completion as well. BPI first segments and rectifies planes, then performs periodicity detection~\cite{li2020perspective} on each plane, and inpaints each plane independently. 
For fair comparison in single-plane NPP image, we only compare with BPI's completion step to remove potential inconsistency introduced from other steps (\eg plane rectification). 
For Huang~\etal, we use their pipeline without modification as their method works directly for a single plane and the completion step cannot be easily separated out.
We will also compare with these two methods in multi-plane NPP images.

\begin{figure*}[!tb]
    \centering
    \begin{tabular}{cc}
        \includegraphics[width=0.99\textwidth]{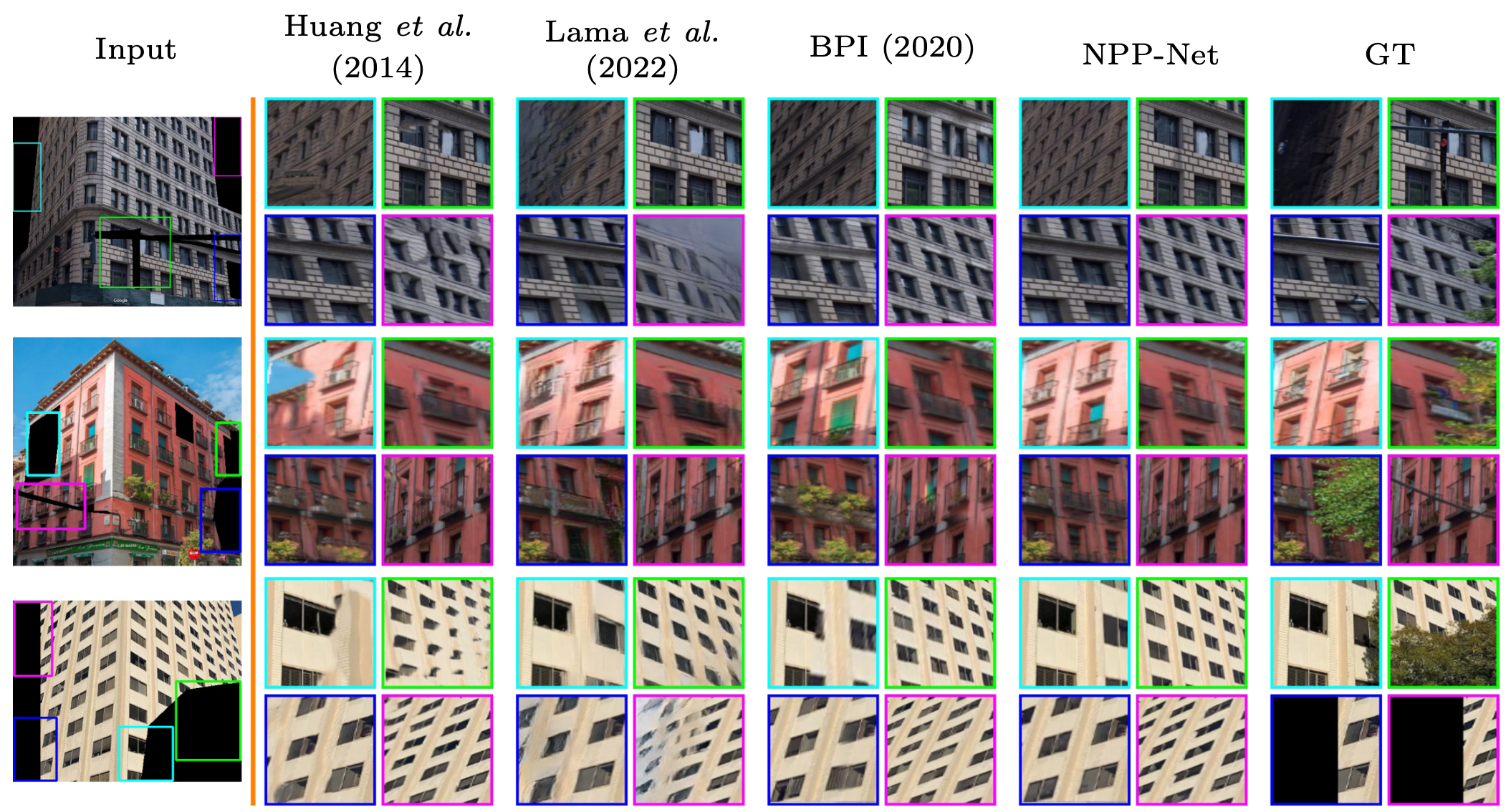}\\
    \end{tabular}
    \caption{ 
 Qualitative comparison for multi-plane NPP completion. We show three baselines, which are either designed for multi-plane NPP scenes (Huang \etal and BPI) or trained on large datasets (Lama). Some zoom-in boxes are resized for visualization. Full results are in supplementary.
    }
    \label{multi_plane_example}
 \end{figure*}

\noindent\textbf{Comparison with Baselines:}
 Table \ref{tab1:completion_baseline} shows the quantitative results for all the methods.
For NRTDB and DTD datasets, among the single-image baselines, BPI obtains the almost best LPIPS because it generates a more reasonable global structure guided by periodicity. 
PatchMatch obtains better SSIM and PSNR even if it generates blurred results for large masks.
Lama achieves the best results among the baselines since it adopts fast Fourier convolution for a larger image receptive field, which allows it to implicitly learn the underlying periodicity from large datasets. 
Our NPP-Net outperforms all baselines on these two datasets by optimizing only on a single image.
For the Facade dataset, even if some non-NPP images are included, NPP-Net performs better than the other baselines (except for Lama).
Lama effectively learns scene prior from large datasets and thus works well for non-NPP images, leading to the best performance in this dataset.

Qualitative results are shown in Figure \ref{comparison_baseline_completion}. 
Large rectangle masks are challenging since there is less information from which to estimate the representation.
Perceptually, PatchMatch works well when the motifs are small (row 3) but results in blur with large masks (row 2 and 6).
Although BPI and Huang \etals perform better than non-periodicity-guided baselines, they generate artifacts since the NPP representation (periodicity) has poor global consistency (row 1-6) or lacks local variations (row 6-9). 
Note that the Top-1 periodicity in row 1 (red dots) is inaccurate - the actual periods are half of the one shown.
We show that NPP-Net can extrapolate NPP images well (row 5), generalize to irregular masks (row 6), and work for scenes that contain non-periodic regions (row 4).
We show that NPP-Net can be extended to handle multi-planar scenes in Figure \ref{multi_plane_example}. Among the baselines, Lama (trained on large datasets) can better handle local variations (row 2 purple box). Although it captures some global structure when the mask is not large, Lama performs worse than BPI when the mask is outside the image border for extrapolation (all rows) and perspective effect is severe (row 2 cyan box). Learning from the Top-K periodicities, NPP-Net produces the best images, maintaining global consistency and local variations. Finally, the output of NPP-Net can in turn improve periodicity detection, leading to better image completion for both BPI and NPP-Net (see supplementary).

\noindent\textbf{Influence of Mask Size:}
We conduct two experiments to study the influence of different mask sizes for image completion. 

First, for each image in NRTDB dataset, if the K-th periodicity has the smallest error among Top-3 periodicities, we assign the image to the K-th periodicity.
We show the number of images assigned to each periodicity with different mask sizes in Figure \ref{comparison_mask_ratio} (left).
While the Top 1st periodicity is the best one for most of the images with small mask size, this number decreases in the large mask case (64\% of the image). This demonstrates that the other periodicities contain better periodicity and leveraging them by our periodicity augmentation strategy can be helpful for learning NPP representation, especially when the mask is large. 

To compute the periodicity error, we manually annotate the periodicity with  the smallest period as ground truth periodicity for the dataset. For each periodicity, we generate a 2D point cloud, defined as $\{\alpha \vect{d}_1 + \beta\vect{d}_2 | \alpha, \beta \in \mathbb{Z}\}$. We also filter points that are out of image range. The periodicity error is calculated using the average L2 distance between every point in proposed point clouds and its nearest neighbor in the ground truth point cloud (one-directional chamfer distance).

Second, we show the LPIPS performances for different mask sizes in Figure \ref{comparison_mask_ratio} (right). In particular, we filter out images that contain large non-periodic regions.
When the mask area is small (4\% of the image), PatchMatch slightly outperforms NPP-Net because the unknown regions may not contain pattern structure, and simply sampling nearby patches is sufficient to produce good results. 
Among single-image methods, Huang \etal, BPI and NPP-Net perform better when the mask size increases since they are guided by periodicity. 
Taking better advantage of periodicity information, NPP-Net is more robust to various mask sizes, especially for larger masks.

\begin{figure}[!t]
\centering
    \captionsetup[subfigure]{labelformat=empty}
        \begin{tabular}{cccc}
         \begin{subfigure}[c]{0.49\linewidth}
  {\includegraphics[width=4.96cm]{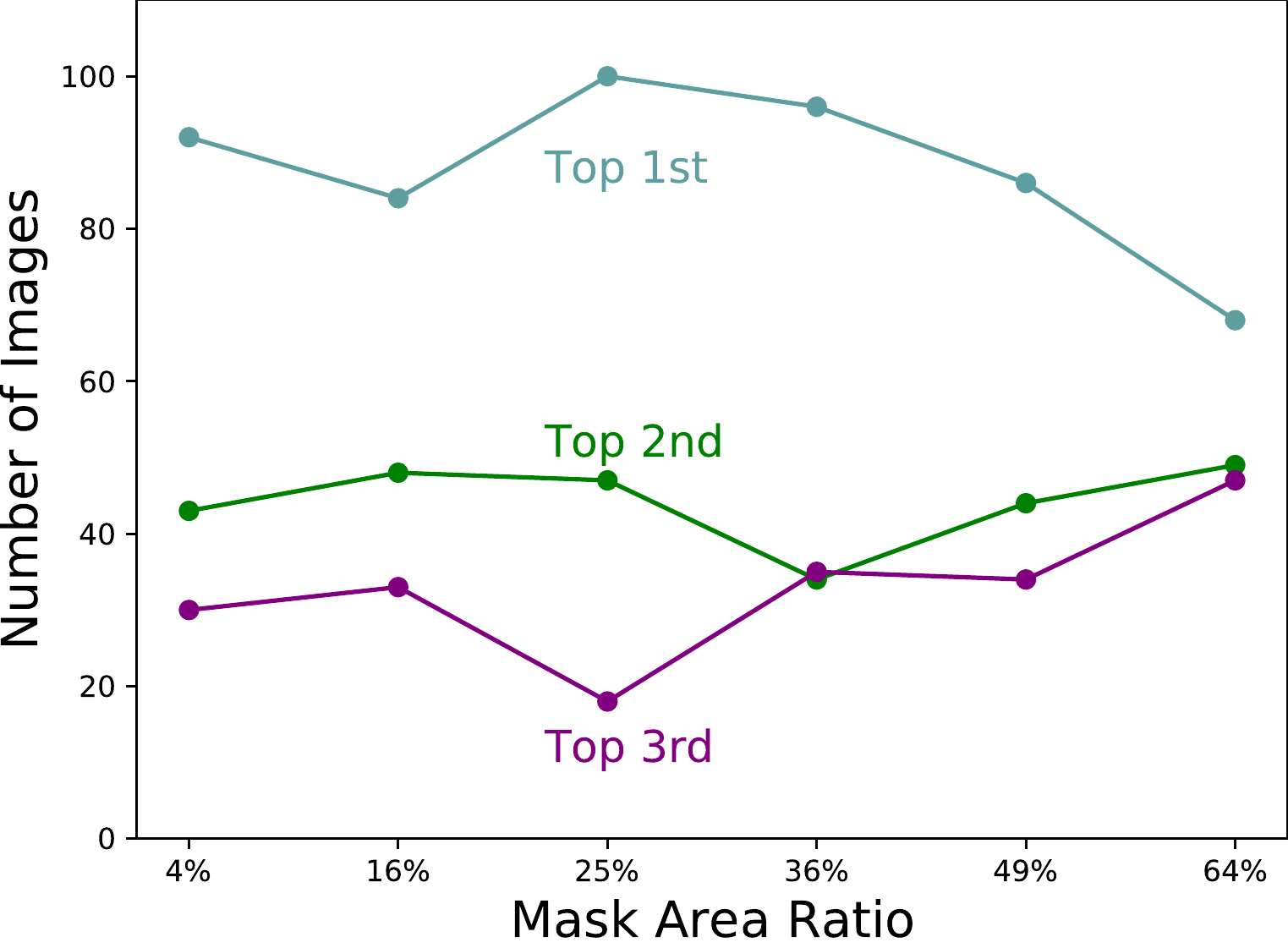}}
\end{subfigure}
 \begin{subfigure}[c]{0.49\linewidth}
  {\includegraphics[width=5.46cm]{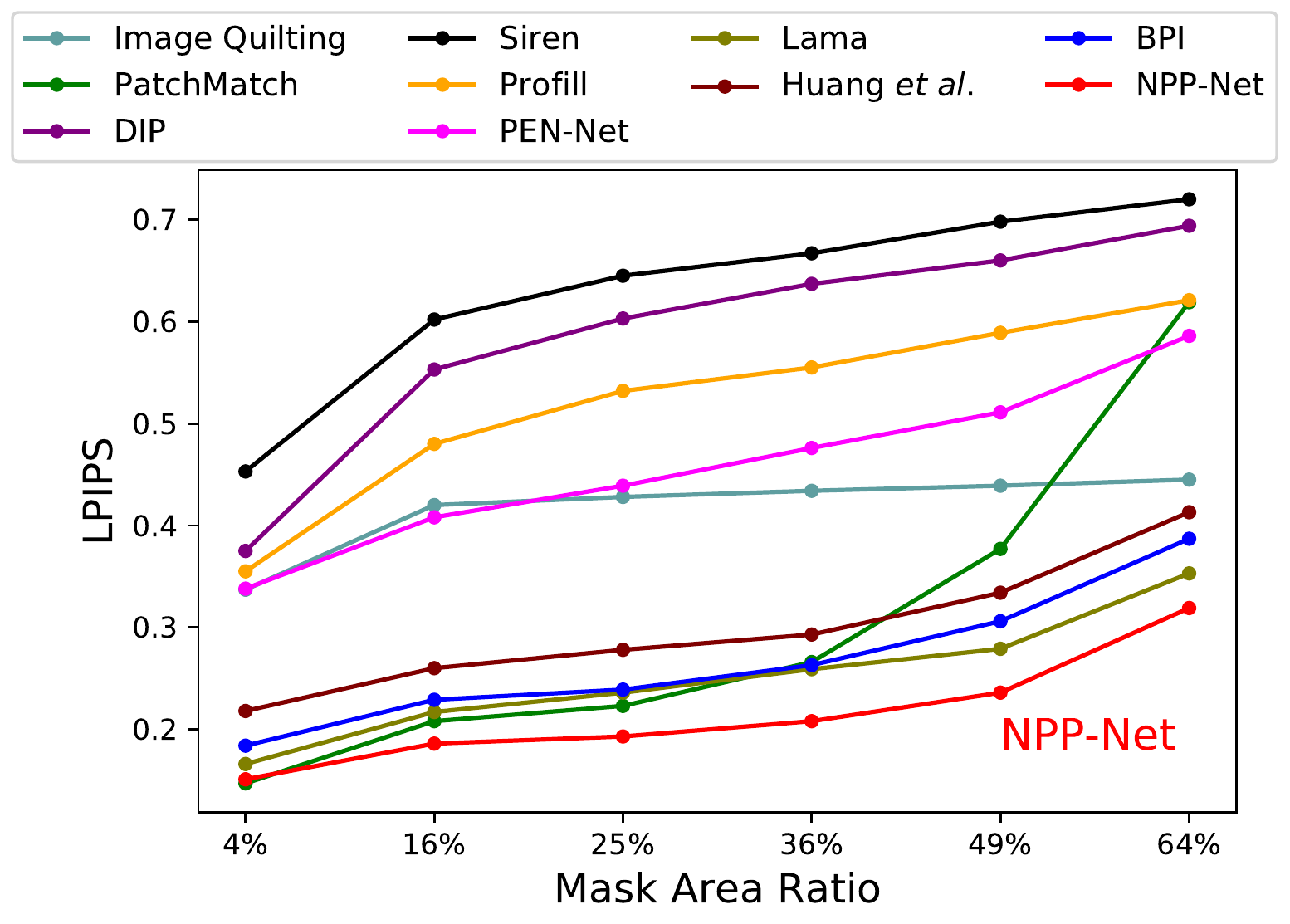}}
\end{subfigure}
\end{tabular}
    \caption{ 
    \textbf{Left}: 
    The number of NPP images for each periodicity that has the smallest periodicity errors (among Top-3) with different mask sizes in NRTDB.  
    As the mask size grows, the best periodicity emerges in the Top 2nd and 3rd periodicity, thus utilizing them in NPP-Net is useful.
    \textbf{Right}: The LPIPS results (lower is better) for different mask sizes in NRTDB.  PatchMatch performs the best when masks are very small but NPP-Net outperforms all the baselines for large masks.
    }
    \label{comparison_mask_ratio}
\end{figure}

\noindent\textbf{Limitations:}
(1) The periodicity proposal cannot be too erroneous, allowing tolerance of about 10\%.
(2) It assumes a multi-planar scene with translated, circular, and potentially other types of symmetrical NPP that can be modeled.

\section{Conclusion}
In conclusion, we show how to learn an effective implicit neural representation of  Near-Periodic Patterns. We design the periodicity proposal, periodicity-aware input warping, periodicity-guided patch loss to maintain global consistency and local variations. We compare NPP-Net with nine baselines and eight variants on three datasets to demonstrate its effectiveness. We believe that NPP-Net is a strong tool to understand a large class of man-made scenes.

\noindent\textbf{Acknowledgement:} This work was supported by a gift from Zillow Group, USA, and NSF Grants \#CNS-2038612,  \#IIS-1900821.

\clearpage
%
%
\bibliographystyle{splncs04}
\bibliography{egbib}
\end{document}


\pagestyle{headings}
\mainmatter
\def\ECCVSubNumber{2589}  

\title{Learning Continuous Implicit Representation for Near-Periodic Patterns}
\newcommand{\etal}{\emph{et al.}}
\newcommand{\etals}{\emph{et al.} }

\newcommand{\eg}{\emph{e.g}., }
\newcommand{\ie}{\emph{i.e}., }


\titlerunning{NPP-Net}
%
\author{Bowei Chen\inst{1}\index{Chen, Bowei} \and
Tiancheng Zhi\inst{1} \and
Martial Hebert\inst{1} \and 
Srinivasa G. Narasimhan\inst{1}
}
\authorrunning{B. Chen et al.}
%
\institute{ Carnegie Mellon University, Pittsburgh PA 15213, USA 
\email{\{boweiche,tzhi,mhebert,srinivas\}@andrew.cmu.edu}
}
\setcounter{tocdepth}{2}
{\def\addcontentsline#1#2#3{}\maketitle}

  \hypersetup{linkcolor=black}
\begingroup
\let\clearpage\relax
{ \hypersetup{hidelinks} \tableofcontents }
\endgroup
\newpage

\section{Dataset}

For PSU Near-Regular Texture Database (NRTDB), we collect all the perspective NPP images, and filter those that fail to be rectified using TILT~\cite{zhang2012tilt}. The resultant dataset has 165 NPP images from PSU Near-Regular Texture Database (NRTDB), which contains facades, friezes, bricks, fences, grounds, Mondrian images, wallpapers, and carpets.

For Describable Textures Dataset (DTD), we only consider the official test set (the first split) since some of our baselines are trained on DTD~\cite{cimpoi14describing}. In the test set, we select the categories whose images are more likely to be near-periodic. These categories are banded, chequered, grid, honeycombed, lined, meshed, perforated, polka-dotted, zigzagged. Then we manually filter out non-NPP images, resulting in 258 NPP images in the dataset. 

For the Facade dataset, we directly use an official CVPR 2010 subset provided by \cite{teboul2010segmentation}, which contains 109 rectified images of facades. Some of these facades are strictly \textit{not} NPP because often the windows are not arranged periodically. But nonetheless we include these to evaluate our approach when the NPP assumption is not strictly satisfied.

\begin{figure}[!h]
    \captionsetup[subfigure]{labelformat=empty, font=small}
\begin{subfigure}{.24\linewidth}
  \centerline{\includegraphics[width=.94\textwidth]{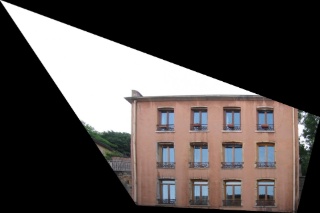}}
\end{subfigure}
\hfill
\begin{subfigure}{.24\linewidth}
  \centerline{\includegraphics[width=.94\textwidth]{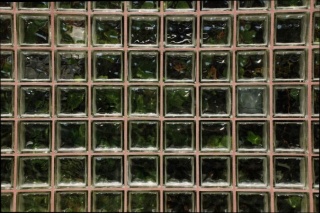}}
\end{subfigure}
\hfill
\begin{subfigure}{.24\linewidth}
  \centerline{\includegraphics[width=.94\textwidth]{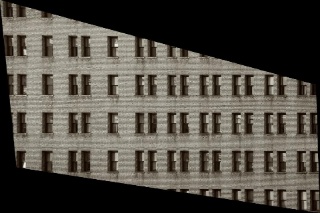}}
\end{subfigure}
\begin{subfigure}{.24\linewidth}
  \centerline{\includegraphics[width=.94\textwidth]{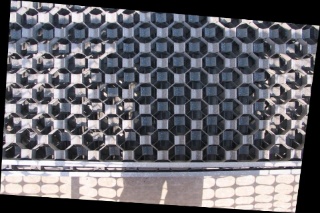}}
\end{subfigure}
\\
\begin{subfigure}{.24\linewidth}
  \centerline{\includegraphics[width=.94\textwidth]{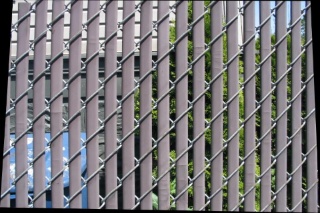}}
\end{subfigure}
\hfill
\begin{subfigure}{.24\linewidth}
  \centerline{\includegraphics[width=.94\textwidth]{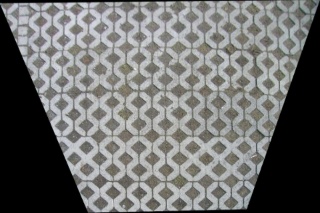}}
\end{subfigure}
\hfill
\begin{subfigure}{.24\linewidth}
  \centerline{\includegraphics[width=.94\textwidth]{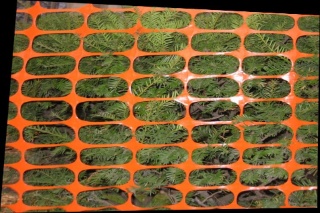}}
\end{subfigure}
\begin{subfigure}{.24\linewidth}
  \centerline{\includegraphics[width=.94\textwidth]{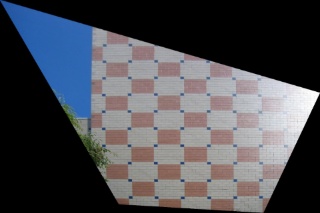}}
\end{subfigure}
\\
\begin{subfigure}{.24\linewidth}
  \centerline{\includegraphics[width=.94\textwidth]{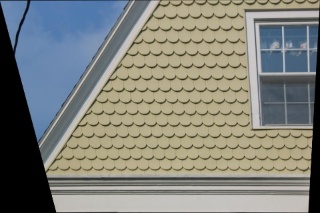}}
\end{subfigure}
\hfill
\begin{subfigure}{.24\linewidth}
  \centerline{\includegraphics[width=.94\textwidth]{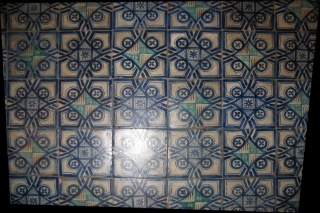}}
\end{subfigure}
\hfill
\begin{subfigure}{.24\linewidth}
  \centerline{\includegraphics[width=.94\textwidth]{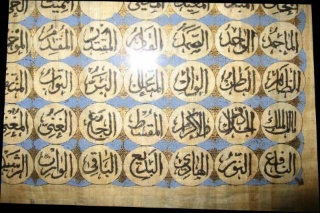}}
\end{subfigure}
\begin{subfigure}{.24\linewidth}
  \centerline{\includegraphics[width=.94\textwidth]{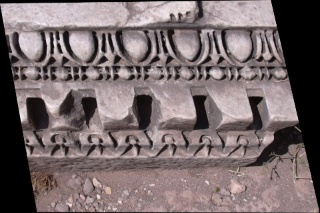}}
\end{subfigure}
\\
\begin{subfigure}{.24\linewidth}
  \centerline{\includegraphics[width=.94\textwidth]{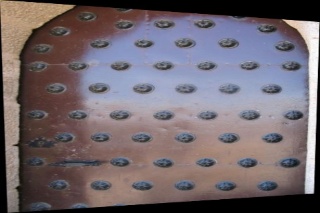}}
\end{subfigure}
\hfill
\begin{subfigure}{.24\linewidth}
  \centerline{\includegraphics[width=.94\textwidth]{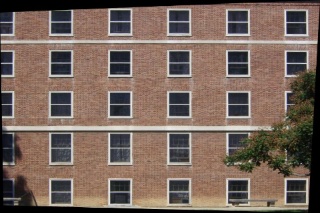}}
\end{subfigure}
\hfill
\begin{subfigure}{.24\linewidth}
  \centerline{\includegraphics[width=.94\textwidth]{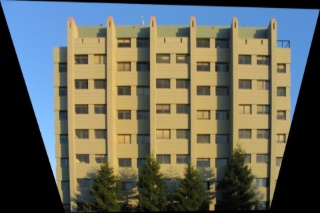}}
\end{subfigure}
\begin{subfigure}{.24\linewidth}
  \centerline{\includegraphics[width=.94\textwidth]{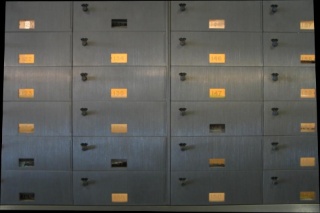}}
\end{subfigure}
\caption{
Sampled NPP images (rectified) of the NRTDB dataset. All examples are resized for the visualization. More images are included in the qualitative comparison. 
}
\label{dataset:NRTDB}
\end{figure}

\begin{figure}[!h]
\vspace{-15mm}
    \captionsetup[subfigure]{labelformat=empty, font=small}
\begin{subfigure}{.24\linewidth}
  \centerline{\includegraphics[width=.94\textwidth]{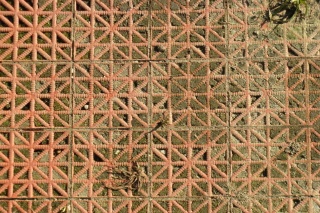}}
\end{subfigure}
\hfill
\begin{subfigure}{.24\linewidth}
  \centerline{\includegraphics[width=.94\textwidth]{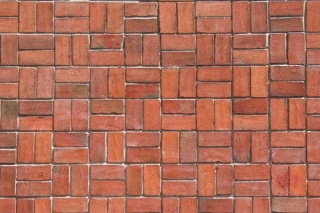}}
\end{subfigure}
\hfill
\begin{subfigure}{.24\linewidth}
  \centerline{\includegraphics[width=.94\textwidth]{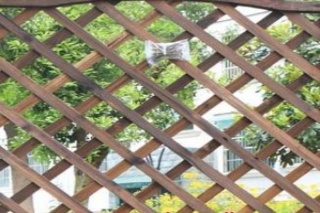}}
\end{subfigure}
\begin{subfigure}{.24\linewidth}
  \centerline{\includegraphics[width=.94\textwidth]{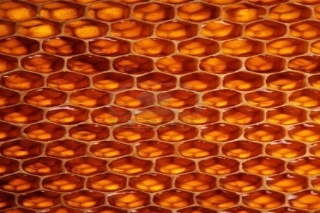}}
\end{subfigure}
\\
\begin{subfigure}{.24\linewidth}
  \centerline{\includegraphics[width=.94\textwidth]{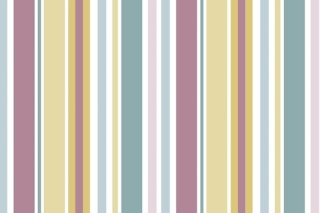}}
\end{subfigure}
\hfill
\begin{subfigure}{.24\linewidth}
  \centerline{\includegraphics[width=.94\textwidth]{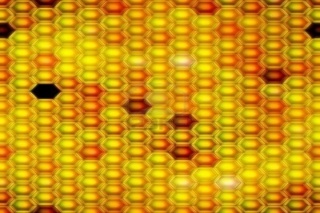}}
\end{subfigure}
\hfill
\begin{subfigure}{.24\linewidth}
  \centerline{\includegraphics[width=.94\textwidth]{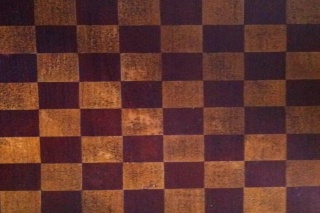}}
\end{subfigure}
\begin{subfigure}{.24\linewidth}
  \centerline{\includegraphics[width=.94\textwidth]{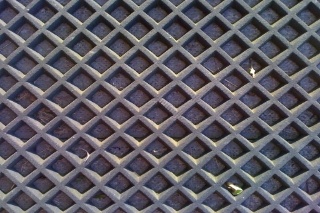}}
\end{subfigure}
\\
\begin{subfigure}{.24\linewidth}
  \centerline{\includegraphics[width=.94\textwidth]{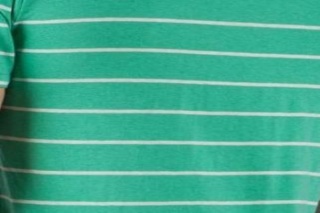}}
\end{subfigure}
\hfill
\begin{subfigure}{.24\linewidth}
  \centerline{\includegraphics[width=.94\textwidth]{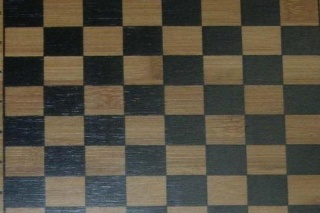}}
\end{subfigure}
\hfill
\begin{subfigure}{.24\linewidth}
  \centerline{\includegraphics[width=.94\textwidth]{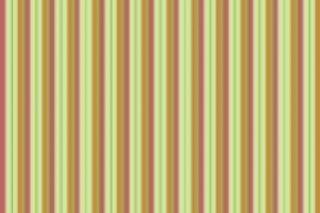}}
\end{subfigure}
\begin{subfigure}{.24\linewidth}
  \centerline{\includegraphics[width=.94\textwidth]{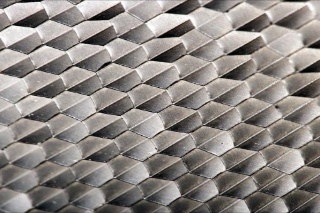}}
\end{subfigure}
\\
\begin{subfigure}{.24\linewidth}
  \centerline{\includegraphics[width=.94\textwidth]{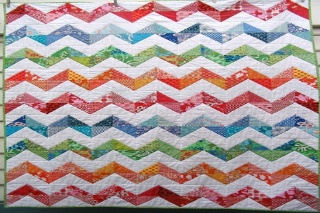}}
\end{subfigure}
\hfill
\begin{subfigure}{.24\linewidth}
  \centerline{\includegraphics[width=.94\textwidth]{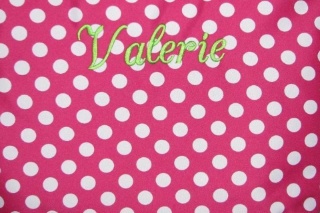}}
\end{subfigure}
\hfill
\begin{subfigure}{.24\linewidth}
  \centerline{\includegraphics[width=.94\textwidth]{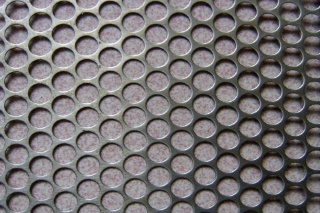}}
\end{subfigure}
\begin{subfigure}{.24\linewidth}
  \centerline{\includegraphics[width=.94\textwidth]{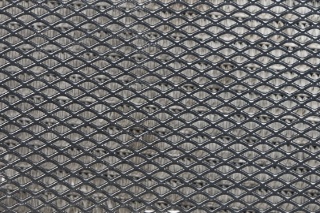}}
\end{subfigure}
\caption{
Sampled NPP images (rectified) of the DTD dataset. All examples are resized for the visualization. More images are included in the qualitative comparison. 
}
\label{dataset:DTD}
\end{figure}

\begin{figure}[!h]
    \captionsetup[subfigure]{labelformat=empty, font=small}
\begin{subfigure}{.23\linewidth}
  \centerline{\includegraphics[width=0.94\textwidth]{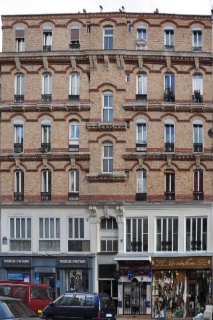}}
\end{subfigure}
\hfill
\begin{subfigure}{.23\linewidth}
  \centerline{\includegraphics[width=0.94\textwidth]{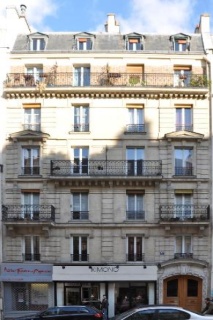}}
\end{subfigure}
\hfill
\begin{subfigure}{.23\linewidth}
  \centerline{\includegraphics[width=0.94\textwidth]{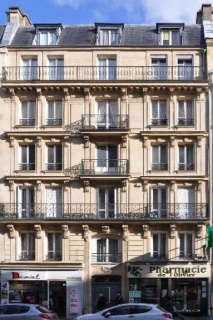}}
\end{subfigure}
\begin{subfigure}{.23\linewidth}
  \centerline{\includegraphics[width=0.94\textwidth]{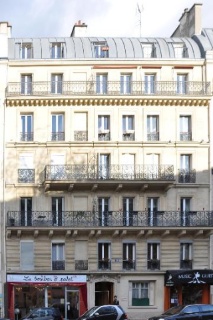}}
\end{subfigure}
\\
\begin{subfigure}{.23\linewidth}
  \centerline{\includegraphics[width=0.94\textwidth]{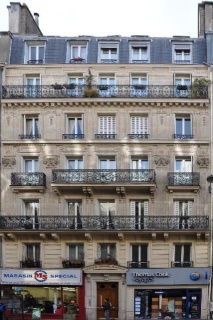}}
\end{subfigure}
\hfill
\begin{subfigure}{.23\linewidth}
  \centerline{\includegraphics[width=0.94\textwidth]{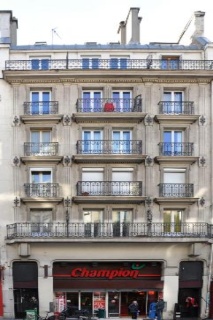}}
\end{subfigure}
\hfill
\begin{subfigure}{.23\linewidth}
  \centerline{\includegraphics[width=0.94\textwidth]{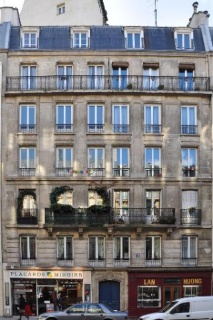}}
\end{subfigure}
\begin{subfigure}{.23\linewidth}
  \centerline{\includegraphics[width=0.94\textwidth]{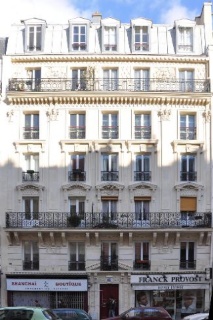}}
\end{subfigure}
\\
\begin{subfigure}{.23\linewidth}
  \centerline{\includegraphics[width=0.94\textwidth]{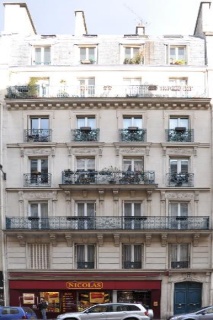}}
\end{subfigure}
\hfill
\begin{subfigure}{.23\linewidth}
  \centerline{\includegraphics[width=0.94\textwidth]{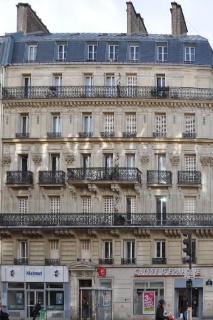}}
\end{subfigure}
\hfill
\begin{subfigure}{.23\linewidth}
  \centerline{\includegraphics[width=0.94\textwidth]{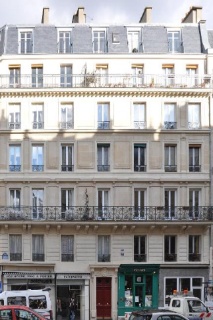}}
\end{subfigure}
\begin{subfigure}{.23\linewidth}
  \centerline{\includegraphics[width=0.94\textwidth]{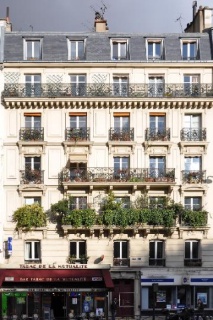}}
\end{subfigure}
\\
\begin{subfigure}{.23\linewidth}
  \centerline{\includegraphics[width=0.94\textwidth]{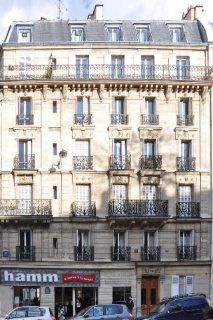}}
\end{subfigure}
\hfill
\begin{subfigure}{.23\linewidth}
  \centerline{\includegraphics[width=0.94\textwidth]{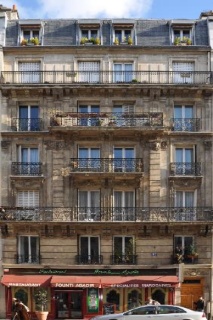}}
\end{subfigure}
\hfill
\begin{subfigure}{.23\linewidth}
  \centerline{\includegraphics[width=0.94\textwidth]{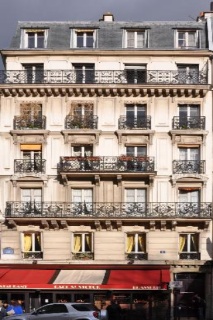}}
\end{subfigure}
\begin{subfigure}{.23\linewidth}
  \centerline{\includegraphics[width=0.94\textwidth]{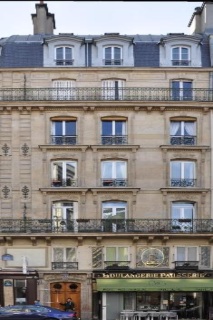}}
\end{subfigure}
\caption{
Sampled NPP images (rectified) of the Facade dataset. All examples are resized for the visualization. More images are included in the qualitative comparison. 
}
\label{dataset:Facade}
\end{figure}

The sampled images of NRTDB, DTD, Facade datasets are shown in Figure \ref{dataset:NRTDB}, Figure \ref{dataset:DTD} and Figure \ref{dataset:Facade}, respectively.
For our self-collected dataset, we show all of them in qualitative results. 


\newpage

\section{NPP-Net}

\subsection{Displacement Vectors Transformation}
	
 \begin{figure*}[h]
\begin{center}
\includegraphics[width=0.4\linewidth,valign=t]{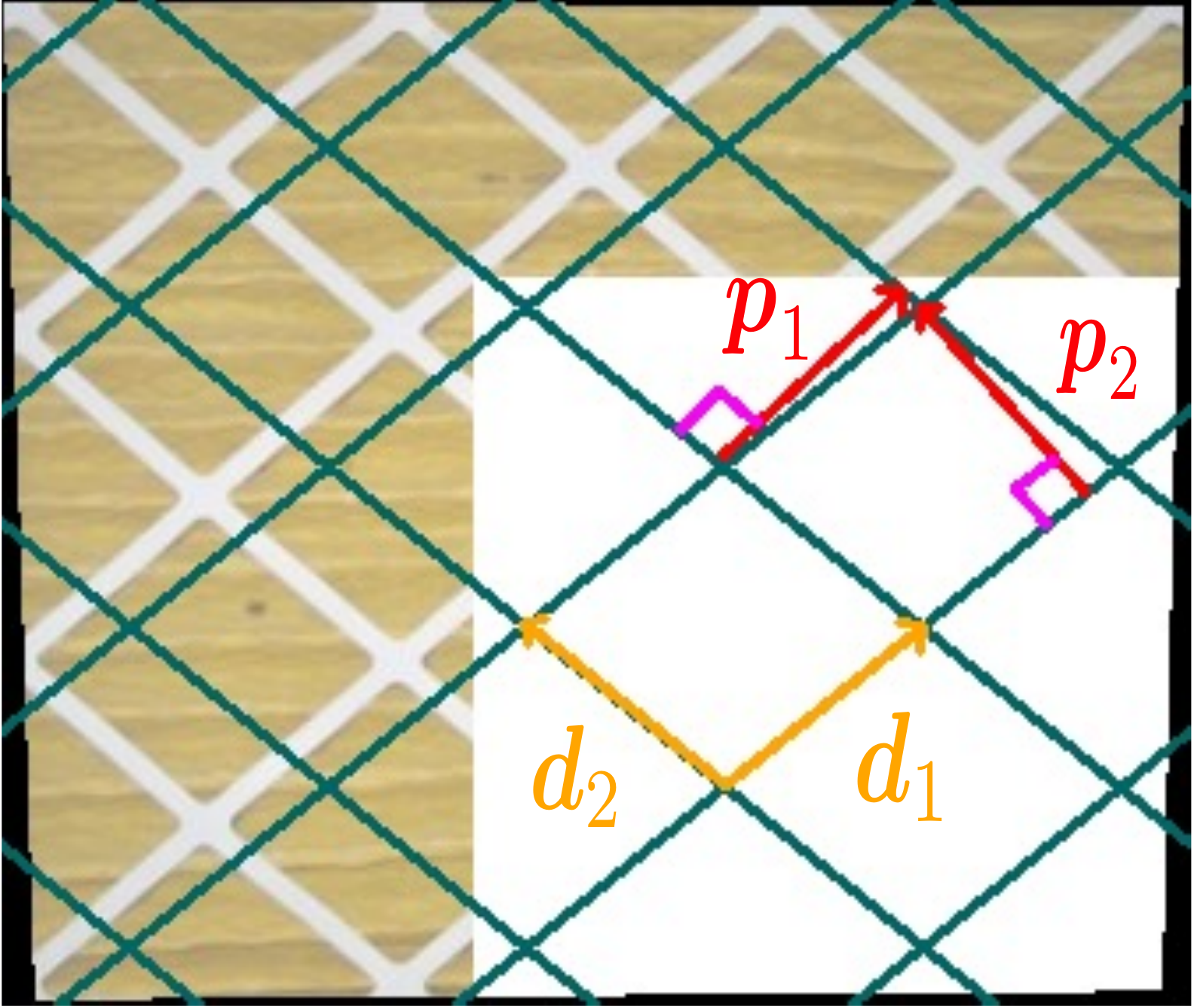}
\end{center} 
    \vspace{-4mm}
    \caption{  The illustration of a periodicity. It consists of two displacement vectors  $\vect{d}_1$ and $\vect{d}_2$ (orange), which can be transformed to periodicity vectors $\vect{p}_1$ and $\vect{p}_2$ (red), respectively. 
    }
    \label{supp_trans}
 \end{figure*}

\subsubsection{Lattice Patterns:}
 
Assuming a 2D lattice arrangement, a periodicity can be represented as two displacement vectors $\vect{d}_1$ and $\vect{d}_2$ (orange arrows in Figure \ref{supp_trans}). A perfect infinite periodic pattern is invariant if shifted by $\alpha \vect{d}_1 + \beta\vect{d}_2 (\alpha, \beta \in \mathbb{Z})$. 
In order to incorporate the displacement vectors into the  Periodicity-Aware Input Warping module, we need to transform $\vect{d}_1$ and $\vect{d}_2$ into pattern periods and orientations, visualized as the magnitudes and orientations of the red arrows ($\vect{p}_1$ and $\vect{p}_2$, called \textbf{periodicity vectors}).

We assume $\vect{d}_1 = (d_{1x}, d_{1y})$ and $\vect{d}_2 = (d_{2x}, d_{2y})$ are known (from Periodicity Proposal). Note that we let $d_{1x}, d_{2x} \in \mathbb{Z}$ and $d_{1y}, d_{2y} \in \mathbb{N}$ to remove identical vectors in another half circle. 
As we mentioned in the main paper, we can obtain the periodicity vectors by solving:
\begin{equation}
    \vect{p}_1 \cdot \vect{d}_2 = \vect{d}_1 \cdot \vect{p}_2 = \vect{0}.
    \label{Eq:dot}
\end{equation}
\begin{equation}
\vect{p}_1 \times \vect{d}_2 = \vect{d}_1 \times \vect{p}_2 = \vect{d}_1 \times \vect{d}_2,
    \label{Eq:cross}
\end{equation}
where the cross product is defined using the corresponding 3D vectors on the plane $z=0$.

Let $\vect{p_1} = (p_1 \cos\theta_1,  p_1 \sin \theta_1)$, where $\theta_1$ is in the range $[0, \pi)$. We can compute $\theta_1$ according to Equation \ref{Eq:dot}, given by:
\begin{equation}
    \begin{split}
        \vect{p_1} \cdot \vect{d_2} =  p_1 \cos\theta_1 d_{2x} + p_1 \sin\theta_1 d_{2y} &= 0 \\
        \tan\theta_1 &= - \frac{d_{2x}}{d_{2y}} \\
        \theta_1 &= \arctan - \frac{d_{2x}}{d_{2y}}
    \end{split}
\end{equation}
Then the magnitude $p_1$ can be solved based on Equation \ref{Eq:cross}, given by:
\begin{equation}
    \begin{split}
       \vect{p_1} \times \vect{d_2} &= \vect{d_1} \times \vect{d_2} \\
       p_1 |d_2| \sin\frac{\pi}{2}  &= |d_1| |d_2| \sin\phi \\
       p_1  &= \sqrt{(d_{1x})^2 + (d_{1y})^2}  \sin\phi
    \end{split}
    \label{mag}
\end{equation}
where $\phi = \arccos \frac{\vect{d_1} \cdot \vect{d_2}}{|d_1| \cdot |d_2|}$ is the angle between $\vect{d_1}$ and $\vect{d_2}$. Similarly, we can obtain $\vect{p_2}$ using Equation \ref{Eq:dot} to Equation \ref{mag}.

\subsubsection{Circular Patterns:}
Our method can be modified to handle circular periodic patterns. Define periodicity of circular NPP by a rotation centroid $\vect{c}=(c_x, c_y)$ and an angular period $p$ (in radians).
The periodicity proposal module can be applied to circular NPP to estimate $\vect{c}$ and $p$. 
In addition, two modifications are needed:

(1) Equation 1 (main paper) is rewritten as two functions:
\begin{equation}
    \begin{split}
       f_{\vect{c},p}(x, y) = \sqrt{(x-c_x)^2 + (y-c_y)^2} \\
       g_{\vect{c},p}(x, y) = atan2^*(y-c_y, x-c_x) \mod p
    \end{split}
    \label{circular_NPP}
\end{equation}
where we define $atan2^*(y,x) = (atan2(y,x) + 2\pi) \mod 2\pi$.

(2) Periodicity-based patch sampling strategy is modified. Let  center of predicted patch be $\vect{x}$. Instead of shifting $\vect{x}$ to obtain centers of known patch $\vect{x}_{\alpha\beta}$ in lattice patterns, we directly rotate the input image around $\vect{c}$ based on $\alpha \cdot p$, where $\alpha$ is an integer constant. Then we sample known patch centers in circular images at position $\vect{x}$.

\subsection{Implementation Details}
\label{ssec:implementation_details}
We apply the following settings below (for the final pipeline) for all applications. Same as \cite{mildenhall2020nerf}, we set the number of frequency $d$ in positional encoding as 10.

\textbf{Periodicity Proposal}. 
In our implementation, the key hyperparameter for periodicity detection method~\cite{li2020perspective} is an integer $q$, which defines the range of possible displacements in $\vect{d}_1$ and $\vect{d}_2$.
Specifically, $\vect{d}_1$ and $\vect{d}_2$ are in the set $R$, given by:
\scriptsize
\begin{equation}
    R = \{ (x, y) | -\frac{W}{q} < x < \frac{W}{q}, 0 \leq y < \frac{H}{q} \} - \{ (x, y) | -\frac{W}{q+1} < x < \frac{W}{q+1}, 0 \leq y < \frac{H}{q+1} \},
\end{equation}
\normalsize
where $H$ and $W$ are the image height and width.
The periodicity detection method outputs the best periodicity in $R$.

In the periodicity searching, we evaluate different $q$ in $\{i | i \in \mathbb{Z}^+, i < 10\}$ for multiple periodicities, which are ranked based on reconstruction errors. Specifically, we generate Top-M (M=3) pseudo square masks whose centers are far away from the image boundary or unknown regions. The mask size for each mask is empirically set to $\frac{5L}{6\sqrt{2}}$, where $L$ is the distance from the center pixel to the nearest invalid pixel. Then we run the initial pipeline for evaluation of each $q$ based on its reconstruction error in these masked regions.

\textbf{Network Architecture}. 
The first MLP contains 9 fully-connected Snake layers~\cite{NEURIPS2020_11604531} with 512 channels.
It also includes a skip connection that concatenates the Top-1 coordinate and original features to the fifth layer's activation. 
The outputs of this MLP are concatenated with the Top 2nd to K-th coordinate features and then fed to the second MLP.
The second MLP has 4 fully-connected Snake layers with 512, 512, 256, and 3 channels. The output features of the first MLP are also concatenated with the second layer activation of the second MLP for the skip connection. 

\textbf{Single-Image Optimization}.
In the patch loss, we sample square ground truth patches for supervision. We set the patch size $s$ in $\{64, 96, 128, 160\}$ based on the Top-1 periodicity, where larger periods are matched with a larger patch size.    
For the patch center $\vect{x}$, we filter out the shifted patch centers $\vect{x}_{\alpha\beta}$ with their known patch area smaller than 70\% of the whole patch area, and we use $N=3$ (size of $T_N$). We set $\lambda_p$ in $\mathcal{L}_{patch}$ to 0.4, and $\lambda_c$ to 1, 10, 5 in completion, remapping and segmentation (explained later), respectively. 
In final Loss $\mathcal{L}$, we set $\lambda_1 = 1$, $\lambda_2 = 0.001$, $|B_1| = 8192$, and $|B_2| = 2$. Further, $B_2$ contains half of  sampled centers in unknown regions and half of those in known regions. 

For training details, we use the Adam optimizer with the learning rate that starts with $5 \times 10^{-4}$ and decays every 500 epochs. Furthermore, for every 2000 epoch, we shrink the patch size by twice and increase the number of the sampled patches by twice to encourage the network to focus on finer-level details. 

\textbf{Runtime.}
The periodicity proposal takes 100 seconds (for 10 candidate periodicities).
The optimization takes around 8 minutes, 5 minutes, and 3 minutes for NPP completion, NPP remapping, and NPP segmentation to converge on a single NVIDIA Titan XP GPU, respectively.

\clearpage

\section{NPP Completion}
\vspace{-2mm}

\subsection{Mask Generation}

We apply different mask generation strategies to test the methods on different scales. 
For each image in NRTDB~\cite{NRTDatabase}, we generate an unknown mask with height and width sampled from $[100, 500]$.  The top-most and left-most coordinates of the mask are sampled from $[ \frac{H}{2} - 250, \frac{H}{2} + 250 ]$ and $[ \frac{W}{2} - 250, \frac{W}{2} + 250 ]$, respectively. We crop the mask when it exceeds the image boundary. 
For each image in DTD~\cite{cimpoi14describing}, we generate an unknown mask with height and width equal to 70\% of the image height and width (starting from the bottom right) respectively. This mask generation strategy allows us to evaluate the extrapolation performance of the methods. 
For each image in Facade, we generate a mask in the center of the image, where the mask size is $(\frac{H}{6}, \frac{W}{3})$.

\vspace{-4mm}

\subsection{More Metrics}

We also evaluate the models using FID~\cite{parmar2021cleanfid} and RMSE.
FID evaluates the distance between the distribution of ground truth and outputs images at the dataset level. But, FID requires all the images to be resized to a fixed size. Thus for images with low resolution, this resizing operation introduces aliasing even if we use the recent proposed clean FID implementation~\cite{parmar2021cleanfid}. RMSE is a pixel-wise metric to compare the difference of pixel values.

\vspace{-4mm}

\subsection{Ablation Study}

\subsubsection{More results for experiments in main paper:}
\begin{figure*}[h]
    \centering
\begin{small}
\begin{flushleft}
\hspace{20pt} 
 \shortstack[c]{ Input  \\ \;} 
\hspace{13pt} 
 \shortstack[c]{ No  \\ Periodicity  } 
\hspace{7pt} 
 \shortstack[c]{ Pixel \\ Only  } 
\hspace{18pt}  
 \shortstack[c]{ Patch \\ Only  } 
\hspace{14pt} 
 \shortstack[c]{ Pixel+ \\ Random  } 
\hspace{10pt}  
 \shortstack[c]{ NPP-Net  \\ \;} 
\hspace{8pt} 
 \shortstack[c]{ Ground  \\ Truth} 
\end{flushleft}
\end{small}
\vspace*{-8pt}
    \begin{tabular}{cc}
        \includegraphics[width=1\textwidth]{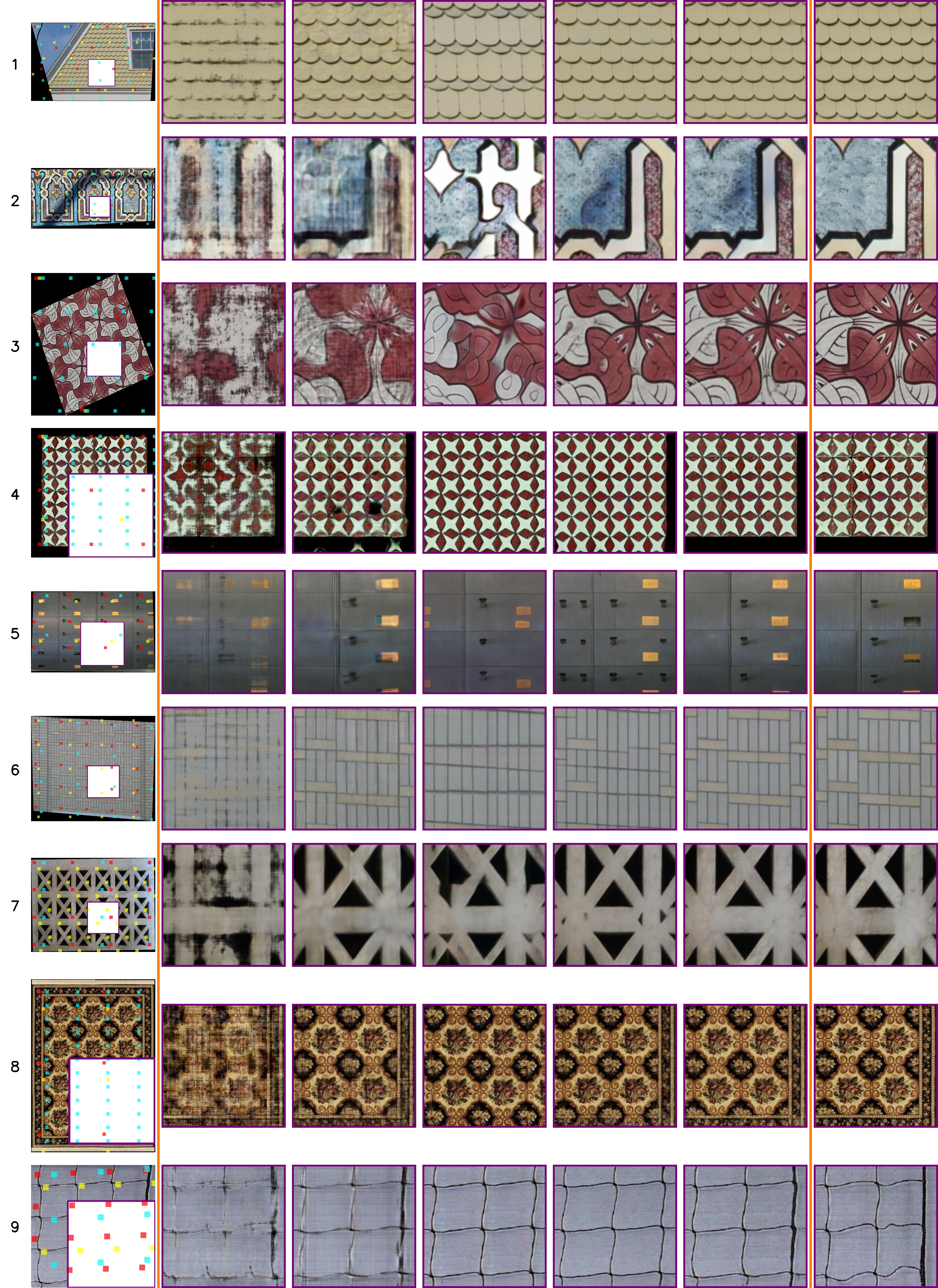}\\
    \end{tabular}
    \caption{ Qualitative results for variants without periodicity prior and with different loss functions. NPP-Net outperforms all other variants. Note that periods in the fifth and seventh row are not scaled by 2. 
    }
    \label{ablation: loss1}
 \end{figure*}

\begin{figure*}[h]
    \centering
\begin{small}
\begin{flushleft}
\hspace{20pt} 
 \shortstack[c]{ Input  \\ \;} 
\hspace{13pt} 
 \shortstack[c]{ No  \\ Periodicity  } 
\hspace{7pt} 
 \shortstack[c]{ Pixel \\ Only  } 
\hspace{18pt}  
 \shortstack[c]{ Patch \\ Only  } 
\hspace{14pt} 
 \shortstack[c]{ Pixel+ \\ Random  } 
\hspace{10pt}  
 \shortstack[c]{ NPP-Net  \\ \;} 
\hspace{8pt} 
 \shortstack[c]{ Ground  \\ Truth} 
\end{flushleft}
\end{small}
\vspace*{-8pt}
    \begin{tabular}{cc}
        \includegraphics[width=0.99\textwidth]{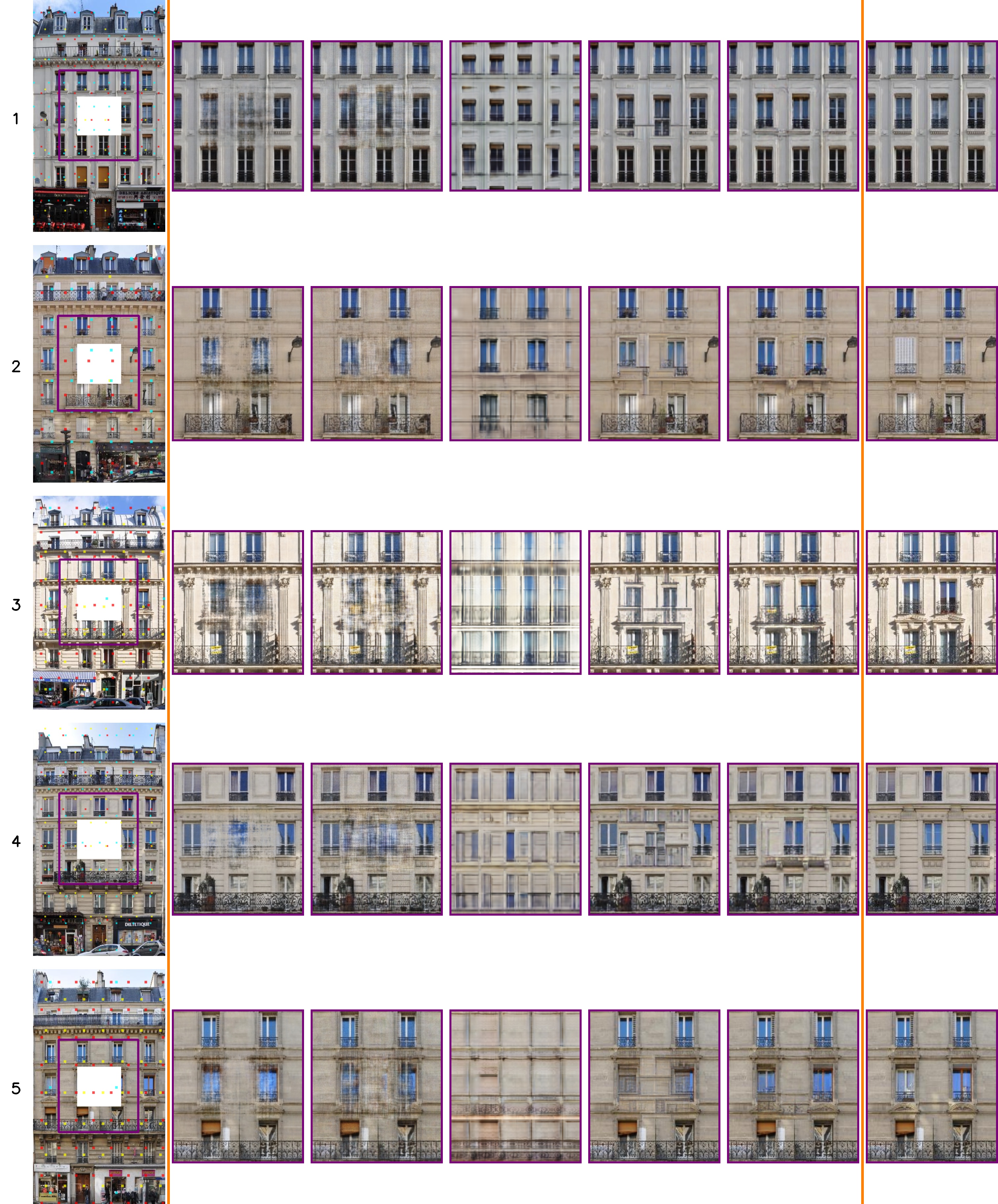}\\
    \end{tabular}
    \caption{ Qualitative results for variants without periodicity prior and with different loss function. NPP-Net outperforms all other variants. 
    }
    \label{ablation: loss2}
 \end{figure*}

\begin{figure*}[h]
    \centering
\begin{small}
\begin{flushleft}
\hspace{20pt} 
 \shortstack[c]{ Input  \\ \;} 
\hspace{13pt} 
 \shortstack[c]{ No  \\ Periodicity  } 
\hspace{7pt} 
 \shortstack[c]{ Pixel \\ Only  } 
\hspace{18pt}  
 \shortstack[c]{ Patch \\ Only  } 
\hspace{14pt} 
 \shortstack[c]{ Pixel+ \\ Random  } 
\hspace{10pt}  
 \shortstack[c]{ NPP-Net  \\ \;} 
\hspace{8pt} 
 \shortstack[c]{ Ground  \\ Truth} 
\end{flushleft}
\end{small}
\vspace*{-8pt}
    \begin{tabular}{cc}
        \includegraphics[width=0.99\textwidth]{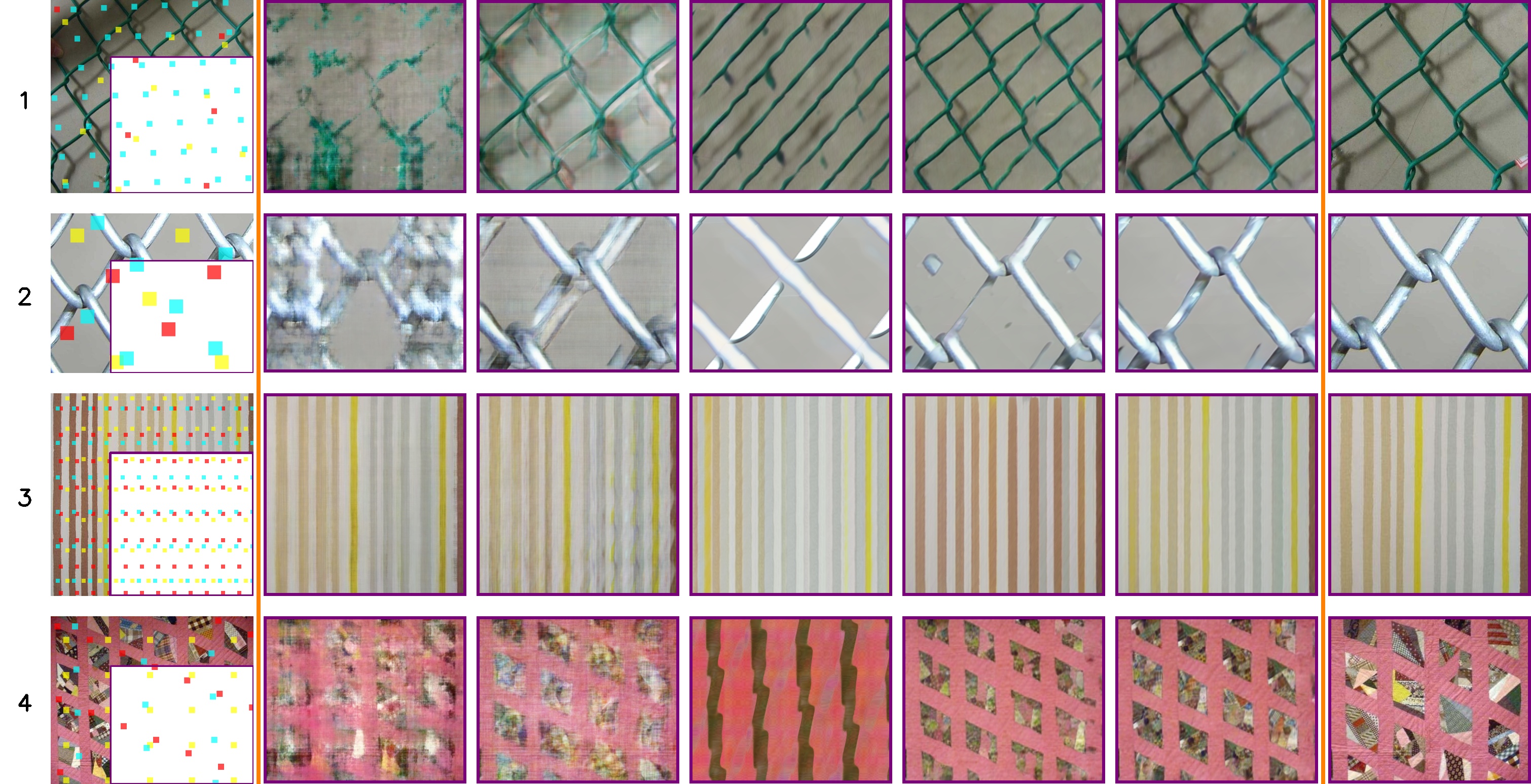}\\
    \end{tabular}
    \caption{ Qualitative results for variants without periodicity prior and with different loss function. NPP-Net outperforms all other variants.  
    }
    \label{ablation: loss3}
 \end{figure*}

\begin{figure*}[h]
 \vspace{-2mm}

    \centering
\begin{small}
\begin{flushleft}
\hspace{20pt} 
 \shortstack[c]{ Input   \\ \;} 
\hspace{15pt} 
 \shortstack[c]{ Initial \\ Pipeline   } 
\hspace{11pt} 
 \shortstack[c]{ Top1 + \\ Offsets  } 
\hspace{11pt}  
 \shortstack[c]{ Top5 + \\ Offsets   } 
\hspace{4pt} 
 \shortstack[c]{ Top3 w/o  \\ Offsets} 
\hspace{1pt}  
 \shortstack[c]{ Top3 + \\ Offsets   \\ (NPP-Net)} 
\hspace{1.2pt} 
 \shortstack[c]{ Ground \\ Truth   \\ \;} 

\end{flushleft}
\end{small}
\vspace*{-8pt}
    \begin{tabular}{cc}
        \includegraphics[width=0.99\textwidth]{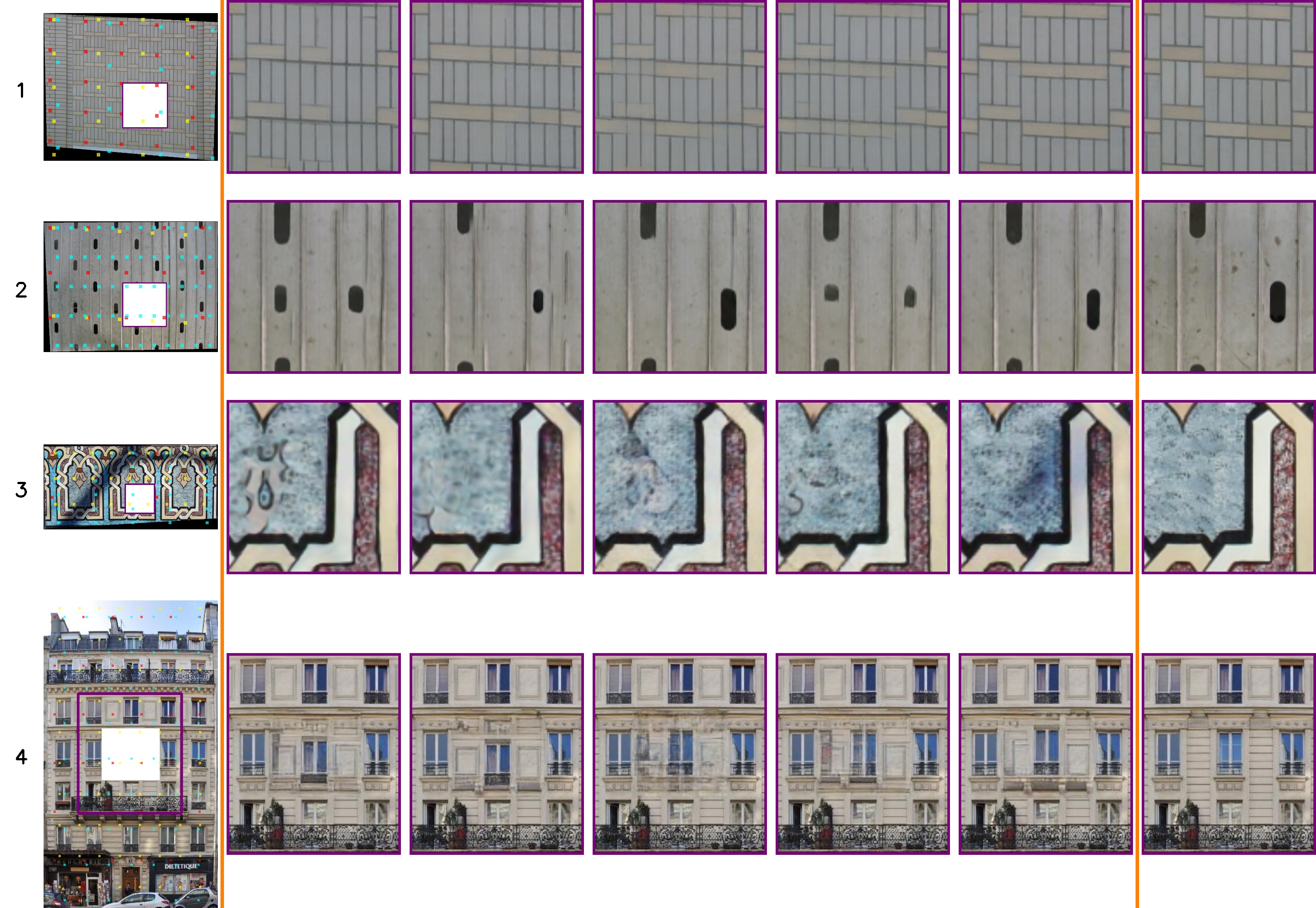}\\
    \end{tabular}
\vspace{-3mm}
    \caption{ Qualitative results for periodicity augmentation. NPP-Net outperforms all other variants.
    }
    \label{ablation: aug}
 \end{figure*}

We show the quantitative results with FID and RMSE (evaluated only in unknown regions) and all metrics (evaluated in the full images) in Table \ref{tab:ablation_nrtdb}, Table \ref{tab:ablation_dtd} and Table \ref{tab:ablation_facade}. 
Note that \enquote{No Periodicity} variant simply overfits the known regions without learning any structural information. Also, the facade dataset may have different performance because it contains some images that are strictly not NPP images. 
Figure \ref{ablation: loss1}, Figure \ref{ablation: loss2}, and Figure \ref{ablation: loss3} show the qualitative results for variants without periodicity prior and those with various combination of loss functions. Figure \ref{ablation: aug} shows the comparison with variants using different settings of periodicity augmentation.  In summary, NPP-Net outperforms all the tested variants on NRTDB and DTD datasets. 

\subsubsection{More experiments for choice of loss:}
We further tested three variants of NPP-Net in NRTDB: (1) replace robust L2 loss by L2 loss, (2) replace contextual loss by perceptual loss, (3) replace perceptual loss by contextual loss. The LPIPS of variants (1)(2)(3) are 0.206, 0.226, and 0.209. They are still worse than the full model (0.188).

\subsubsection{Choice of activation function:}
We replaced SNAKE activation function in our MLP with Sine function (following SIREN), and the LPIPS in NRTDB is 0.194 (3\% worse).
This is because the periodic activation function in SIREN helps in handling fine details but not necessarily complex periodic patterns.

\begin{table}[ !h]
	\begin{center}
		\scalebox{0.81}{
	 \begin{tabular}{ccccccccc}	\toprule
	 \multirow{ 2}{*}{Category} &
	  \multirow{ 2}{*}{Method}  & \multicolumn{2}{c}{Only Unknown Regions} &  \multicolumn{5}{c}{Full Images} \\
\cmidrule(lr){3-4} \cmidrule(lr){5-9}  
	 &	   &  RMSE $\downarrow$  & FID $\downarrow$  & LPIPS $\downarrow$  & SSIM $\uparrow$  & PSNR$\uparrow$ & RMSE $\downarrow$  & FID $\downarrow$  \\
	 	\midrule 
\multicolumn{1}{c|}{ \multirow{8}{*}{\shortstack[l]{NPP-Net\\ Variants}}}	 	 &	No Periodicity   & 8.890  & 223.7 & 0.134 & 0.861 & 23.34 & \textbf{4.526} &  47.85\\  
\multicolumn{1}{c|}{}&	 	Pixel Only   & 8.703 & 153.5  & 0.180 & 0.859 & 23.65 & 5.756  & 47.84 \\  
\multicolumn{1}{c|}{}&	 	Patch Only   &  9.270 & 128.1 &    0.313 & 0.426 & 15.98 & 8.698  & 98.41 \\  
\multicolumn{1}{c|}{}&	 	Pixel + Random & 8.241 &100.0  & 0.104 & 0.902 & 24.52 & 4.770 &   26.06 \\  
\multicolumn{1}{c|}{}&	 	Initial Pipeline   &  8.246 & 81.99   &  0.124 & 0.893 & 24.29 & 5.160 &  27.46  \\  
\multicolumn{1}{c|}{}&	 	Top1 + Offsets   & 8.228 & 74.83 & 0.109 & 0.900 & 24.48 & 4.856 &  23.51\\  
\multicolumn{1}{c|}{}&	 	Top5 + Offsets   &  8.266 & 74.83& \textbf{0.100} & 0.900 & 24.55 & 4.721 &  23.08\\  
\multicolumn{1}{c|}{}&	 	Top3 w/o Offsets   &  8.259 & 78.20  & 0.119 & 0.892 & 24.31 & 5.038 & 26.01\\  
	 		 	
	 	\midrule
\multicolumn{1}{c|}{NPP-Net}	& 	Top3 + Offsets   & \textbf{8.164} &  \textbf{70.43}  & \textbf{0.100} & \textbf{0.908} & \textbf{24.71} & 4.839 &  \textbf{20.46} \\   
	 	\bottomrule
	 	\vspace{-2mm}
	 \end{tabular}}
	 	 		\caption{ \footnotesize
	 	 	Comparing different variants of NPP-Net for NPP completion in NRTDB. NPP-Net outperforms all other variants. Note that \enquote{No Periodicity} variant overfits the known regions in the full images case. \normalsize}
	 	\label{tab:ablation_nrtdb}
	\end{center}
	\end{table}

\vspace{-2mm}
\begin{table}[ !h]
	\begin{center}
		\scalebox{0.81}{
	 \begin{tabular}{ccccccccc}	\toprule
	 \multirow{ 2}{*}{Category} &
	  \multirow{ 2}{*}{Method}  & \multicolumn{2}{c}{Only Unknown Regions} &  \multicolumn{5}{c}{Full Images} \\
\cmidrule(lr){3-4} \cmidrule(lr){5-9}  
	 &	   &  RMSE $\downarrow$  & FID $\downarrow$  & LPIPS $\downarrow$  & SSIM $\uparrow$  & PSNR$\uparrow$ & RMSE $\downarrow$  & FID $\downarrow$  \\
	 	\midrule 
\multicolumn{1}{c|}{ \multirow{8}{*}{\shortstack[l]{NPP-Net\\ Variants}}}	 	 &	
No Periodicity   & 8.758  & 186.3 & 0.256 & 0.661 & 19.35 & \textbf{6.082} &  86.97\\  
\multicolumn{1}{c|}{}&	 	Pixel Only   & 8.400 & 145.7  & 0.263 & 0.685 & 20.43 & 6.545  & 81.80 \\  
\multicolumn{1}{c|}{}&	 	Patch Only   &  9.097 & 122.4 &    0.382 & 0.335 & 13.62 & 9.020  &  115.0 \\  
\multicolumn{1}{c|}{}&	 	Pixel + Random & 7.981 & 89.26  &  0.173 & 0.742 & 20.85 & 6.202 &   54.09 \\  
\multicolumn{1}{c|}{}&	 	Initial Pipeline   &  8.100 & 96.64   &  0.212 & 0.704 & 20.00 & 6.516 & 65.91\\  
\multicolumn{1}{c|}{}&	 	Top1 + Offsets   &  8.061 & 93.93 & 0.201 & 0.714 & 20.42 & 6.431 &  61.02\\  
\multicolumn{1}{c|}{}&	 	Top5 + Offsets   &  8.024 & 86.97 & 0.171 & 0.730 & 20.85 & 6.264 &  50.97\\  
\multicolumn{1}{c|}{}&	 	Top3 w/o Offsets   &  8.079 & 90.48  & 0.193 & 0.718 & 20.15 & 6.407 & 57.70\\  
	 		 	
	 	\midrule
\multicolumn{1}{c|}{NPP-Net}	& 	Top3 + Offsets   & \textbf{7.918} &  \textbf{85.39}  & \textbf{0.162} & \textbf{0.744} & \textbf{21.15} & 6.161 &  \textbf{50.39} \\   
	 	\bottomrule
	 	\vspace{-2mm}
	 \end{tabular}}
	 	 		\caption{ \footnotesize
	 	 	Comparing different variants of NPP-Net for NPP completion in DTD. NPP-Net outperforms all other variants. 
	 	 	\normalsize}
	 	\label{tab:ablation_dtd}
	\end{center}
	\end{table}

\vspace{-8mm}
\begin{table}[ !h]
	\begin{center}
		\scalebox{0.81}{
	 \begin{tabular}{ccccccccc}	\toprule
	 \multirow{ 2}{*}{Category} &
	  \multirow{ 2}{*}{Method}  & \multicolumn{2}{c}{Only Unknown Regions} &  \multicolumn{5}{c}{Full Images} \\
\cmidrule(lr){3-4} \cmidrule(lr){5-9}  
	 &	   &  RMSE $\downarrow$  & FID $\downarrow$  & LPIPS $\downarrow$  & SSIM $\uparrow$  & PSNR$\uparrow$ & RMSE $\downarrow$  & FID $\downarrow$  \\
	 	\midrule 
\multicolumn{1}{c|}{ \multirow{8}{*}{\shortstack[l]{NPP-Net\\ Variants}}}	 	 &	
No Periodicity   & 9.597  & 264.3 &  \textbf{0.040} & \textbf{0.961} & \textbf{27.20} & \textbf{4.646} &  \textbf{16.13}\\  
\multicolumn{1}{c|}{}&	 	Pixel Only   & 9.902 & 281.6  & 0.137 & 0.914 & 24.80 & 7.196  & 53.69 \\  
\multicolumn{1}{c|}{}&	 	Patch Only   &  9.987 & 156.6 &    0.535 & 0.166 & 11.04 & 9.993  &  234.9 \\  
\multicolumn{1}{c|}{}&	 	Pixel + Random & 9.540 & 132.7  &   0.073 & 0.939 & 25.43 & 6.289 &   18.15 \\  
\multicolumn{1}{c|}{}&	 	Initial Pipeline   &   \textbf{9.399} & 102.1   &  0.092 & 0.932 & 24.70 & 6.607 & 18.65\\  
\multicolumn{1}{c|}{}&	 	Top1 + Offsets   &   9.472 &  101.5 & 0.083 & 0.933 & 24.70 & 6.463 & 18.37\\  
\multicolumn{1}{c|}{}&	 	Top5 + Offsets   &  9.474 & 92.34 & 0.067 & 0.936 & 25.55 & 6.149 &  17.14\\  
\multicolumn{1}{c|}{}&	 	Top3 w/o Offsets   &  9.403 & 96.38 & 0.085 & 0.935 & 24.96 & 6.475 & 17.68\\  
	 		 	
	 	\midrule
\multicolumn{1}{c|}{NPP-Net}	& 	Top3 + Offsets   & 9.464 &  \textbf{91.05}  & 0.063 & 0.944 & 25.51 & 6.155 &  16.94 \\   
	 	\bottomrule
	 	\vspace{-2mm}
	 \end{tabular}}
	 	 		\caption{ \footnotesize
	 	 	Comparing different variants of NPP-Net for NPP completion in Facade dataset. NPP-Net outperforms all other variants.  Besides, the Facade dataset contains a number of non-NPP images, resulting in a different performance from the other two datasets. \enquote{No Periodicity} variant overfits the known regions in the full images case.	 	 	 \normalsize}
	 	\label{tab:ablation_facade}
	\end{center}
	\end{table}

\clearpage

\subsection{Comparison with Baselines}

We show the quantitative results with FID and RMSE (evaluated only in unknown regions) and all metrics (evaluated in the full images) in Table \ref{tab:baseline_nrtdb}, Table \ref{tab:baseline_dtd}, and Table \ref{tab:baseline_facade}. 
For NRTDB and DTD datasets, NPP-Net outperforms all the baselines. 
For RMSE in Table \ref{tab:baseline_nrtdb}, DIP and Siren perform better in the full image even if they generate bad results in unknown regions because they simply overfit the known ones.
For FID in Table \ref{tab:baseline_dtd}, image quilting performs better. One possible reason is that the images in this dataset have smaller variations in the motifs thus simply tiling them may produce reasonable results. 

The Facade dataset contains some images that are not strictly NPP images. Thus the performance is different from the other two datasets. But NPP-Net can still outperform other baselines (except for Lama that trained on large datasets) by optimizing only on a single image. Similarly, DIP and Siren perform better in the full image since they simply overfit the known ones.

\begin{figure*}[h]
\begin{center}
\begin{tabular}[t]{c}
 \\[.3cm]
 \rotatebox[origin=c]{0}{
 \shortstack[c]{\scriptsize Input   \\ \;}
 }  \\[.8cm]
 \rotatebox[origin=c]{0}{
 \shortstack[c]{\scriptsize Image   \\\scriptsize Quilting}
 }\\[1.1cm]
 \rotatebox[origin=c]{0}{
 \shortstack[c]{ \scriptsize PatchMatch \\ \;} 
 }\\[1.1cm]
 \rotatebox[origin=c]{0}{
 \shortstack[c]{\scriptsize DIP \\ \;} 
 }\\[1.3cm]
  \rotatebox[origin=c]{0}{
  \shortstack[c]{\scriptsize Siren \\ \;}
 }\\[1.2cm]
  \rotatebox[origin=c]{0}{
 \shortstack[c]{\scriptsize PEN-Net \\ \;}  }\\[1.1cm]
 \rotatebox[origin=c]{0}{
 \shortstack[c]{\scriptsize ProFill \\ \;}  }\\[1.3cm]
  \rotatebox[origin=c]{0}{
 \shortstack[c]{\scriptsize Lama \\ \;}  }\\[1.cm]
 \rotatebox[origin=c]{0}{
 \shortstack[c]{\scriptsize Huang \\\scriptsize \etal }  }\\[1.cm]
 \rotatebox[origin=c]{0}{
 \shortstack[c]{\scriptsize BPI \\ \;}   }\\[1.2cm]
 \rotatebox[origin=c]{0}{
 \shortstack[c]{\scriptsize NPP-Net \\ \;}   }\\[1.1cm]
 \rotatebox[origin=c]{0}{
 \shortstack[c]{\scriptsize Ground \\   \scriptsize Truth}}  
\end{tabular}
\includegraphics[width=0.83\linewidth,valign=t]{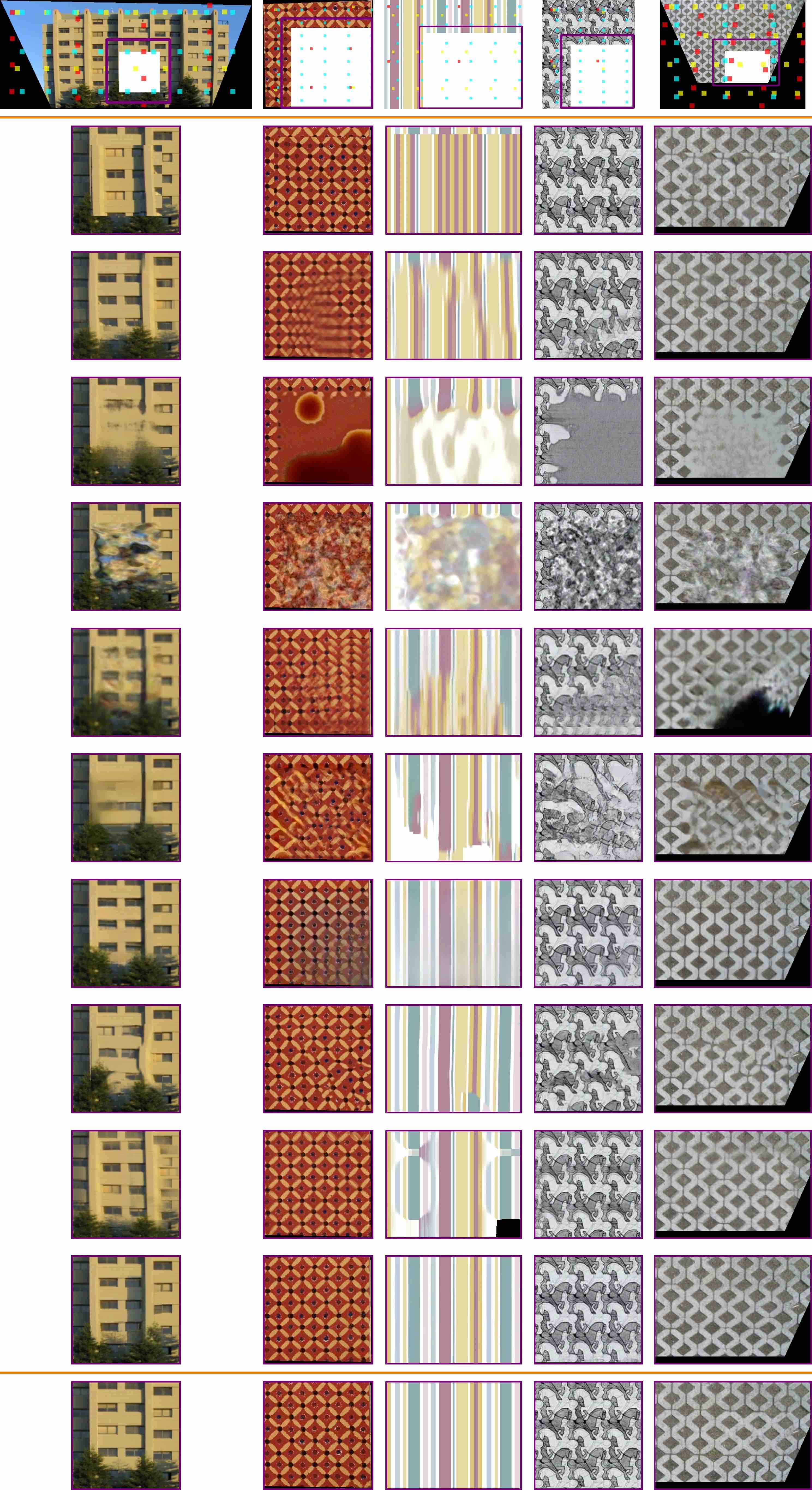}
\end{center} 
    \vspace{-4mm}
    \caption{ Comparison with other baselines for image completion. 
    }
    \label{completion_baseline_1}
 \end{figure*}

\begin{figure*}[h]
\begin{center}
\begin{tabular}[t]{c}
 \\[.3cm]
 \rotatebox[origin=c]{0}{
 \shortstack[c]{\scriptsize Input   \\ \;}
 }  \\[.8cm]
 \rotatebox[origin=c]{0}{
 \shortstack[c]{\scriptsize Image   \\\scriptsize Quilting}
 }\\[1.1cm]
 \rotatebox[origin=c]{0}{
 \shortstack[c]{ \scriptsize PatchMatch \\ \;} 
 }\\[1.1cm]
 \rotatebox[origin=c]{0}{
 \shortstack[c]{\scriptsize DIP \\ \;} 
 }\\[1.3cm]
  \rotatebox[origin=c]{0}{
  \shortstack[c]{\scriptsize Siren \\ \;}
 }\\[1.2cm]
  \rotatebox[origin=c]{0}{
 \shortstack[c]{\scriptsize PEN-Net \\ \;}  }\\[1.1cm]
 \rotatebox[origin=c]{0}{
 \shortstack[c]{\scriptsize ProFill \\ \;}  }\\[1.3cm]
  \rotatebox[origin=c]{0}{
 \shortstack[c]{\scriptsize Lama \\ \;}  }\\[1.cm]
 \rotatebox[origin=c]{0}{
 \shortstack[c]{\scriptsize Huang \\\scriptsize \etal }  }\\[1.cm]
 \rotatebox[origin=c]{0}{
 \shortstack[c]{\scriptsize BPI \\ \;}   }\\[1.2cm]
 \rotatebox[origin=c]{0}{
 \shortstack[c]{\scriptsize NPP-Net \\ \;}   }\\[1.1cm]
 \rotatebox[origin=c]{0}{
 \shortstack[c]{\scriptsize Ground \\   \scriptsize Truth}}  
\end{tabular}
\includegraphics[width=0.77\linewidth,valign=t]{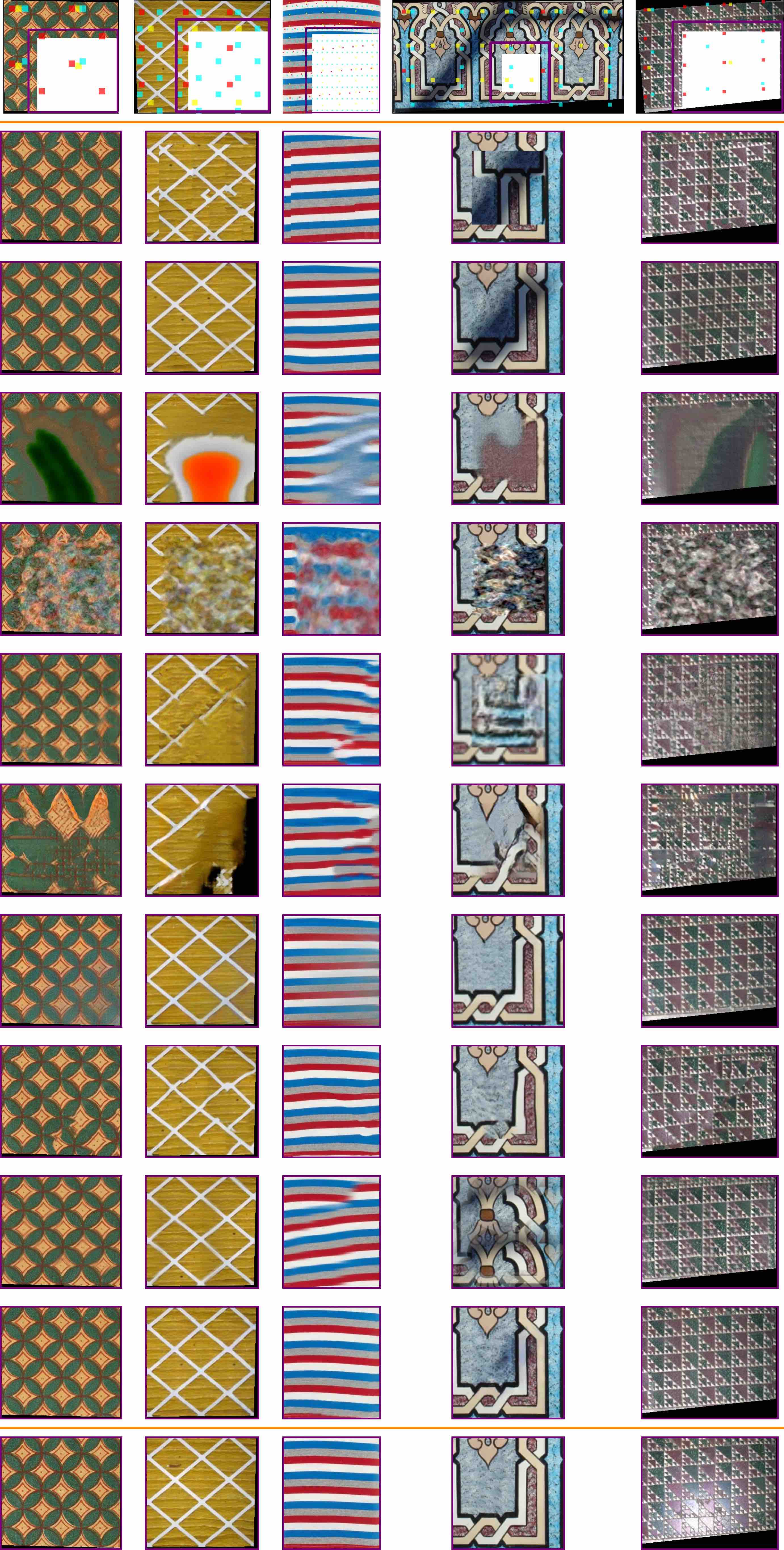}
\end{center} 
    \vspace{-4mm}
    \caption{ Comparison with other baselines for image completion. 
Note that periods in the second column are not scaled by 2.     }
    \label{completion_baseline_2}
 \end{figure*}

\begin{figure*}[h]
\begin{center}
\begin{tabular}[t]{c}
 \\[.3cm]
 \rotatebox[origin=c]{0}{
 \shortstack[c]{\scriptsize Input   \\ \;}
 }  \\[.8cm]
 \rotatebox[origin=c]{0}{
 \shortstack[c]{\scriptsize Image   \\\scriptsize Quilting}
 }\\[1.1cm]
 \rotatebox[origin=c]{0}{
 \shortstack[c]{ \scriptsize PatchMatch \\ \;} 
 }\\[1.1cm]
 \rotatebox[origin=c]{0}{
 \shortstack[c]{\scriptsize DIP \\ \;} 
 }\\[1.cm]
  \rotatebox[origin=c]{0}{
  \shortstack[c]{\scriptsize Siren \\ \;}
 }\\[1.2cm]
  \rotatebox[origin=c]{0}{
 \shortstack[c]{\scriptsize PEN-Net \\ \;}  }\\[1.1cm]
 \rotatebox[origin=c]{0}{
 \shortstack[c]{\scriptsize ProFill \\ \;}  }\\[1.1cm]
  \rotatebox[origin=c]{0}{
 \shortstack[c]{\scriptsize Lama \\ \;}  }\\[1.cm]
 \rotatebox[origin=c]{0}{
 \shortstack[c]{\scriptsize Huang \\\scriptsize \etal }  }\\[1.cm]
 \rotatebox[origin=c]{0}{
 \shortstack[c]{\scriptsize BPI \\ \;}   }\\[1.2cm]
 \rotatebox[origin=c]{0}{
 \shortstack[c]{\scriptsize NPP-Net \\ \;}   }\\[.9cm]
 \rotatebox[origin=c]{0}{
 \shortstack[c]{\scriptsize Ground \\   \scriptsize Truth}}  
\end{tabular}
\includegraphics[width=0.84\linewidth,valign=t]{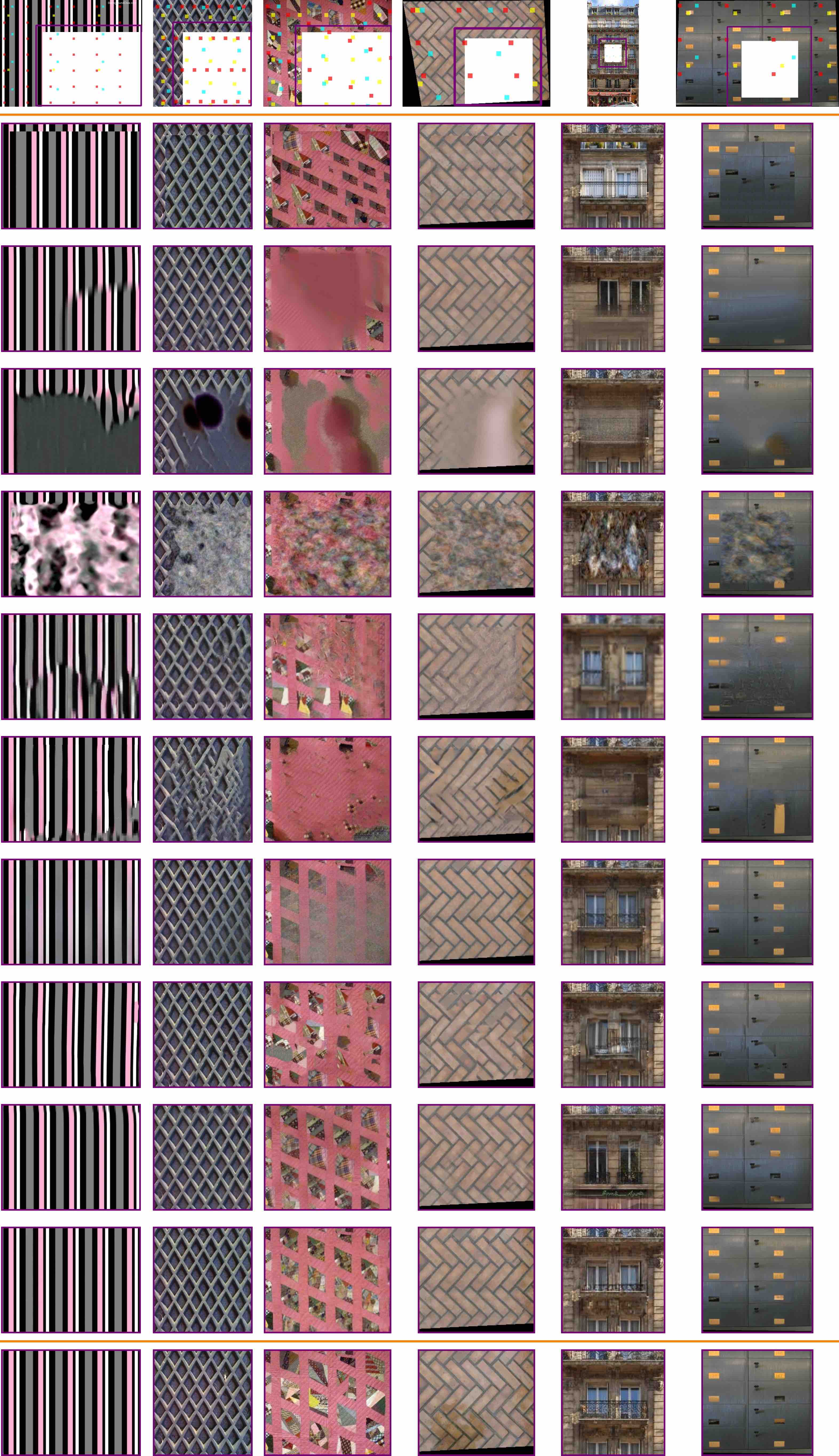}
\end{center} 
    \vspace{-4mm}
    \caption{ Comparison with other baselines for image completion. Note that periods in the fifth column are not scaled by 2. 
    }
    \label{completion_baseline_3}
 \end{figure*}

\begin{figure*}[h]
\begin{center}
\begin{tabular}[t]{c}
 \\[.3cm]
 \rotatebox[origin=c]{0}{
 \shortstack[c]{\scriptsize Input   \\ \;}
 }  \\[.8cm]
 \rotatebox[origin=c]{0}{
 \shortstack[c]{\scriptsize Image   \\\scriptsize Quilting}
 }\\[1.2cm]
 \rotatebox[origin=c]{0}{
 \shortstack[c]{ \scriptsize PatchMatch \\ \;} 
 }\\[1.2cm]
 \rotatebox[origin=c]{0}{
 \shortstack[c]{\scriptsize DIP \\ \;} 
 }\\[1.2cm]
  \rotatebox[origin=c]{0}{
  \shortstack[c]{\scriptsize Siren \\ \;}
 }\\[1.2cm]
  \rotatebox[origin=c]{0}{
 \shortstack[c]{\scriptsize PEN-Net \\ \;}  }\\[1.1cm]
 \rotatebox[origin=c]{0}{
 \shortstack[c]{\scriptsize ProFill \\ \;}  }\\[1.3cm]
  \rotatebox[origin=c]{0}{
 \shortstack[c]{\scriptsize Lama \\ \;}  }\\[1.1cm]
 \rotatebox[origin=c]{0}{
 \shortstack[c]{\scriptsize Huang \\\scriptsize \etal }  }\\[1.2cm]
 \rotatebox[origin=c]{0}{
 \shortstack[c]{\scriptsize BPI \\ \;}   }\\[1.2cm]
 \rotatebox[origin=c]{0}{
 \shortstack[c]{\scriptsize NPP-Net \\ \;}   }\\[1.1cm]
 \rotatebox[origin=c]{0}{
 \shortstack[c]{\scriptsize Ground \\   \scriptsize Truth}}  
\end{tabular}
\includegraphics[width=0.76\linewidth,valign=t]{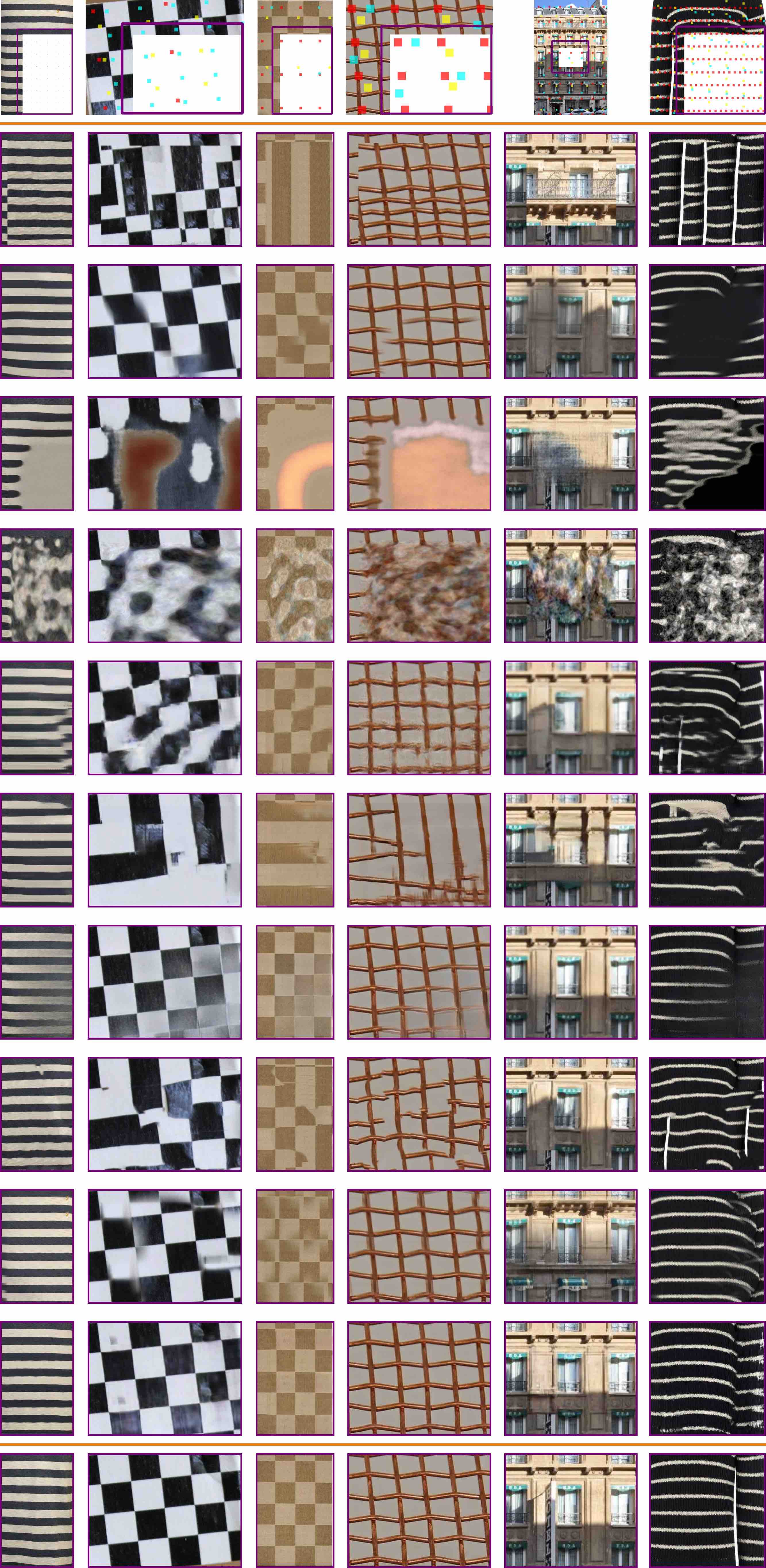}
\end{center} 
    \vspace{-4mm}
    \caption{ Comparison with other baselines for image completion. 
    }
    \label{completion_baseline_4}
 \end{figure*}

\begin{figure*}[h]
\begin{center}
\begin{tabular}[t]{c}
 \\[.3cm]
 \rotatebox[origin=c]{0}{
 \shortstack[c]{\scriptsize Input   \\ \;}
 }  \\[.8cm]
 \rotatebox[origin=c]{0}{
 \shortstack[c]{\scriptsize Image   \\\scriptsize Quilting}
 }\\[1.1cm]
 \rotatebox[origin=c]{0}{
 \shortstack[c]{ \scriptsize PatchMatch \\ \;} 
 }\\[1.1cm]
 \rotatebox[origin=c]{0}{
 \shortstack[c]{\scriptsize DIP \\ \;} 
 }\\[1.cm]
  \rotatebox[origin=c]{0}{
  \shortstack[c]{\scriptsize Siren \\ \;}
 }\\[1.2cm]
  \rotatebox[origin=c]{0}{
 \shortstack[c]{\scriptsize PEN-Net \\ \;}  }\\[1.1cm]
 \rotatebox[origin=c]{0}{
 \shortstack[c]{\scriptsize ProFill \\ \;}  }\\[1.cm]
  \rotatebox[origin=c]{0}{
 \shortstack[c]{\scriptsize Lama \\ \;}  }\\[1.cm]
 \rotatebox[origin=c]{0}{
 \shortstack[c]{\scriptsize Huang \\\scriptsize \etal }  }\\[1.cm]
 \rotatebox[origin=c]{0}{
 \shortstack[c]{\scriptsize BPI \\ \;}   }\\[1.1cm]
 \rotatebox[origin=c]{0}{
 \shortstack[c]{\scriptsize NPP-Net \\ \;}   }\\[.9cm]
 \rotatebox[origin=c]{0}{
 \shortstack[c]{\scriptsize Ground \\   \scriptsize Truth}}  
\end{tabular}
\includegraphics[width=0.76\linewidth,valign=t]{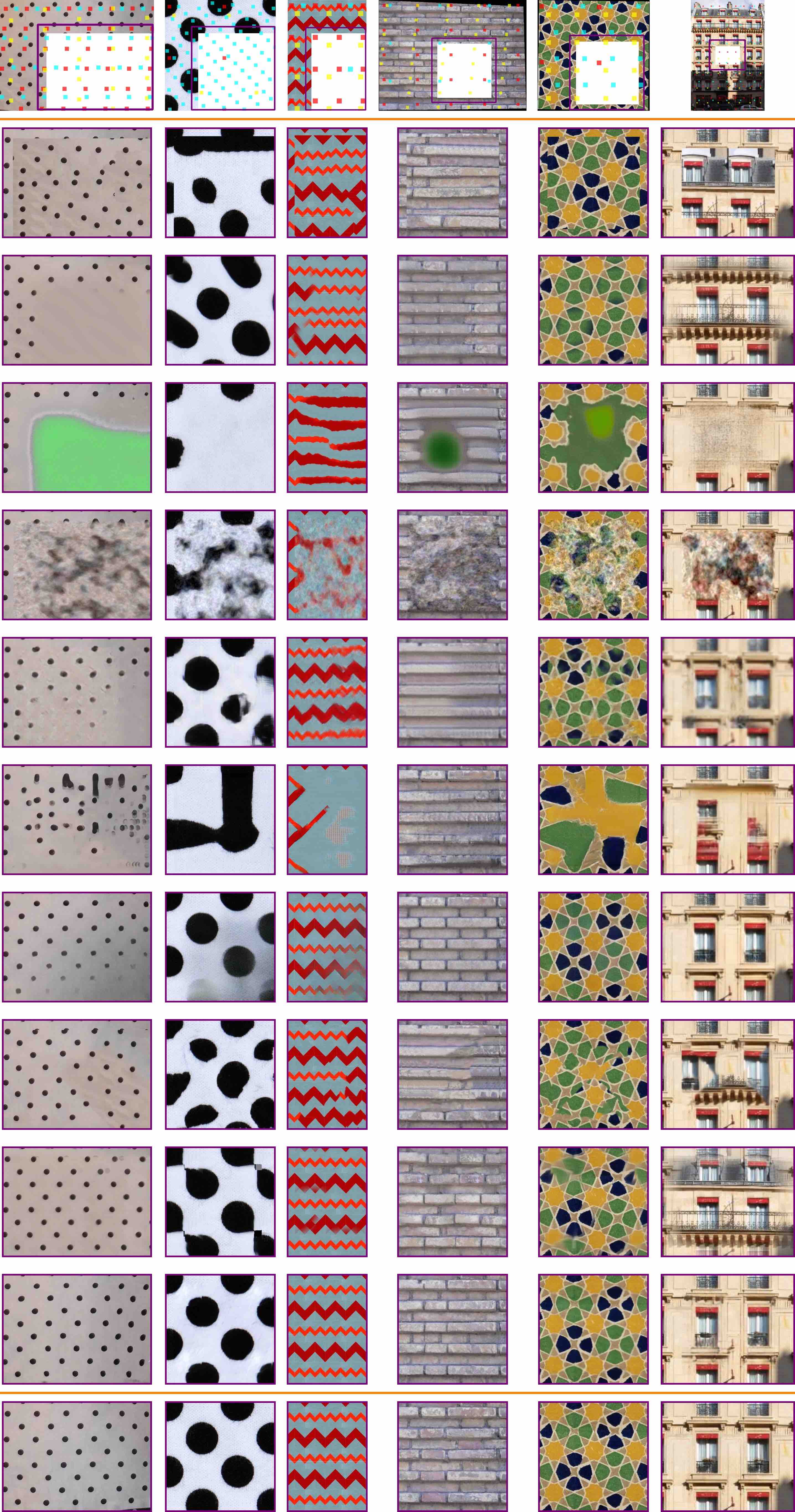}
\end{center} 
    \vspace{-4mm}
    \caption{ Comparison with other baselines for image completion.  Note that periods in the fifth column are not scaled by 2. 
    }
    \label{completion_baseline_5}
 \end{figure*}

\begin{figure*}[h]
\begin{center}
\begin{tabular}[t]{c}
 \\[.3cm]
 \rotatebox[origin=c]{0}{
 \shortstack[c]{\scriptsize Input   \\ \;}
 }  \\[.8cm]
 \rotatebox[origin=c]{0}{
 \shortstack[c]{\scriptsize Image   \\\scriptsize Quilting}
 }\\[1.1cm]
 \rotatebox[origin=c]{0}{
 \shortstack[c]{ \scriptsize PatchMatch \\ \;} 
 }\\[1.1cm]
 \rotatebox[origin=c]{0}{
 \shortstack[c]{\scriptsize DIP \\ \;} 
 }\\[1.3cm]
  \rotatebox[origin=c]{0}{
  \shortstack[c]{\scriptsize Siren \\ \;}
 }\\[1.2cm]
  \rotatebox[origin=c]{0}{
 \shortstack[c]{\scriptsize PEN-Net \\ \;}  }\\[1.2cm]
 \rotatebox[origin=c]{0}{
 \shortstack[c]{\scriptsize ProFill \\ \;}  }\\[1.3cm]
  \rotatebox[origin=c]{0}{
 \shortstack[c]{\scriptsize Lama \\ \;}  }\\[1.cm]
 \rotatebox[origin=c]{0}{
 \shortstack[c]{\scriptsize Huang \\\scriptsize \etal }  }\\[1.cm]
 \rotatebox[origin=c]{0}{
 \shortstack[c]{\scriptsize BPI \\ \;}   }\\[1.3cm]
 \rotatebox[origin=c]{0}{
 \shortstack[c]{\scriptsize NPP-Net \\ \;}   }\\[1.1cm]
 \rotatebox[origin=c]{0}{
 \shortstack[c]{\scriptsize Ground \\   \scriptsize Truth}}  
\end{tabular}
\includegraphics[width=0.67\linewidth,valign=t]{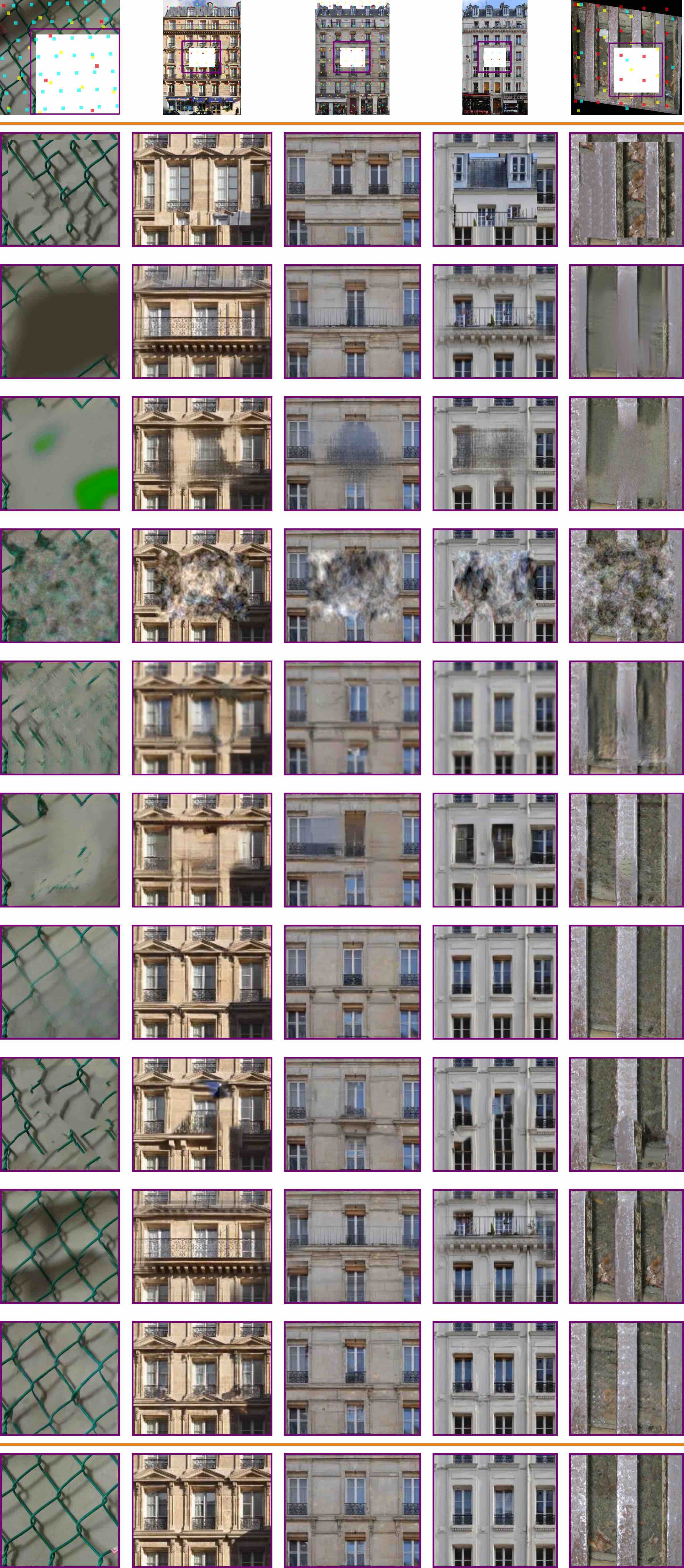}
\end{center} 
    \vspace{-4mm}
    \caption{ Comparison with other baselines for image completion. 
    }
    \label{completion_baseline_6}
 \end{figure*}

\begin{figure*}[h]
\begin{center}
\begin{tabular}[t]{c}
 \\[.3cm]
 \rotatebox[origin=c]{0}{
 \shortstack[c]{\scriptsize Input   \\ \;}
 }  \\[.8cm]
 \rotatebox[origin=c]{0}{
 \shortstack[c]{\scriptsize Image   \\\scriptsize Quilting}
 }\\[1.1cm]
 \rotatebox[origin=c]{0}{
 \shortstack[c]{ \scriptsize PatchMatch \\ \;} 
 }\\[1.1cm]
 \rotatebox[origin=c]{0}{
 \shortstack[c]{\scriptsize DIP \\ \;} 
 }\\[1.3cm]
  \rotatebox[origin=c]{0}{
  \shortstack[c]{\scriptsize Siren \\ \;}
 }\\[1.2cm]
  \rotatebox[origin=c]{0}{
 \shortstack[c]{\scriptsize PEN-Net \\ \;}  }\\[1.1cm]
 \rotatebox[origin=c]{0}{
 \shortstack[c]{\scriptsize ProFill \\ \;}  }\\[1.3cm]
  \rotatebox[origin=c]{0}{
 \shortstack[c]{\scriptsize Lama \\ \;}  }\\[1.cm]
 \rotatebox[origin=c]{0}{
 \shortstack[c]{\scriptsize Huang \\\scriptsize \etal }  }\\[1.cm]
 \rotatebox[origin=c]{0}{
 \shortstack[c]{\scriptsize BPI \\ \;}   }\\[1.2cm]
 \rotatebox[origin=c]{0}{
 \shortstack[c]{\scriptsize NPP-Net \\ \;}   }\\[1.1cm]
 \rotatebox[origin=c]{0}{
 \shortstack[c]{\scriptsize Ground \\   \scriptsize Truth}}  
\end{tabular}
\includegraphics[width=0.66\linewidth,valign=t]{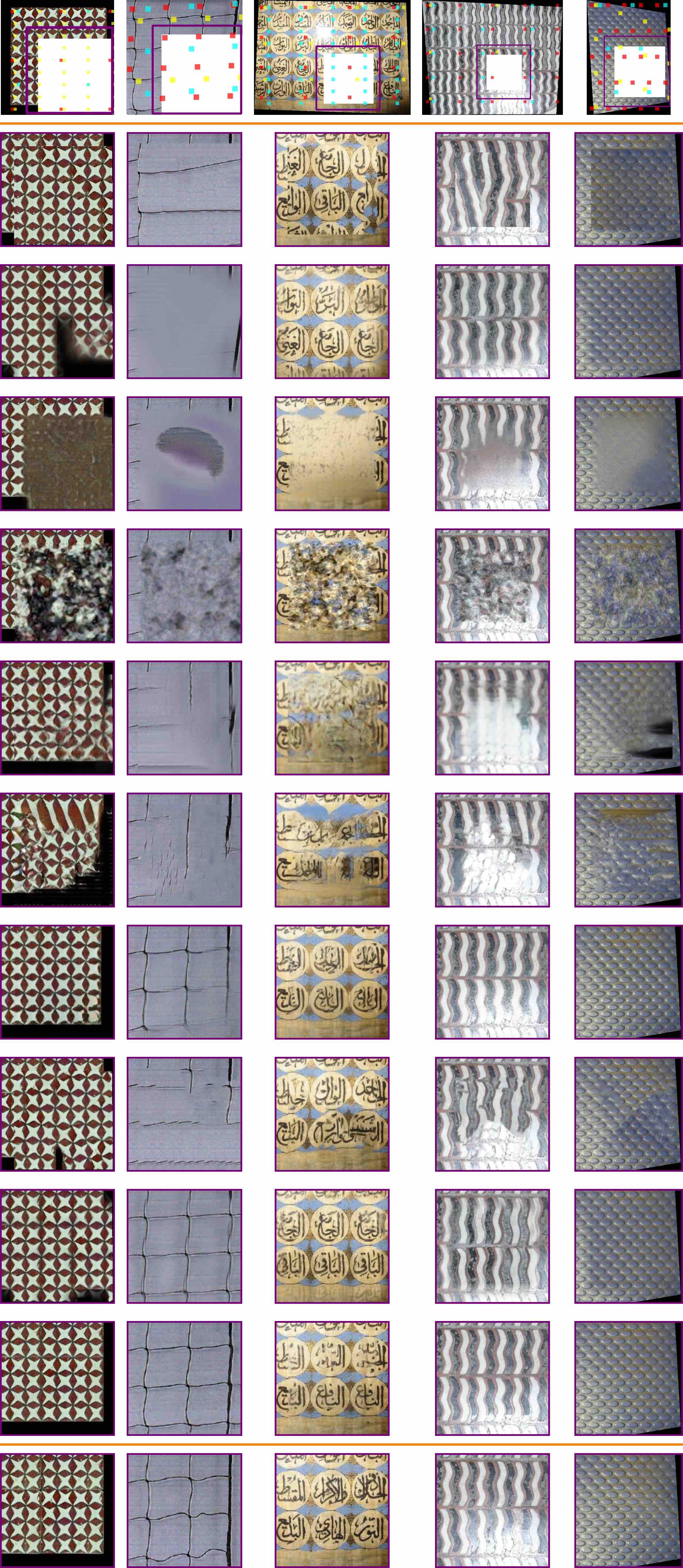}
\end{center} 
    \vspace{-4mm}
    \caption{ Comparison with other baselines for image completion. 
    }
    \label{completion_baseline_7}
 \end{figure*}

\begin{figure*}[h]
\begin{center}
\begin{tabular}[t]{c}
 \\[.3cm]
 \rotatebox[origin=c]{0}{
 \shortstack[c]{\scriptsize Input   \\ \;}
 }  \\[.8cm]
 \rotatebox[origin=c]{0}{
 \shortstack[c]{\scriptsize Image   \\\scriptsize Quilting}
 }\\[1.1cm]
 \rotatebox[origin=c]{0}{
 \shortstack[c]{ \scriptsize PatchMatch \\ \;} 
 }\\[1.1cm]
 \rotatebox[origin=c]{0}{
 \shortstack[c]{\scriptsize DIP \\ \;} 
 }\\[1.3cm]
  \rotatebox[origin=c]{0}{
  \shortstack[c]{\scriptsize Siren \\ \;}
 }\\[1.2cm]
  \rotatebox[origin=c]{0}{
 \shortstack[c]{\scriptsize PEN-Net \\ \;}  }\\[1.1cm]
 \rotatebox[origin=c]{0}{
 \shortstack[c]{\scriptsize ProFill \\ \;}  }\\[1.3cm]
  \rotatebox[origin=c]{0}{
 \shortstack[c]{\scriptsize Lama \\ \;}  }\\[1.cm]
 \rotatebox[origin=c]{0}{
 \shortstack[c]{\scriptsize Huang \\\scriptsize \etal }  }\\[1.cm]
 \rotatebox[origin=c]{0}{
 \shortstack[c]{\scriptsize BPI \\ \;}   }\\[1.2cm]
 \rotatebox[origin=c]{0}{
 \shortstack[c]{\scriptsize NPP-Net \\ \;}   }\\[1.1cm]
 \rotatebox[origin=c]{0}{
 \shortstack[c]{\scriptsize Ground \\   \scriptsize Truth}}  
\end{tabular}
\includegraphics[width=0.68\linewidth,valign=t]{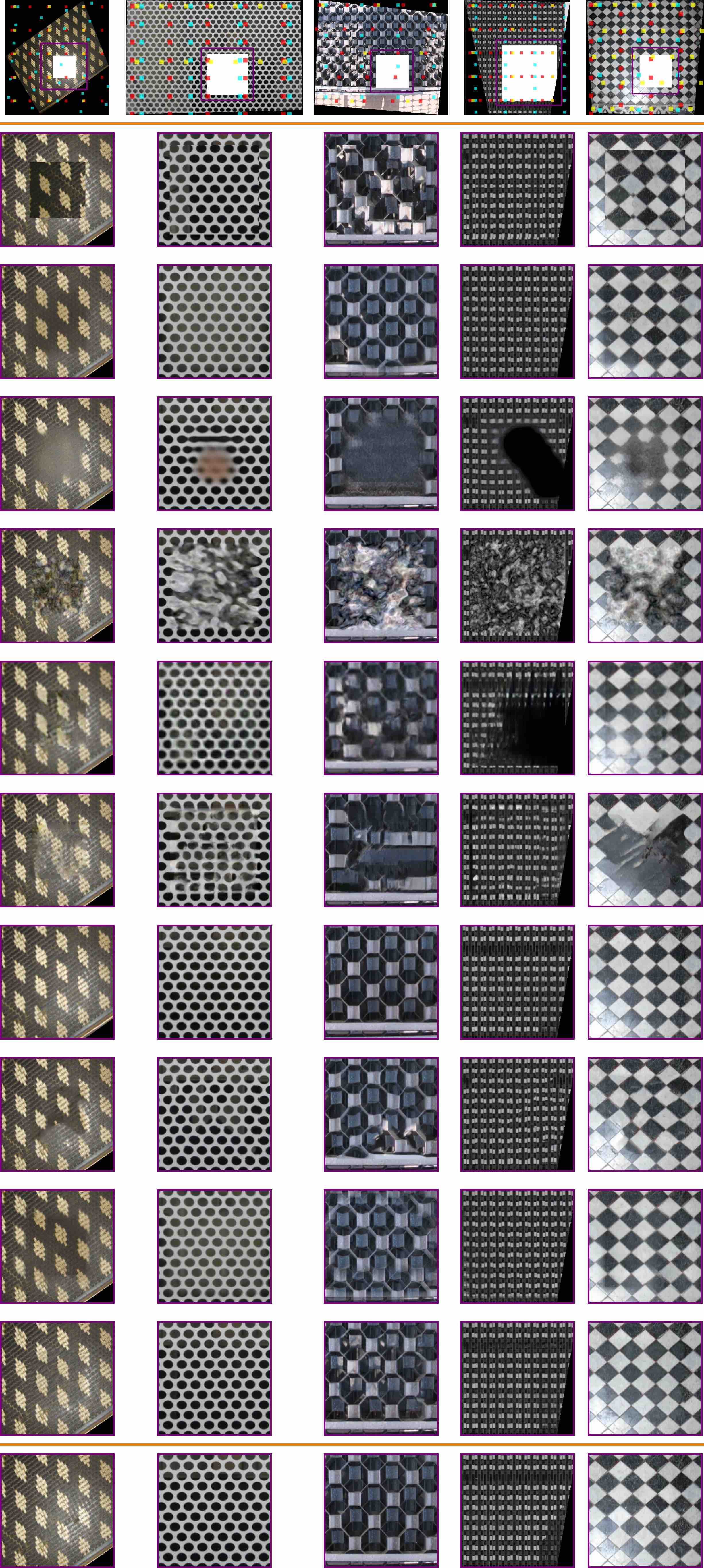}
\end{center} 
    \vspace{-4mm}
    \caption{ Comparison with other baselines for image completion. 
    }
    \label{completion_baseline_8}
 \end{figure*}

We also show more qualitative results with all baselines. Figure \ref{completion_baseline_1} to Figure \ref{completion_baseline_6} show that NPP-Net outperforms baselines in terms of global consistency. Figure \ref{completion_baseline_7} and Figure \ref{completion_baseline_8} are for local variations, including boundaries and lighting effects. 
As shown in the results, NPP-Net can handle the NPP images that only have periodic patterns along one direction. 
This is because NPP-Net tends to focus on this direction while ignoring the second one since this minimizes the loss in the known regions better.
Note that pretrained PEN-Net only accepts images with a fixed resolution,  thus we resize the input image to satisfy the constraint, which may lead to blurry results. 
Lama also performs well since it implicitly learns scene prior, such as periodicity, from large datasets, while NPP-Net is only optimized on a single image.

\begin{figure*}[h]
\begin{center}
\begin{tabular}[t]{c}
 \\[.3cm]
 \rotatebox[origin=c]{0}{
 \shortstack[c]{\scriptsize Input   \\ \;}
 }  \\[.9cm]
 \rotatebox[origin=c]{0}{
 \shortstack[c]{\scriptsize Image   \\\scriptsize Quilting}
 }\\[1.2cm]
 \rotatebox[origin=c]{0}{
 \shortstack[c]{ \scriptsize PatchMatch \\ \;} 
 }\\[1.1cm]
 \rotatebox[origin=c]{0}{
 \shortstack[c]{\scriptsize DIP \\ \;} 
 }\\[1.3cm]
  \rotatebox[origin=c]{0}{
  \shortstack[c]{\scriptsize Siren \\ \;}
 }\\[1.2cm]
  \rotatebox[origin=c]{0}{
 \shortstack[c]{\scriptsize PEN-Net \\ \;}  }\\[1.1cm]
 \rotatebox[origin=c]{0}{
 \shortstack[c]{\scriptsize ProFill \\ \;}  }\\[1.2cm]
  \rotatebox[origin=c]{0}{
 \shortstack[c]{\scriptsize Lama \\ \;}  }\\[.9cm]
 \rotatebox[origin=c]{0}{
 \shortstack[c]{\scriptsize Huang \\\scriptsize \etal }  }\\[1.2cm]
 \rotatebox[origin=c]{0}{
 \shortstack[c]{\scriptsize BPI \\ \;}   }\\[1.1cm]
 \rotatebox[origin=c]{0}{
 \shortstack[c]{\scriptsize NPP-Net \\ \;}   }\\[1.1cm]
 \rotatebox[origin=c]{0}{
 \shortstack[c]{\scriptsize Ground \\   \scriptsize Truth}}  
\end{tabular}
\includegraphics[width=0.77\linewidth,valign=t]{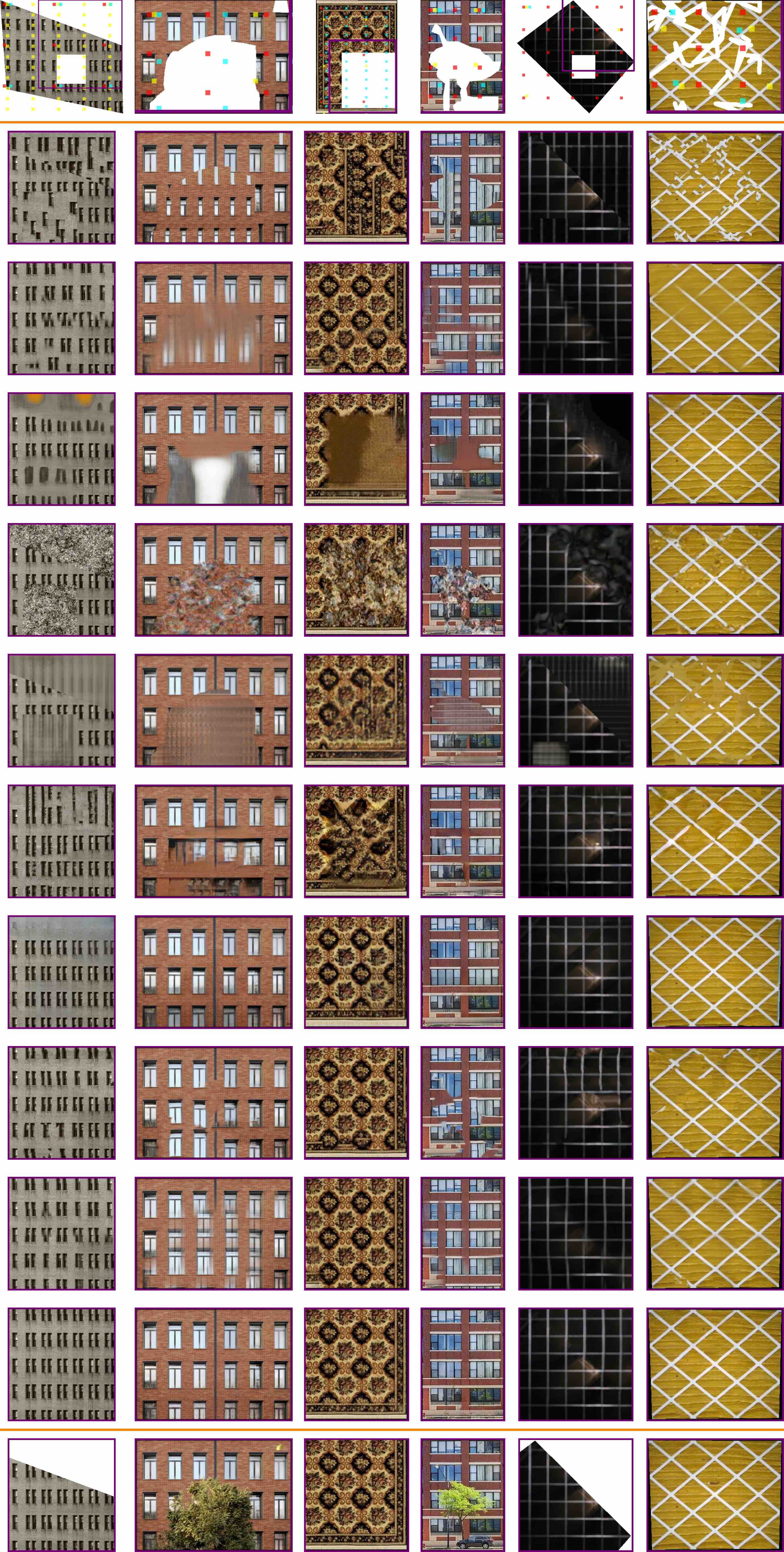}
\end{center} 
    \vspace{-4mm}
    \caption{ Comparison with other baselines for image completion. Note that periods in the last column are not scaled by 2. 
    }
    \label{completion_baseline_11}
 \end{figure*}

\begin{figure*}[h]
\begin{center}
\begin{tabular}[t]{c}
 \\[.3cm]
 \rotatebox[origin=c]{0}{
 \shortstack[c]{\scriptsize Input   \\ \;}
 }  \\[.8cm]
 \rotatebox[origin=c]{0}{
 \shortstack[c]{\scriptsize Image   \\\scriptsize Quilting}
 }\\[1.1cm]
 \rotatebox[origin=c]{0}{
 \shortstack[c]{ \scriptsize PatchMatch \\ \;} 
 }\\[1.1cm]
 \rotatebox[origin=c]{0}{
 \shortstack[c]{\scriptsize DIP \\ \;} 
 }\\[1.2cm]
  \rotatebox[origin=c]{0}{
  \shortstack[c]{\scriptsize Siren \\ \;}
 }\\[1.2cm]
  \rotatebox[origin=c]{0}{
 \shortstack[c]{\scriptsize PEN-Net \\ \;}  }\\[1.1cm]
 \rotatebox[origin=c]{0}{
 \shortstack[c]{\scriptsize ProFill \\ \;}  }\\[1.1cm]
  \rotatebox[origin=c]{0}{
 \shortstack[c]{\scriptsize Lama \\ \;}  }\\[1.cm]
 \rotatebox[origin=c]{0}{
 \shortstack[c]{\scriptsize Huang \\\scriptsize \etal }  }\\[1.cm]
 \rotatebox[origin=c]{0}{
 \shortstack[c]{\scriptsize BPI \\ \;}   }\\[1.cm]
 \rotatebox[origin=c]{0}{
 \shortstack[c]{\scriptsize NPP-Net \\ \;}   }\\[1.1cm]
 \rotatebox[origin=c]{0}{
 \shortstack[c]{\scriptsize Ground \\   \scriptsize Truth}}  
\end{tabular}
\includegraphics[width=0.84\linewidth,valign=t]{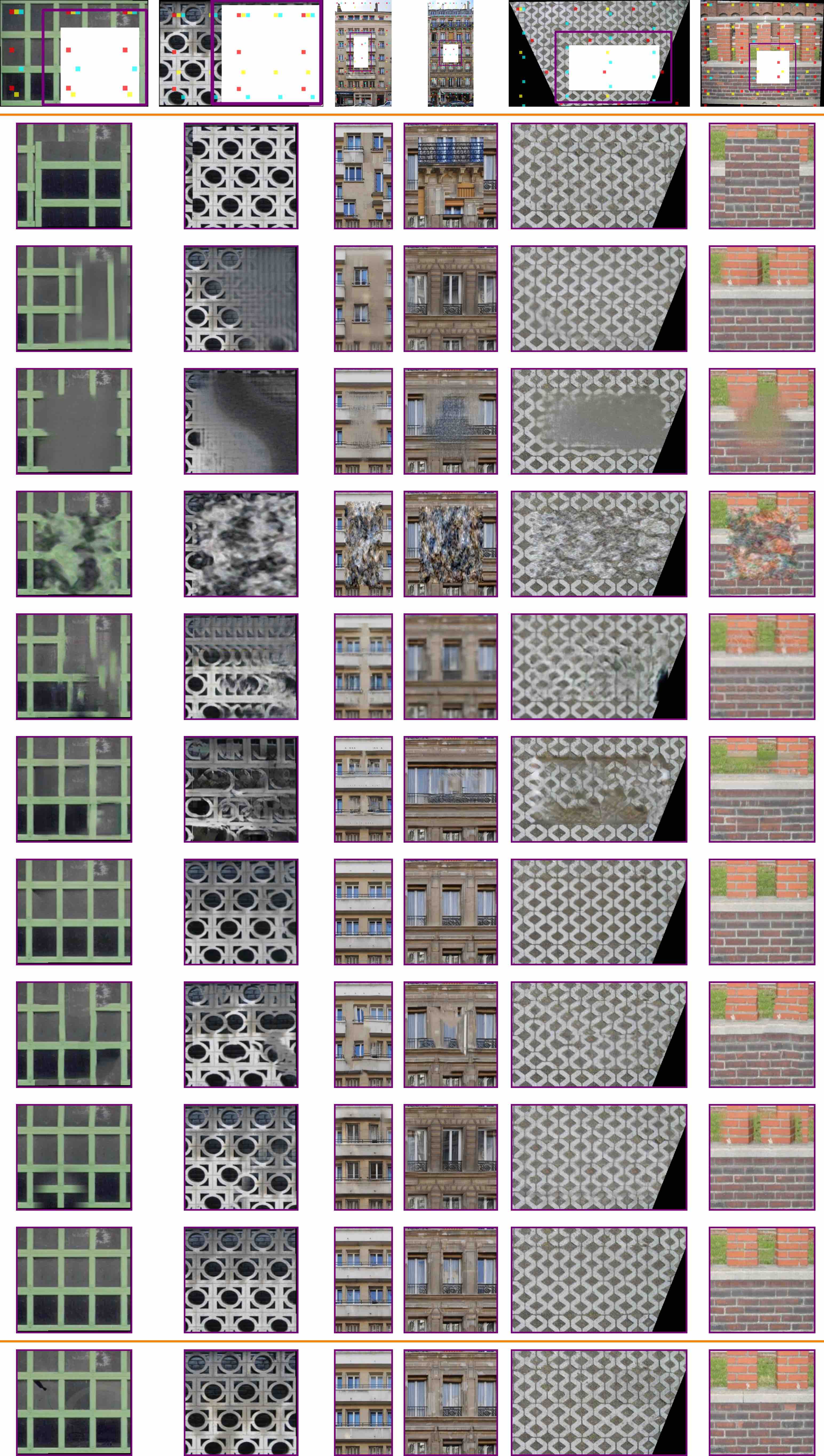}
\end{center} 
    \vspace{-4mm}
    \caption{ Comparison with other baselines for image completion. 
    }
    \label{completion_baseline_10}
 \end{figure*}

In addition, We show the results in the main paper with all the baselines in Figure \ref{completion_baseline_11} and Figure \ref{completion_baseline_10}. Besides, Figure \ref{completion_baseline_11} also shows more examples for extrapolation and different mask shapes. 
The results show NPP-Net outperforms baselines (especially for those single image-based counterparts) in terms of global consistency and local variations with various shapes and sizes of masks. 

Please note that, although we minimizes LPIPS in known regions during training, we provides evaluation of all methods only in unknown regions (Table 1 in main paper). 
Even removing LPIPS loss, NPP-Net still obtains 0.204 LPIPS (24\% better than the best baseline (BPI) that does not optimize LPIPS). In addition, NPP-Net performs the best in SSIM and FID (main paper and supp), which are not explicitly minimized by all methods.

\begin{table}[ !h]
	\begin{center}
		\scalebox{0.85}{
	 \begin{tabular}{ccccccccccc}	\toprule
	 \multirow{ 2}{*}{Category} &
	  \multirow{ 2}{*}{Method}  & \multicolumn{2}{c}{Only Unknown Regions} &  \multicolumn{5}{c}{Full Images} \\
\cmidrule(lr){3-4} \cmidrule(lr){5-9}  
	 &	   &  RMSE $\downarrow$  & FID $\downarrow$  & LPIPS $\downarrow$  & SSIM $\uparrow$  & PSNR$\uparrow$ & RMSE $\downarrow$  & FID $\downarrow$  \\
	 	\midrule 
  \multicolumn{1}{c|}{\multirow{3}{*}{\shortstack[l]{Large\\ Datasets}}}	 &	PEN-Net~\cite{yan2019PENnet}   & 9.231  & 204.1 & - & - & - & - & - \\
\multicolumn{1}{c|}{}	 &	ProFill~\cite{zeng2020high}   & 9.137  & 187.4 & - & - & - & - & -  \\
\multicolumn{1}{c|}{}	&    Lama~\cite{suvorov2022resolution}   & {\underline{8.567}}  & {{86.18}} & - & - & - & - & - \\  
	    	 	\midrule 
\multicolumn{1}{c|}{ \multirow{6}{*}{\shortstack[l]{Single\\ Image}} }	 &	 	Image Quilting~\cite{image_quilting} &9.739  & 89.53 & - & - & - & - & - \\ 
\multicolumn{1}{c|}{}	 &	PatchMatch~\cite{barnes2009patchmatch}  &   8.956  & 86.67 & - & - & - & - & - \\
\multicolumn{1}{c|}{}	& 	DIP~\cite{ulyanov2018deep}  & 8.906  & 242.3 & \underline{0.177} & \underline{0.829} & \underline{22.32} & \underline{4.702}  & \underline{79.65}\\
\multicolumn{1}{c|}{}&	 	Siren~\cite{sitzmann2020implicit} & 9.859  &  303.4 & 0.183 & 0.791 & 21.17 & \textbf{4.418} & 85.18 \\
\multicolumn{1}{c|}{}	& 	Huang \etal~\cite{huang2014image}   & 8.921  & 104.7 & - & - & - & -& -  \\  
\multicolumn{1}{c|}{}	& 	BPI~\cite{li2020multi}    & 9.213  & \underline{71.77} & - & - & - & -& -  \\  
 \cmidrule(lr){2-9} 
\multicolumn{1}{c|}{}	& 	NPP-Net   &  \textbf{8.164} &  \textbf{70.43}  & \textbf{0.100} & \textbf{0.908} & \textbf{24.71} & 4.839 &  \textbf{50.39} \\   
	 	\bottomrule
	 	\vspace{-2mm}
	 \end{tabular}}
	 	 		\caption{ \footnotesize
	 	 	Comparing with different baselines for NPP completion in NRTDB dataset. NPP-Net outperforms all baselines.  
	 	 	 \normalsize}
	 	\label{tab:baseline_nrtdb}
	\end{center}
	\end{table}

\begin{table}[ !h]
	 	\vspace{-5mm}
	\begin{center}
		\scalebox{0.85}{
	 \begin{tabular}{ccccccccccc}	\toprule
	 \multirow{ 2}{*}{Category} &
	  \multirow{ 2}{*}{Method}  & \multicolumn{2}{c}{Only Unknown Regions} &  \multicolumn{5}{c}{Full Images} \\
\cmidrule(lr){3-4} \cmidrule(lr){5-9}  
	 &	   &  RMSE $\downarrow$  & FID $\downarrow$  & LPIPS $\downarrow$  & SSIM $\uparrow$  & PSNR$\uparrow$ & RMSE $\downarrow$  & FID $\downarrow$  \\
	 	\midrule 
  \multicolumn{1}{c|}{\multirow{3}{*}{\shortstack[l]{Large\\ Datasets}}}	 &	PEN-Net~\cite{yan2019PENnet}   & 8.952  & 166.5 & - & - & - & - & - \\
\multicolumn{1}{c|}{}	 &	ProFill~\cite{zeng2020high}   & 9.024  & 199.7 & - & - & - & - & -  \\
\multicolumn{1}{c|}{}	&    Lama~\cite{suvorov2022resolution}   & {9.137}  & {{93.14}} & - & - & - & - & - \\  
	    	 	\midrule 
\multicolumn{1}{c|}{ \multirow{6}{*}{\shortstack[l]{Single\\ Image}} }	 &	 	
Image Quilting~\cite{image_quilting} & 9.431  & \textbf{83.39} & - & - & - & - & - \\ 
\multicolumn{1}{c|}{}	 &	PatchMatch~\cite{barnes2009patchmatch}  &   8.637  &105.4 & - & - & - & - & - \\
\multicolumn{1}{c|}{}	& 	DIP~\cite{ulyanov2018deep}  & 9.350  &  263.32& \underline{0.344} & \underline{0.604} & \underline{16.23} & 6.549  & \underline{153.74}\\
\multicolumn{1}{c|}{}&	 	Siren~\cite{sitzmann2020implicit} & 9.895  &  314.1  & 0.393 & 0.544 & 16.21 & \underline{6.220} & 174.9 \\
\multicolumn{1}{c|}{}	& 	Huang \etal~\cite{huang2014image}   & \underline{8.520}  & 97.97& - & - & - & -& -  \\  
\multicolumn{1}{c|}{}	& 	BPI~\cite{li2020multi}    & 8.910  &  85.93 & - & - & - & -& -  \\  
	 		  \cmidrule(lr){2-9} 
\multicolumn{1}{c|}{}	& 	NPP-Net   & \textbf{7.918} &  \underline{85.39}  & \textbf{0.162} & \textbf{0.744} & \textbf{21.15} & \textbf{6.161} &  \textbf{50.39} \\   
	 	\bottomrule
	 	\vspace{-2mm}
	 \end{tabular}}
	 	 		\caption{ \footnotesize
	 	 	Comparing with different baselines for NPP completion in DTD dataset. NPP-Net outperforms all baselines.  
	 	 	 \normalsize}
	 	\label{tab:baseline_dtd}
	\end{center}
	\end{table}

\begin{table}[ !h]
	 	\vspace{-5mm}

	\begin{center}
		\scalebox{0.85}{
	 \begin{tabular}{ccccccccccc}	\toprule
	 \multirow{ 2}{*}{Category} &
	  \multirow{ 2}{*}{Method}  & \multicolumn{2}{c}{Only Unknown Regions} &  \multicolumn{5}{c}{Full Images} \\
\cmidrule(lr){3-4} \cmidrule(lr){5-9}  
	 &	   &  RMSE $\downarrow$  & FID $\downarrow$  & LPIPS $\downarrow$  & SSIM $\uparrow$  & PSNR$\uparrow$ & RMSE $\downarrow$  & FID $\downarrow$  \\
	 	\midrule 
  \multicolumn{1}{c|}{\multirow{3}{*}{\shortstack[l]{Large\\ Datasets}}}	 &	PEN-Net~\cite{yan2019PENnet}   & 9.745  & 133.3 & - & - & - & - & - \\
\multicolumn{1}{c|}{}	 &	ProFill~\cite{zeng2020high}   & 9.566  & 138.0 & - & - & - & - & -  \\
\multicolumn{1}{c|}{}	&    Lama~\cite{suvorov2022resolution}   & \textbf{9.434}  & \textbf{83.07} & - & - & - & - & - \\  
	    	 	\midrule 
\multicolumn{1}{c|}{ \multirow{6}{*}{\shortstack[l]{Single\\ Image}} }	 &	 	
Image Quilting~\cite{image_quilting} &  10.16  & 121.46 & - & - & - & - & - \\ 
\multicolumn{1}{c|}{}	 &	PatchMatch~\cite{barnes2009patchmatch}  &   9.741  &   \underline{95.34} & - & - & - & - & - \\
\multicolumn{1}{c|}{}	& 	DIP~\cite{ulyanov2018deep}  & 9.660  &   239.9 & \textbf{0.055} & \textbf{0.951} & \textbf{ 26.71} & \underline{5.080}  & \underline{20.98}\\
\multicolumn{1}{c|}{}&	 	Siren~\cite{sitzmann2020implicit} & 10.17  &   404.7  & \underline{0.060} & 0.938 & 24.39 & \textbf{2.730} & 33.54 \\
\multicolumn{1}{c|}{}	& 	Huang \etal~\cite{huang2014image}   & 9.447  &  104.9 & - & - & - & -& -  \\  
\multicolumn{1}{c|}{}	& 	BPI~\cite{li2020multi}    & 9.992 &   98.52 & - & - & - & -& -  \\  
	 	 \cmidrule(lr){2-9} 
\multicolumn{1}{c|}{}	& 	NPP-Net   & \underline{9.464} &  99.50  & 0.063 & \underline{0.944} & \underline{25.51} & 6.155 &  \textbf{16.94} \\   
	 	\bottomrule
	 	\vspace{-2mm}
	 \end{tabular}}
	 	 		\caption{ \footnotesize
	 	 	Comparing with different baselines for NPP completion in Facade dataset.  Note that, Facade dataset has different performance since it contains some non-NPP images. 
	 	 	 \normalsize}
	 	\label{tab:baseline_facade}
	\end{center}
	\end{table}

\clearpage
 \subsection{Influence of Mask Size}

\begin{figure}[!h]
\vspace{5mm}
    \captionsetup[subfigure]{labelformat=empty, font=small}
        \captionsetup[subfigure]{labelformat=empty, font=small}
\begin{subfigure}{.5\linewidth}
  \centerline{\includegraphics[width=\textwidth]{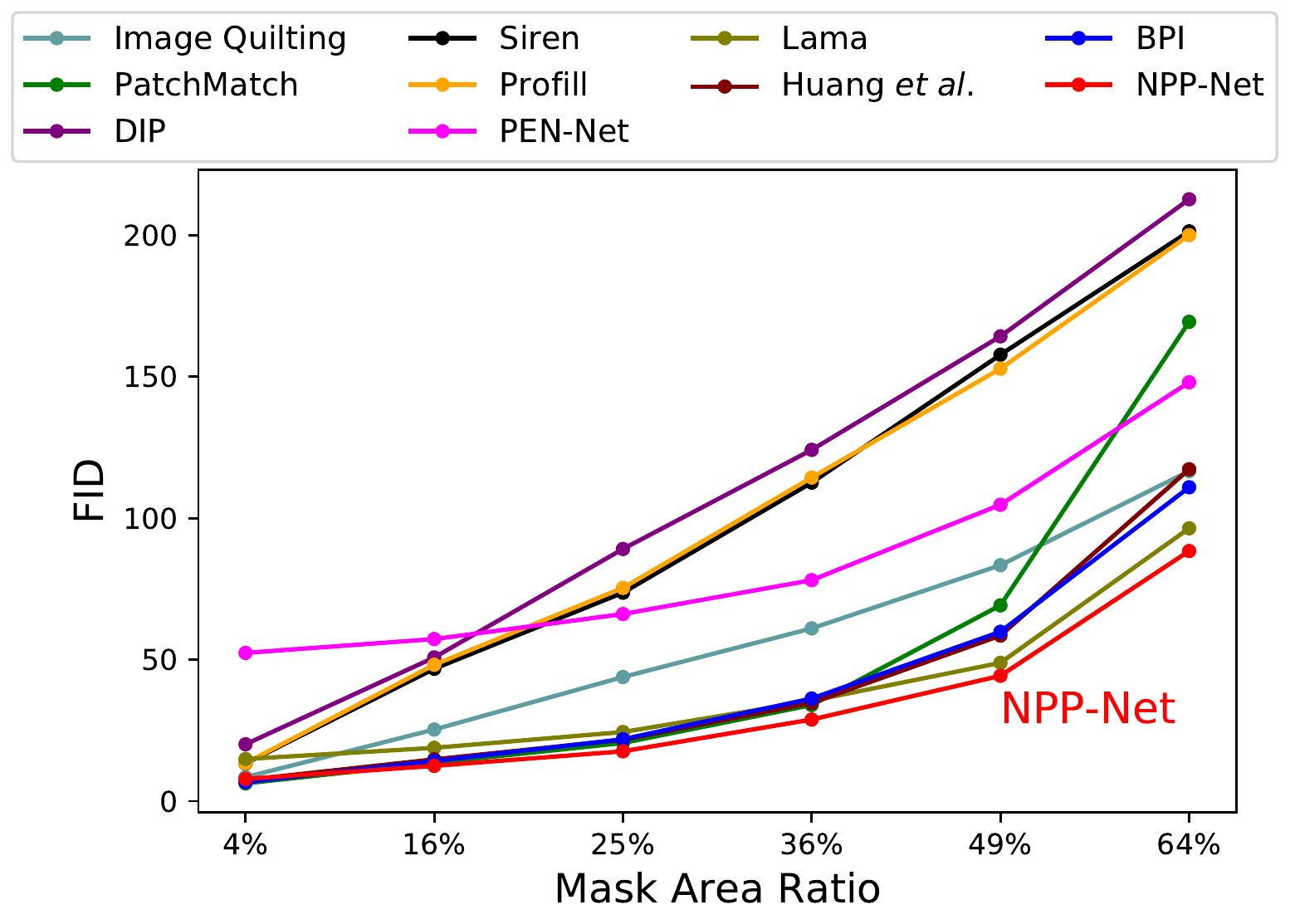}}
  \caption{\normalsize (a) FID \normalsize}
    \vspace{5mm}
\end{subfigure}
\hfill
\begin{subfigure}{.5\linewidth}
  \centerline{\includegraphics[width=\textwidth]{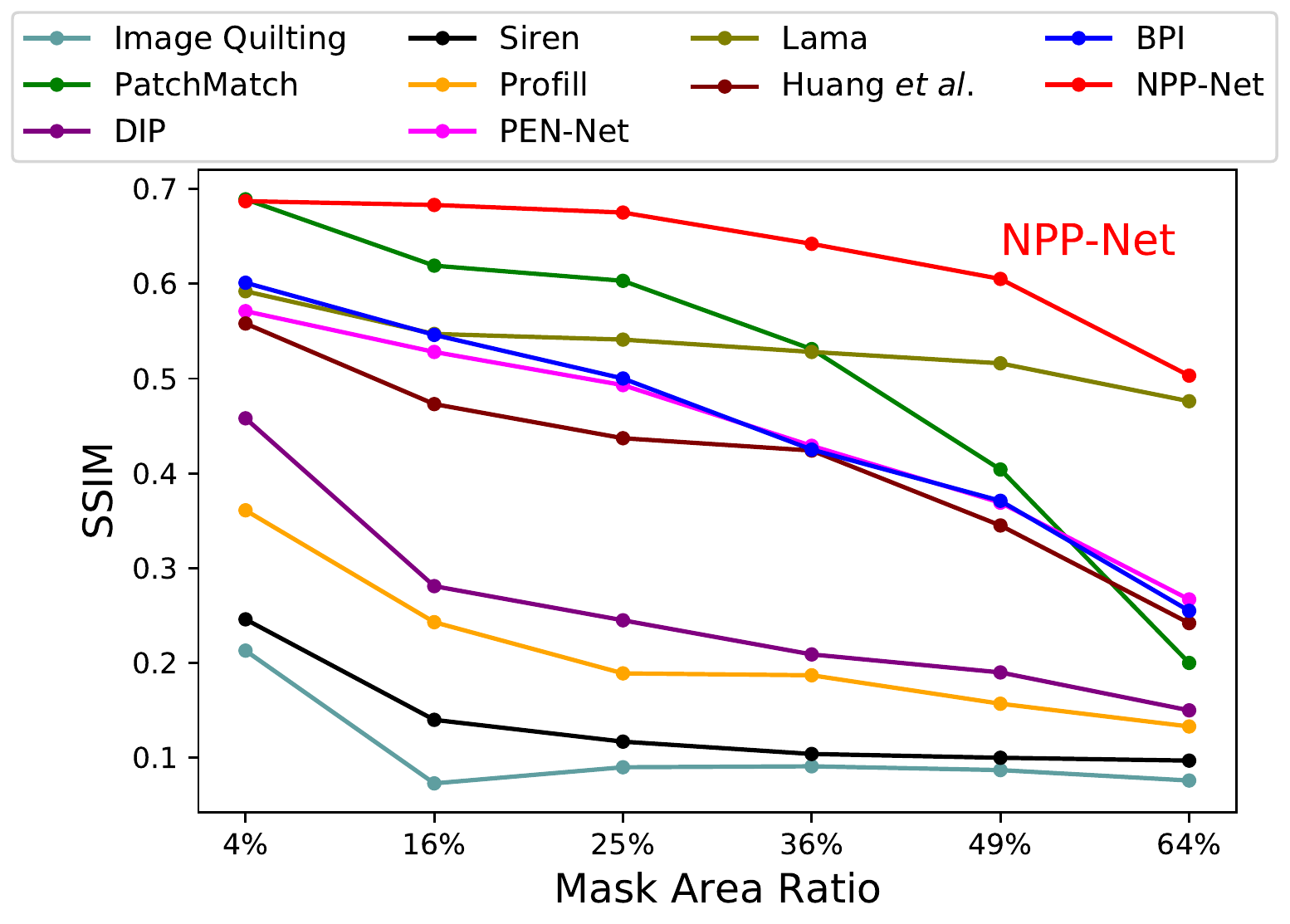}}
  \caption{\normalsize (b) SSIM \normalsize}
  \vspace{5mm}
\end{subfigure}
\\
\begin{subfigure}{.5\linewidth}
  \centerline{\includegraphics[width=\textwidth]{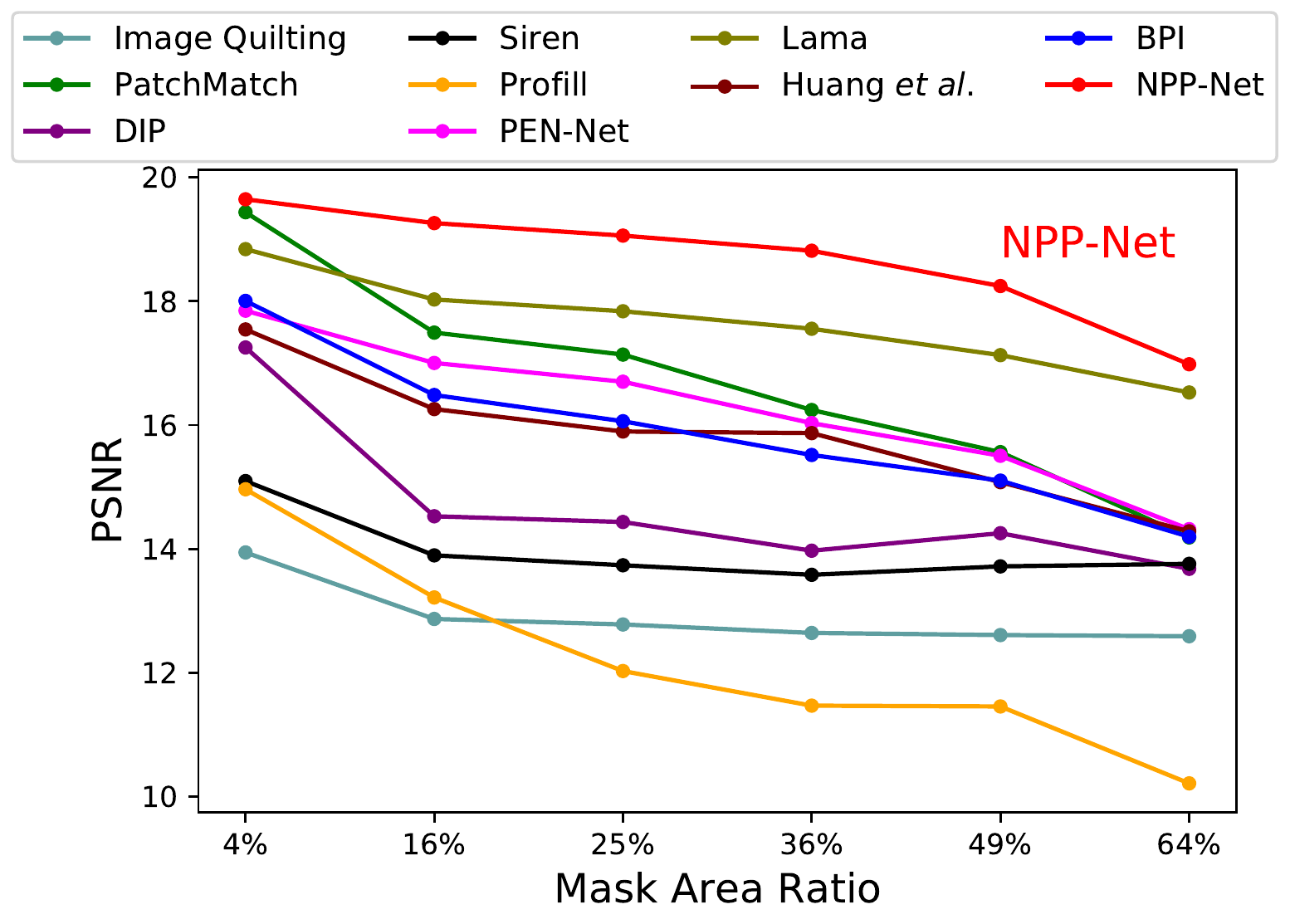}}
  \caption{\normalsize (c) PSNR \normalsize}
\end{subfigure}
\hfill
\begin{subfigure}{.5\linewidth}
  \centerline{\includegraphics[width=\textwidth]{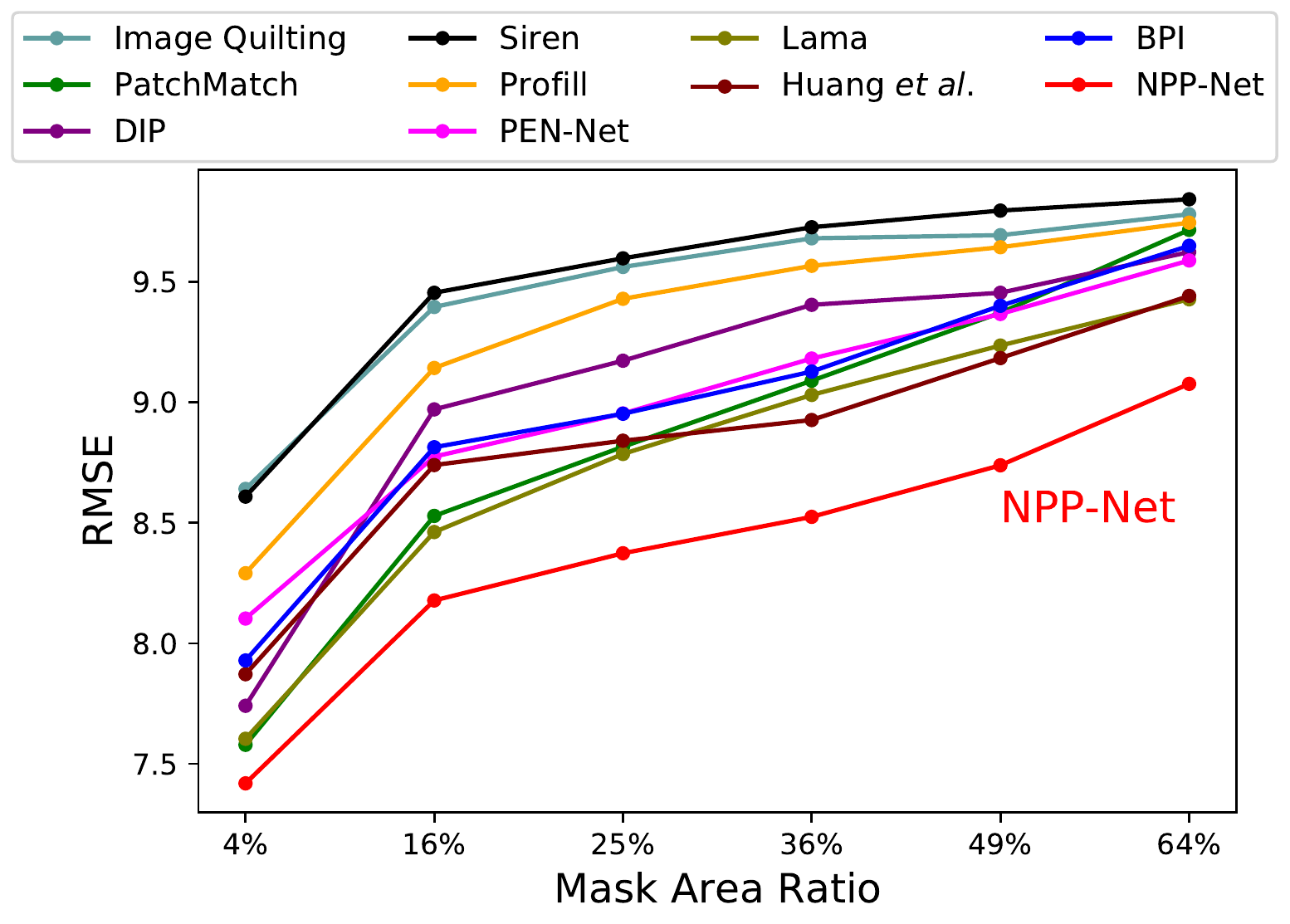}}
  \caption{\normalsize (d) RMSE   \normalsize}
\end{subfigure}
\caption{
Comparison of model performances for different mask sizes in the NRTDB dataset. FID is evaluated in the full image, and the other three metrics are tested in the unknown regions.  Note that, LPIPS result on NRTDB   has already been shown in the main paper. 
}
\vspace{-0.2in}
\label{mask_size_nrtdb}
\end{figure}

\begin{figure}[!h]
\vspace{5mm}
    \captionsetup[subfigure]{labelformat=empty, font=small}
    \begin{subfigure}{1.\linewidth}
  \centerline{\includegraphics[width=.5\textwidth]{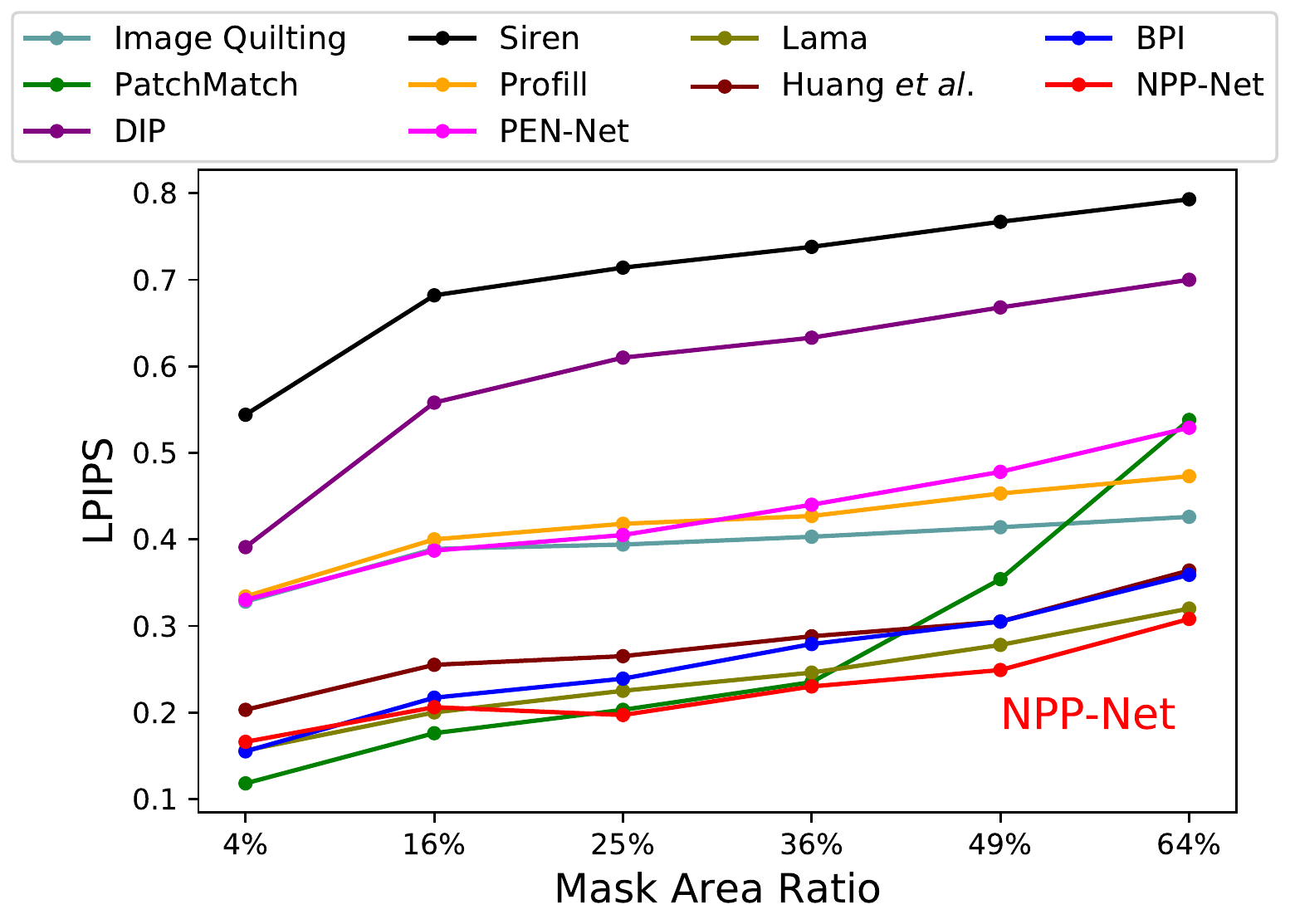}}
  \caption{\normalsize (a) LPIPS \normalsize}
    \vspace{5mm}
\end{subfigure}
\\
\begin{subfigure}{.5\linewidth}
  \centerline{\includegraphics[width=\textwidth]{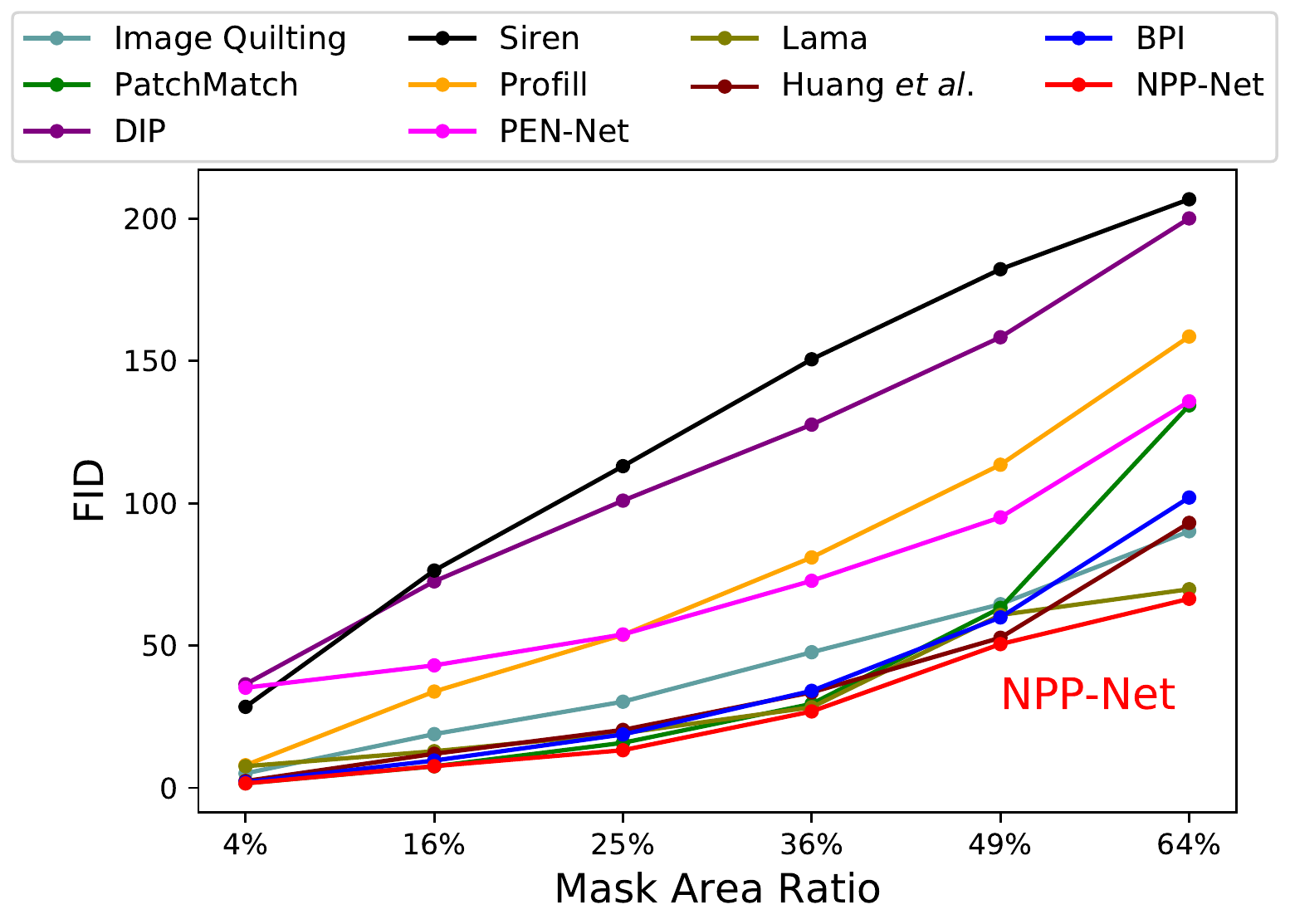}}
  \caption{\normalsize (b) FID \normalsize}
    \vspace{5mm}
\end{subfigure}
\hfill
\begin{subfigure}{.5\linewidth}
  \centerline{\includegraphics[width=\textwidth]{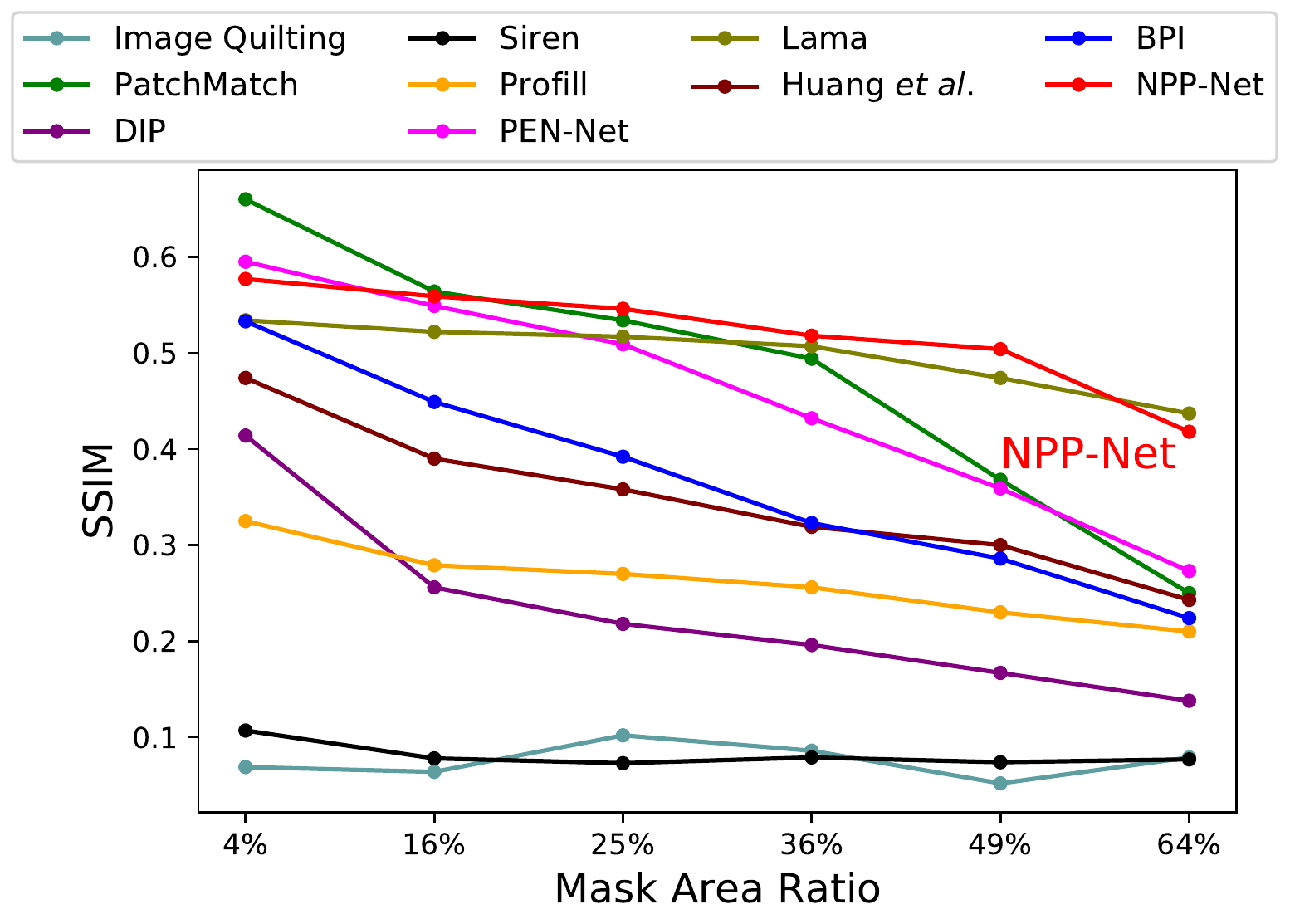}}
  \caption{\normalsize (c) SSIM \normalsize}
  \vspace{5mm}
\end{subfigure}
\\
\begin{subfigure}{.5\linewidth}
  \centerline{\includegraphics[width=\textwidth]{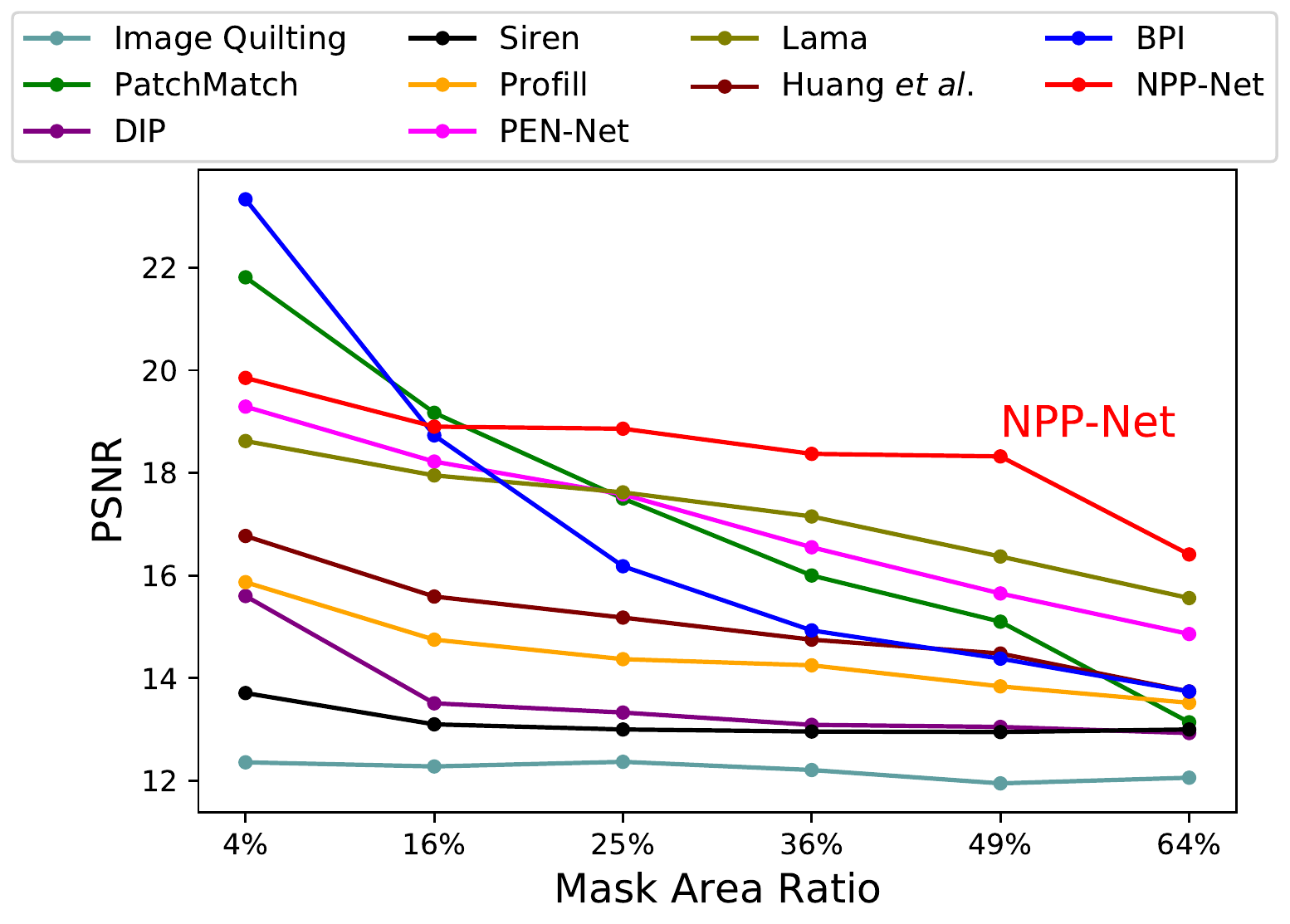}}
  \caption{\normalsize (d) PSNR \normalsize}
\end{subfigure}
\hfill
\begin{subfigure}{.5\linewidth}
  \centerline{\includegraphics[width=\textwidth]{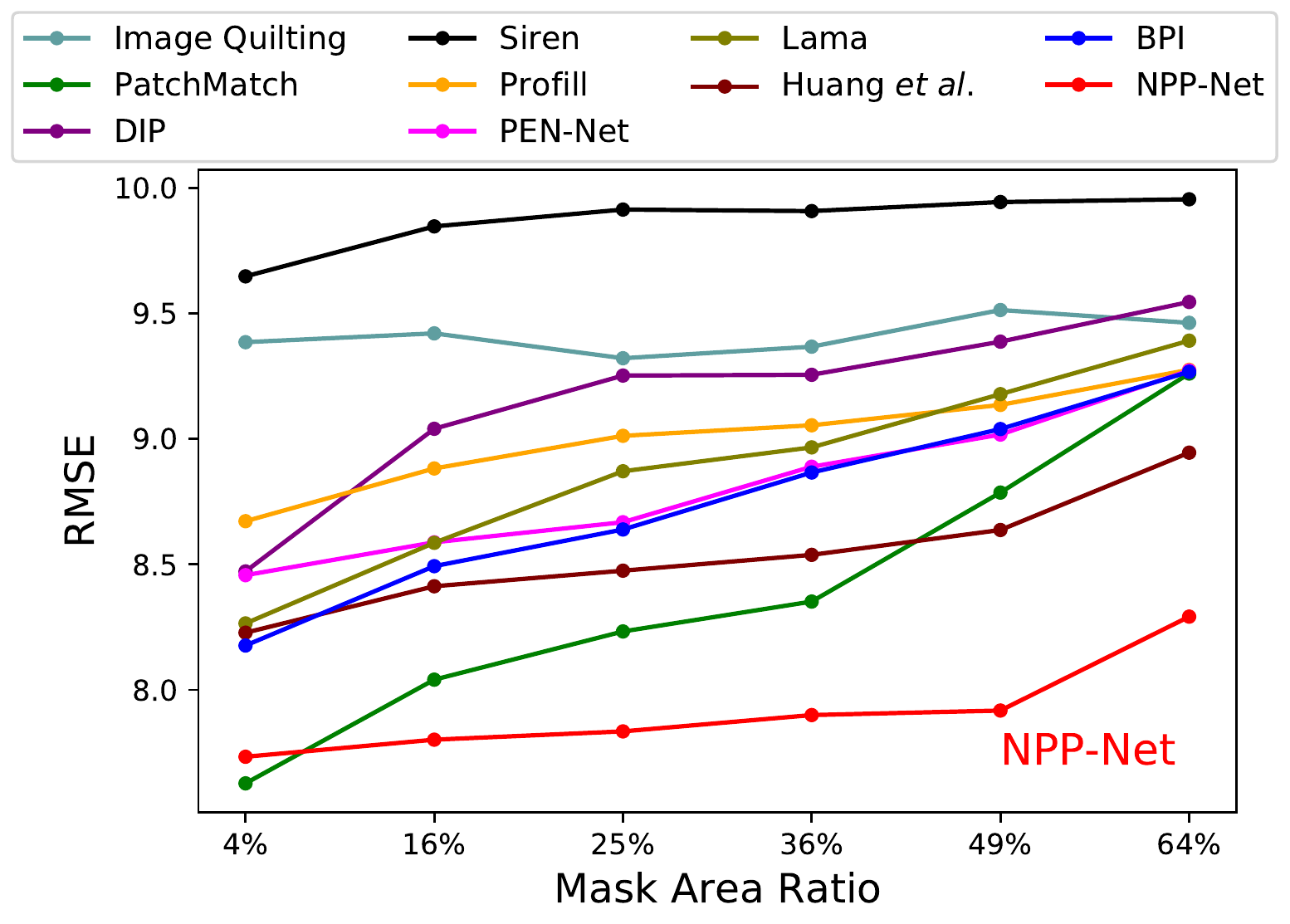}}
  \caption{\normalsize (e) RMSE   \normalsize}
\end{subfigure}
\caption{
Comparison of model performances for different mask sizes in the DTD dataset. FID is evaluated in the full image, and the other four metrics are tested in the unknown regions.  
}
\vspace{-0.2in}
\label{mask_size_dtd}
\end{figure}

\begin{figure*}[h]
    \centering
\begin{scriptsize}
\begin{flushleft}
\hspace{75pt} 
4 \% 
\hspace{25pt} 
16 \%
\hspace{25pt} 
25 \%
\hspace{25pt} 
36 \%
\hspace{25pt} 
49 \%
\hspace{28pt} 
64 \%
\end{flushleft}
\end{scriptsize}
 \vspace{-7mm}
\begin{center}
\begin{tabular}[t]{c}
 \\[.2cm]
 \rotatebox[origin=c]{0}{
 \shortstack[c]{\scriptsize Input   \\ \;}
 }  \\[.7cm]
 \rotatebox[origin=c]{0}{
 \shortstack[c]{\scriptsize Image   \\\scriptsize Quilting}
 }\\[1.0cm]
 \rotatebox[origin=c]{0}{
 \shortstack[c]{ \scriptsize PatchMatch \\ \;} 
 }\\[1.1cm]
 \rotatebox[origin=c]{0}{
 \shortstack[c]{\scriptsize DIP \\ \;} 
 }\\[1.2cm]
  \rotatebox[origin=c]{0}{
  \shortstack[c]{\scriptsize Siren \\ \;}
 }\\[1.1cm]
  \rotatebox[origin=c]{0}{
 \shortstack[c]{\scriptsize PEN-Net \\ \;}  }\\[1.1cm]
 \rotatebox[origin=c]{0}{
 \shortstack[c]{\scriptsize ProFill \\ \;}  }\\[1.1cm]
  \rotatebox[origin=c]{0}{
 \shortstack[c]{\scriptsize Lama \\ \;}  }\\[.8cm]
 \rotatebox[origin=c]{0}{
 \shortstack[c]{\scriptsize Huang \\\scriptsize \etal }  }\\[1.2cm]
 \rotatebox[origin=c]{0}{
 \shortstack[c]{\scriptsize BPI \\ \;}   }\\[1.0cm]
 \rotatebox[origin=c]{0}{
 \shortstack[c]{\scriptsize NPP-Net \\ \;}   }\\[1.0cm]
 \rotatebox[origin=c]{0}{
 \shortstack[c]{\scriptsize Ground \\   \scriptsize Truth}}  
\end{tabular}
\includegraphics[width=0.84\linewidth,valign=t]{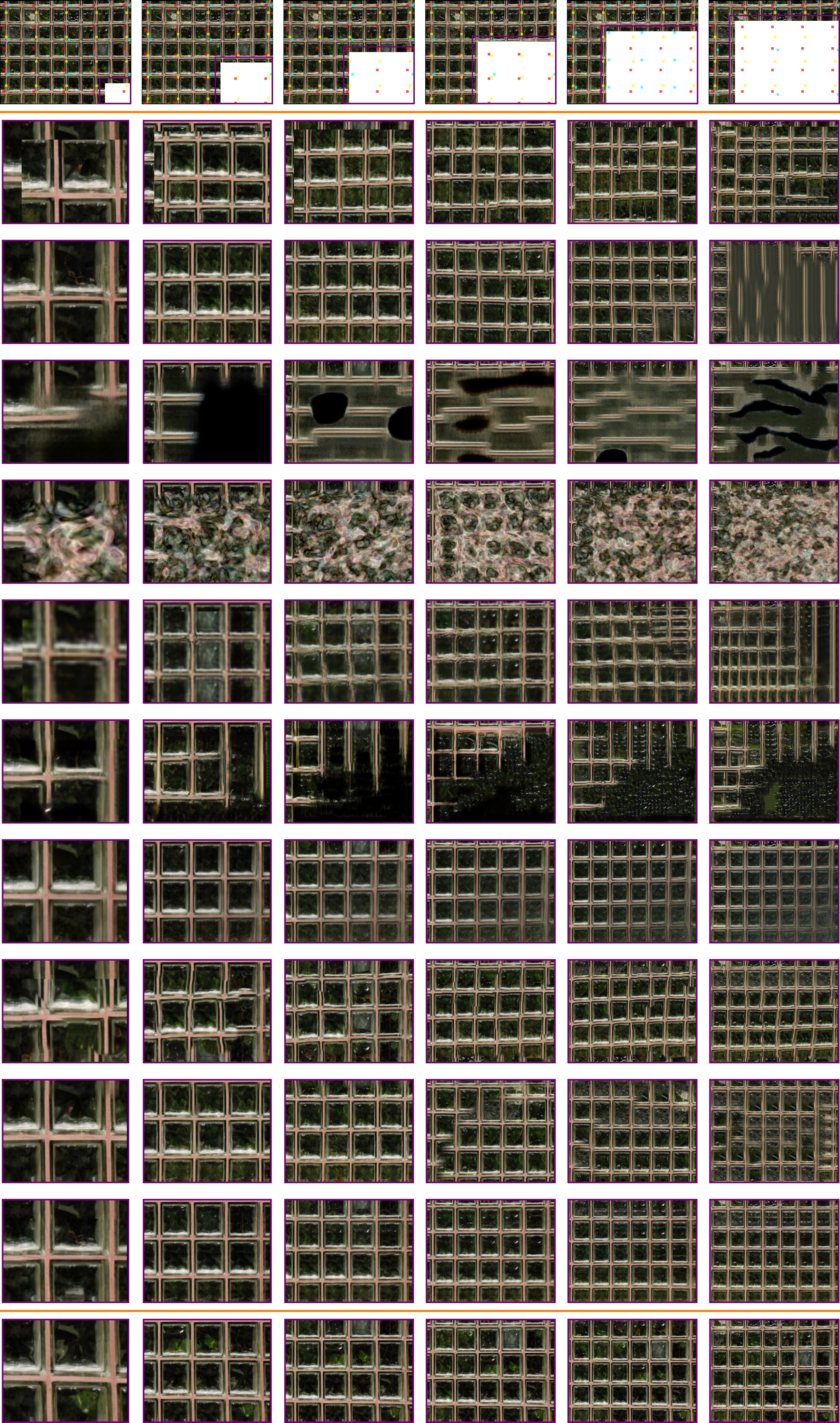}
\end{center} 
    \vspace{-4.5mm}
    \caption{  Qualitative results for different size of masks. From left to right, the mask sizes are 4\%, 16\%, 25\%, 36\%, 49\% and 64\%. NPP-Net outperforms all the baselines.  
    }
    \label{mask_ratio:mask_ratio_20150914132954-b854d373-me}
 \end{figure*}

\begin{figure*}[h]
    \centering
\begin{scriptsize}
\begin{flushleft}
\hspace{75pt} 
4 \% 
\hspace{25pt} 
16 \%
\hspace{25pt} 
25 \%
\hspace{25pt} 
36 \%
\hspace{25pt} 
49 \%
\hspace{28pt} 
64 \%
\end{flushleft}
\end{scriptsize}
 \vspace{-7mm}
\begin{center}
\begin{tabular}[t]{c}
 \\[.1cm]
 \rotatebox[origin=c]{0}{
 \shortstack[c]{\scriptsize Input   \\ \;}
 }  \\[.6cm]
 \rotatebox[origin=c]{0}{
 \shortstack[c]{\scriptsize Image   \\\scriptsize Quilting}
 }\\[.8cm]
 \rotatebox[origin=c]{0}{
 \shortstack[c]{ \scriptsize PatchMatch \\ \;} 
 }\\[.7cm]
 \rotatebox[origin=c]{0}{
 \shortstack[c]{\scriptsize DIP \\ \;} 
 }\\[.8cm]
  \rotatebox[origin=c]{0}{
  \shortstack[c]{\scriptsize Siren \\ \;}
 }\\[.9cm]
  \rotatebox[origin=c]{0}{
 \shortstack[c]{\scriptsize PEN-Net \\ \;}  }\\[.85cm]
 \rotatebox[origin=c]{0}{
 \shortstack[c]{\scriptsize ProFill \\ \;}  }\\[.85cm]
  \rotatebox[origin=c]{0}{
 \shortstack[c]{\scriptsize Lama \\ \;}  }\\[.65cm]
 \rotatebox[origin=c]{0}{
 \shortstack[c]{\scriptsize Huang \\\scriptsize \etal }  }\\[.85cm]
 \rotatebox[origin=c]{0}{
 \shortstack[c]{\scriptsize BPI \\ \;}   }\\[.85cm]
 \rotatebox[origin=c]{0}{
 \shortstack[c]{\scriptsize NPP-Net \\ \;}   }\\[.75cm]
 \rotatebox[origin=c]{0}{
 \shortstack[c]{\scriptsize Ground \\   \scriptsize Truth}}  
\end{tabular}
\includegraphics[width=0.84\linewidth,valign=t]{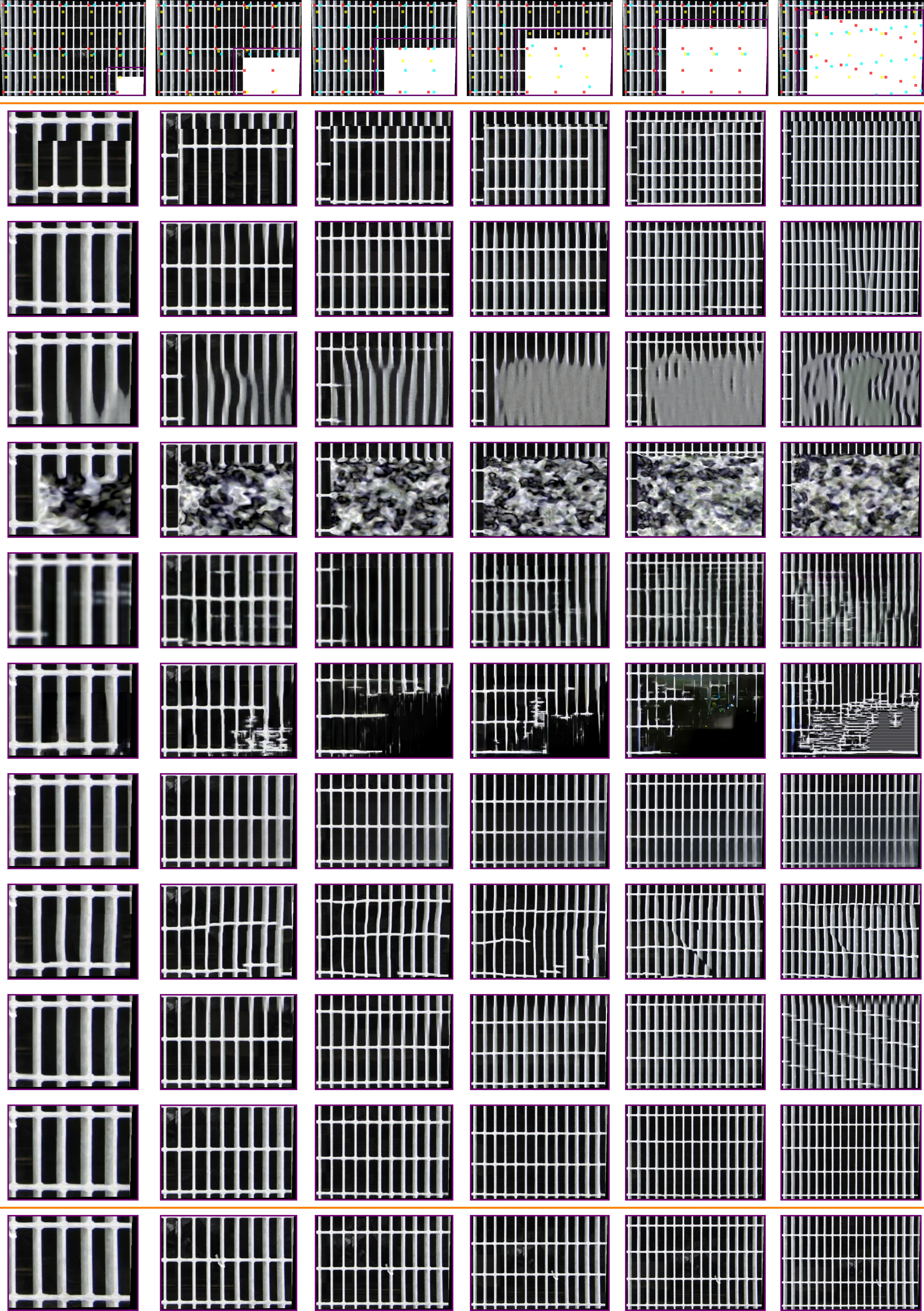}
\end{center} 
    \vspace{-4.5mm}
    \caption{  Qualitative results for different size of masks. From left to right, the mask sizes are 4\%, 16\%, 25\%, 36\%, 49\% and 64\%. NPP-Net outperforms all the baselines.  
    }
    \label{mask_ratio:mask_ratio_20150911152116-7d1f022c-me}
 \end{figure*}

We show the all the evaluation metrics for the influence of mask size in NRTDB and DTD datasets in Figure \ref{mask_size_nrtdb} and Figure \ref{mask_size_dtd}, respectively. 
While LPIPS, SSIM, PSNR, and RMSE are evaluated only in unknown regions, we evaluate FID in the full image since FID is inaccurate if the image resolution is very small (\eg 4\% of original image). NPP-Net outperforms other baselines, especially when the mask size is large. 
Figure \ref{mask_ratio:mask_ratio_20150914132954-b854d373-me} and Figure \ref{mask_ratio:mask_ratio_20150911152116-7d1f022c-me}  show the qualitative results, where NPP-Net performs the best among all methods.

\clearpage
 \subsection{Periodicity Refinement}

The completed output of NPP-Net can also improve the periodicity detection since more known and reliable information is available for the detection, especially for large masks. The better periodicity can help BPI and NPP-Net achieve better results. Note that, we do not adopt this strategy on Huang \etals because it is not easy to incorporate the detection method into their pipeline. 

Specifically, given a masked NPP image, we perform our pipeline to complete the image. Then we input the completed image to our pipeline for periodicity proposal, but still, train NPP-Net only in the originally known regions for better completion results. A similar process is also applied to BPI. In practice, we only refine the masked images with the ratio of unknown regions to the sum of unknown and known ones larger than 40\%. 

\begin{figure*}[h]
\vspace{-16pt} 
    \centering
\begin{small}
\begin{flushleft}
\hspace{13pt} 
 \shortstack[c]{ Input   \\ \;} 
\hspace{11pt} 
 \shortstack[c]{ Input   \\ (Refine)} 
\hspace{19pt} 
 \shortstack[c]{ BPI   \\ \;} 
\hspace{18pt}  
 \shortstack[c]{ BPI   \\ (Refine)} 
\hspace{14pt} 
 \shortstack[c]{ NPP-Net   \\ \;} 
\hspace{8pt}  
 \shortstack[c]{ NPP-Net   \\ (Refine)} 
\hspace{14pt} 
 \shortstack[c]{ GT   \\ \;} 

\end{flushleft}
\end{small}
\vspace*{-8pt}
    \begin{tabular}{cc}
        \includegraphics[width=0.99\textwidth]{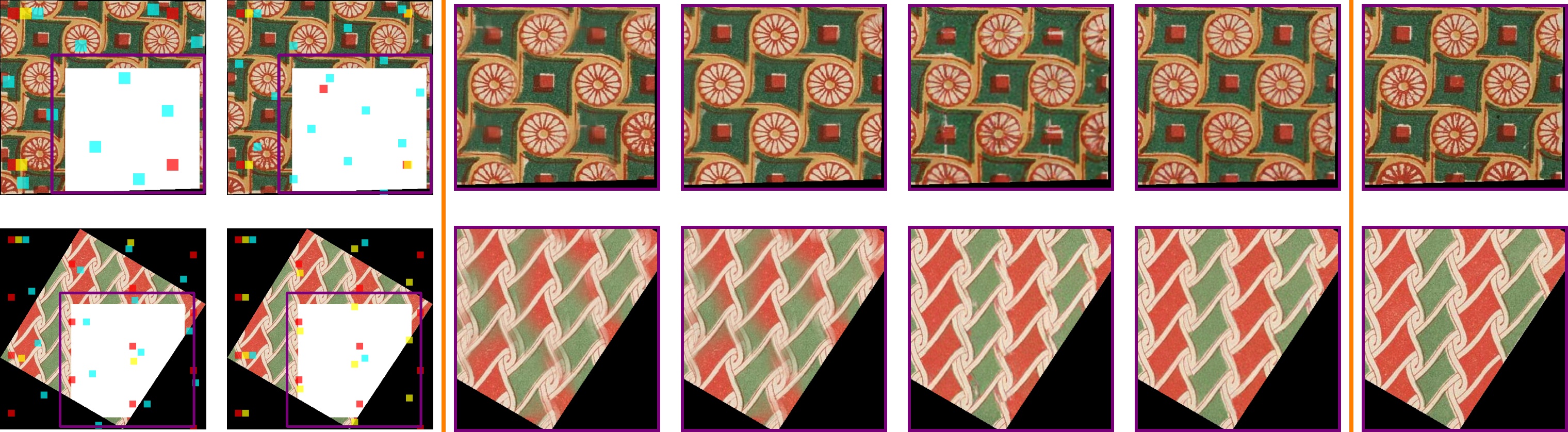}\\
    \end{tabular}
    \caption{ Qualitative results for periodicity refinement. NPP-Net with refinement outperforms all other compared methods. 
    }
    \label{refinement}
 \end{figure*}

Table \ref{tab3:refinement} and  Figure \ref{refinement} show the quantitative and qualitative results for the refinement in NRTDB dataset, respectively. Methods with refinement perform better than those without refinement. Among all the methods, NPP-Net with refinement performs the best, demonstrating the effectiveness of refinement for large masks.

\begin{table*}[!h]
\vspace{-5mm}
\centering
			\scalebox{1}{
	 \begin{tabular}{cccccc}	
	 \toprule 
	  \multirow{ 2}{*}{Method}  & \multicolumn{5}{c}{Only Unknown Regions} \\
\cmidrule(lr){2-6} 
	 	   & LPIPS $\downarrow$ & SSIM $\uparrow$ & PSNR $\uparrow$& FID $\downarrow$ & RMSE $\downarrow$    \\
	 	 \midrule 
	 	BPI    & 0.252 & 0.385 &  14.83  & 82.77  & 9.496\\  
	 	
	 	BPI (Refine)   & 0.244 & 0.402 &  14.93  & 90.16 &  9.462 \\  
	 		 	\midrule
	 	NPP-Net   & 0.186 & 0.626  & 17.55 & 74.10  & 9.001   \\   
	 	NPP-Net (Refine)    & \textbf{ 0.171} & \textbf{0.636}  & \textbf{17.87} &  \textbf{68.20} & \textbf{8.904}   \\   
	 	\bottomrule
	 \end{tabular}}	 	
	 \vspace{2mm}
	 	 		\caption{Comparison of BPI and NPP-Net with and without periodicity refinement. NPP-Net with refinement outperforms all other compared methods.  }
	 	\label{tab3:refinement}
	\end{table*}

\subsection{Learnable Periodicity}
\begin{figure}[!h]
\vspace{5mm}
    \captionsetup[subfigure]{labelformat=empty, font=small}
        \captionsetup[subfigure]{labelformat=empty, font=small}
    \begin{subfigure}{.49\linewidth}
  \centerline{\includegraphics[width=.85\textwidth]{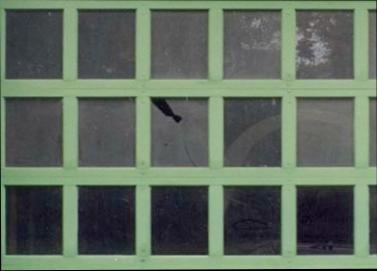}}
  \caption{\normalsize (a) Input \normalsize}
    \vspace{5mm}
\end{subfigure}
    \begin{subfigure}{.49\linewidth}
  \centerline{\includegraphics[width=.6\textwidth]{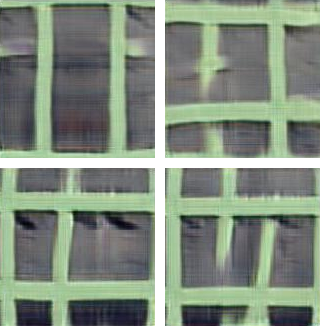}}
  \caption{\normalsize (b) PSGAN Output \normalsize}
    \vspace{5mm}
\end{subfigure}
\caption{
Texture synthesis results of PSGAN. The left image shows the input and the right image shows four synthesized samples of PSGAN.
}
\vspace{-0.2in}
\label{psgan}
\end{figure}

There are several  texture synthesis methods~\cite{chen2022exemplar,jetchev2016texture} that directly treat periodicity as a learnable  parameter during training. However, as mentioned in the main paper, this does not work for many real-world NPP images due to the presence of local variations. 
Predicting periodicity before training stabilizes the network optimization. 
Figure \ref{psgan} shows four texture synthesis samples from PSGAN.

\clearpage

\section{NPP Segmentation}
\subsection{Method}

We aim to segment the non-periodic regions in an NPP image in an unsupervised manner. The key idea is to detect initial non-periodic regions, treat them as the unknown mask, and relabel regions with low reconstruction error as periodic regions. 

Specifically, we first apply a traditional image segmentation method~\cite{Segmentation2017} to generate an initial segmentation for non-periodic regions, treated as unknown regions. This segmentation method first divides image pixels as superpixels and uses Gaussian Mixture Model for a coarse segmentation, which is further refined by GraphCut. We treat the class that contains the largest number of pixels as the periodic class, and the rest of the classes as non-periodic classes for initialization. 
One drawback of this method is that it often over-segments the non-periodic regions. Thus we use NPP-Net to refine the initial segmentation. In detail, we use the same pipeline of image completion to complete the unknown (non-periodic) regions. For every pixel $\vect{x}$ in non-periodic regions, we can compute the reconstruction error using two metrics since the ground-truth value of $\vect{x}$ is known. The first metric is the L1 distance, comparing the difference between output and ground truth pixels, given by:
\begin{equation}
    S_1(\vect{x}) = |\hat{G}(\vect{x}) - G(\vect{x})|, 
\end{equation}
where $\hat{G}(\vect{x})$ and $G(\vect{x})$ are the  grayscale value of output and ground truth at $\vect{x}$, respectively. The second metric is perceptual distance~\cite{zhang2018perceptual} based on pretrained network, given by:
\begin{equation}
    S_2(\vect{x}) = || P_{\hat{I}}(\vect{x}) - P_I(\vect{x})||_2,
\end{equation}
where $P_{\hat{I}}$ and $P_{{I}}$ denote the normalized perceptual activation image (first layer of AlexNet) of the output and ground truth NPP image, respectively. Only $\vect{x}$ with low reconstruction error is eligible for relabelling, and these $\vect{x}$ are defined as a set $S$, given by: 
\begin{equation}
    S = \{\vect{x}|  S_1(\vect{x}) < \epsilon_1, S_2(\vect{x}) < \epsilon_2\},
\end{equation}
where $\epsilon_1$ and $\epsilon_2$ are constants. Finally, we relabel $\vect{x} \in S$ to periodic class to obtain our segmentation. For implementation details, we set  $\epsilon_1 = 0.15$ and $\epsilon_2 = 0.3$. Also, we blur the input NPP image before using our pipeline to remove fine details in the image because we only focus on the global structure.

\vspace{-4mm}
\subsection{Comparison with Baselines}

We compare with HRNet~\cite{SunXLW19} and Borovec \etal~\cite{Segmentation2017}. The first one is a network trained on a large dataset (ADE20K~\cite{zhou2017scene}), and the latter one is an unsupervised method based on Gaussian Mixture Model and Graphcut.

\begin{figure*}[h]
\vspace{20pt} 
    \centering
\begin{small}
\begin{flushleft}
\hspace{40pt} 
 \shortstack[c]{ Input   } 
\hspace{45pt} 
 \shortstack[c]{ HRNet} 
\hspace{30pt} 
 \shortstack[c]{ Borovec \etal} 
\hspace{30pt} 
 \shortstack[c]{ NPP-Net } 
\end{flushleft}
\end{small}
\vspace*{-8pt}
    \begin{tabular}{cc}
        \includegraphics[width=0.9\textwidth]{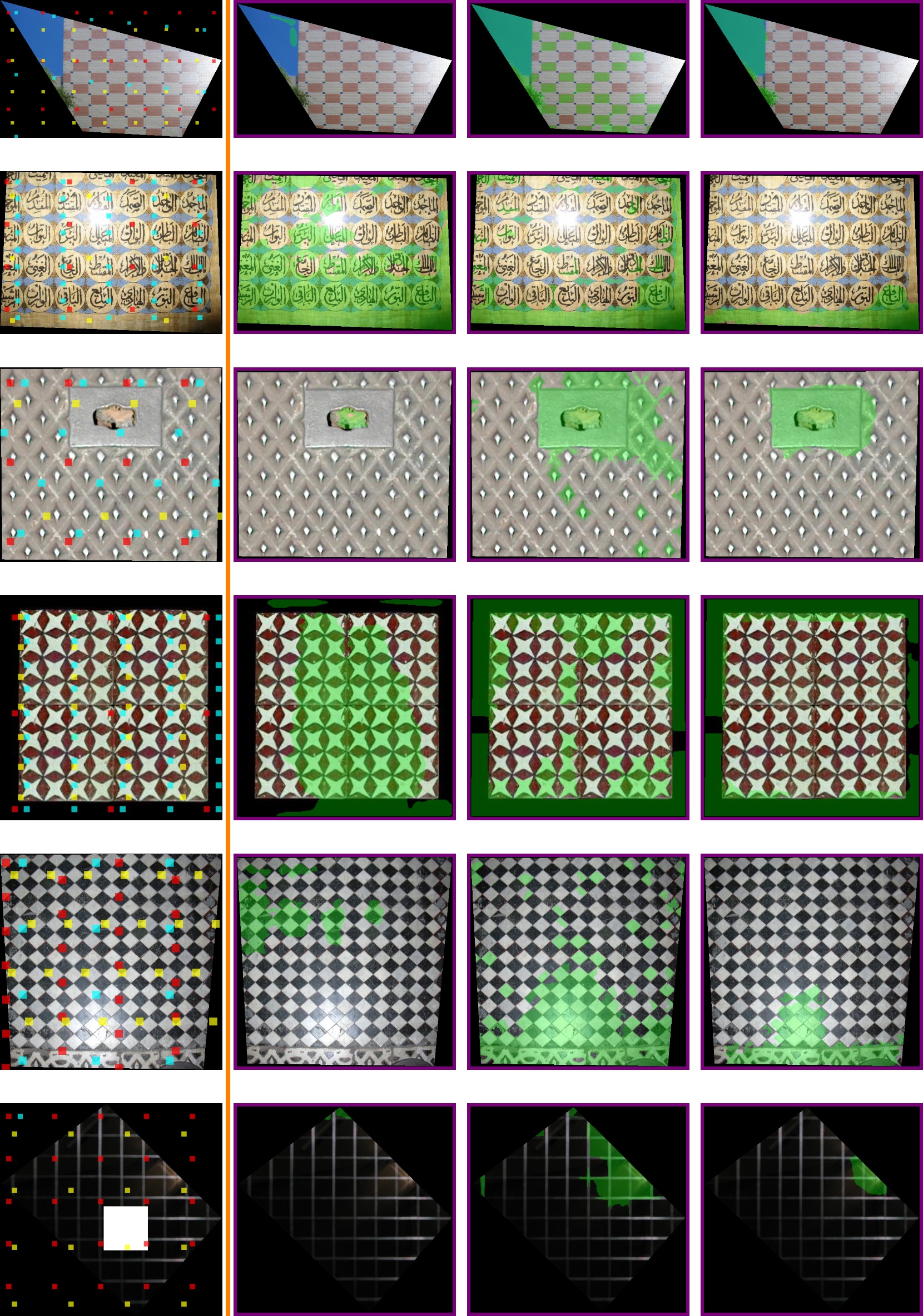}\\
    \end{tabular}
    \vspace{-3mm}
    \caption{ Qualitative results of non-periodic region segmentation. The green regions refer to non-periodic regions.  
    }
    \label{comparison_seg}
 \end{figure*}

We show NPP segmentation results in Figure \ref{comparison_seg}, where NPP-Net outperforms all the chosen baselines. HRNet cannot produce reasonable results because it is trained on a dataset of objects and scenes rather than periodic patterns. Borovec \etals often over-segments the non-periodic regions since it is based on low-level features without high-level periodicity understanding.

 \clearpage
 \subsection{NPP Classification}
 
The NPP segmentation can be extended to classify whether an input image is NPP or not. This can serve as a pre-filtering step before applying NPP-Net for completion. Also, it can be adopted to collect NPP images from a large collection of natural images.  

Given an arbitrary image with an unknown mask, we label the regions around the unknown mask as initial non-periodic regions. The remaining known regions are treated as periodic regions for training. Then we apply NPP segmentation to relabel the non-periodic regions. If more than 50\% of pixels are relabeled, we classify it as an NPP image.

We test this classification method on the DTD dataset.
We select 40 non-NPP images from the class \enquote{potholed} in the original dataset, together with 258 images in the chosen DTD  dataset for experiments. 
We generate an unknown mask with height and width equal to 50\% of the image height and width (starting from the bottom right) respectively.
We initialize the initial non-periodic region as a mask with height and width equal to 70\% of the image height and width (but excluding the unknown region).

Quantitatively, the precision of NPP classification for all images (298 images), only non-NPP images (258 images), and only NPP images (40 images) are 82.2\%, 81.9\%, 84.6\% respectively.

\begin{figure}[!h]
    \captionsetup[subfigure]{labelformat=empty, font=small}
\begin{subfigure}{.23\linewidth}
  \centerline{\includegraphics[width=0.94\textwidth]{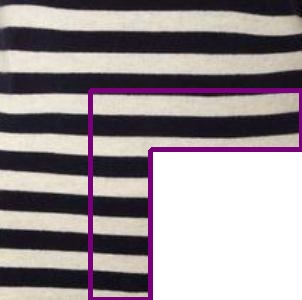}}
\end{subfigure}
\hfill
\begin{subfigure}{.23\linewidth}
  \centerline{\includegraphics[width=0.94\textwidth]{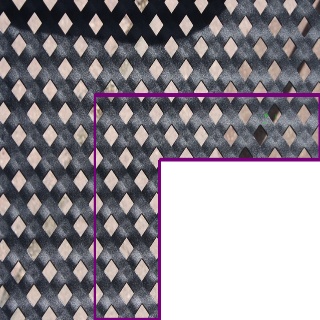}}
\end{subfigure}
\hfill
\begin{subfigure}{.23\linewidth}
  \centerline{\includegraphics[width=0.94\textwidth]{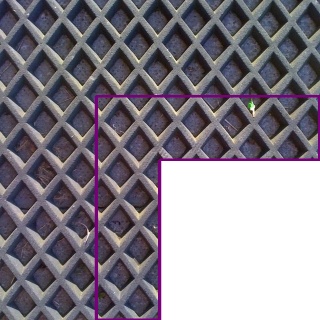}}
\end{subfigure}
\begin{subfigure}{.23\linewidth}
  \centerline{\includegraphics[width=0.94\textwidth]{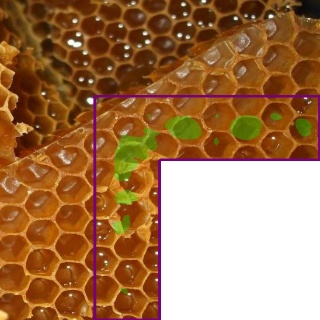}}
\end{subfigure}
\\
\begin{subfigure}{.23\linewidth}
  \centerline{\includegraphics[width=0.94\textwidth]{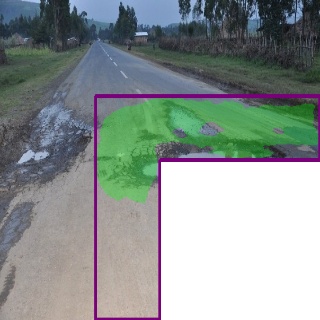}}
\end{subfigure}
\hfill
\begin{subfigure}{.23\linewidth}
  \centerline{\includegraphics[width=0.94\textwidth]{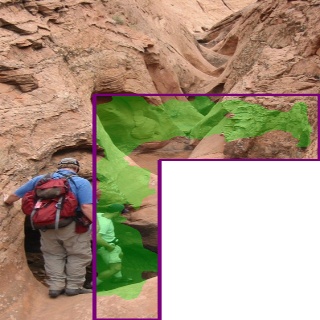}}
\end{subfigure}
\hfill
\begin{subfigure}{.23\linewidth}
  \centerline{\includegraphics[width=0.94\textwidth]{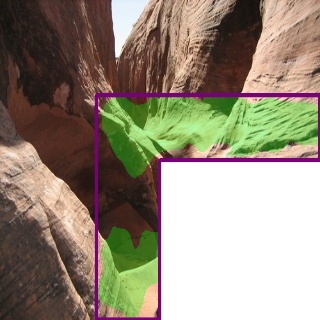}}
\end{subfigure}
\begin{subfigure}{.23\linewidth}
  \centerline{\includegraphics[width=0.94\textwidth]{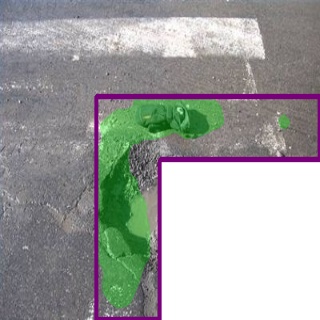}}
\end{subfigure}
\caption{
The qualitative results of NPP classification. The regions inside the purple box are the initial non-periodic regions, and the regions highlighted in green refer to the final periodic regions. 
NPP-Net correctly classifies images in the first row (NPP images) and second row (non-NPP images).
The more regions in green, the more likely the image is a non-NPP image. 
}
\label{classifcation_good}
\end{figure}

\begin{figure}[!t]
    \captionsetup[subfigure]{labelformat=empty, font=small}
\begin{subfigure}{.23\linewidth}
  \centerline{\includegraphics[width=0.94\textwidth]{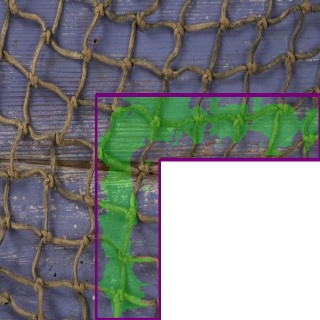}}
\end{subfigure}
\hfill
\begin{subfigure}{.23\linewidth}
  \centerline{\includegraphics[width=0.94\textwidth]{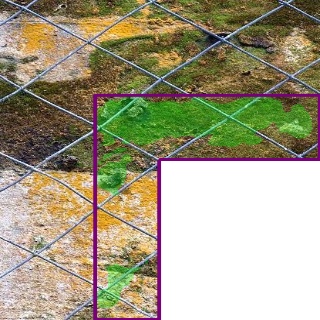}}
\end{subfigure}
\hfill
\begin{subfigure}{.23\linewidth}
  \centerline{\includegraphics[width=0.94\textwidth]{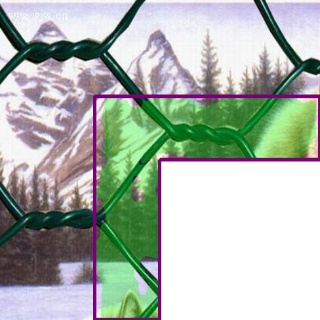}}
\end{subfigure}
\begin{subfigure}{.23\linewidth}
  \centerline{\includegraphics[width=0.94\textwidth]{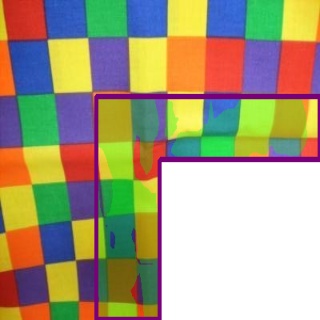}}
\end{subfigure}
\\
\begin{subfigure}{.23\linewidth}
  \centerline{\includegraphics[width=0.94\textwidth]{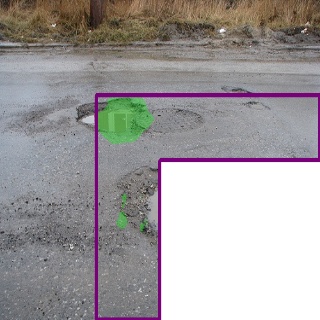}}
\end{subfigure}
\hfill
\begin{subfigure}{.23\linewidth}
  \centerline{\includegraphics[width=0.94\textwidth]{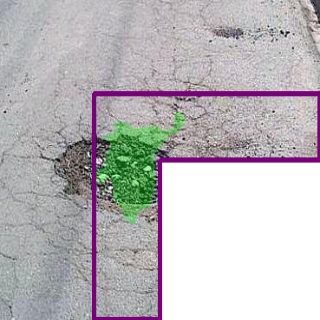}}
\end{subfigure}
\hfill
\begin{subfigure}{.23\linewidth}
  \centerline{\includegraphics[width=0.94\textwidth]{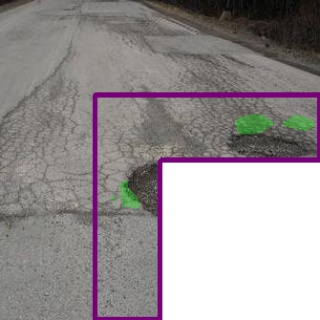}}
\end{subfigure}
\begin{subfigure}{.23\linewidth}
  \centerline{\includegraphics[width=0.94\textwidth]{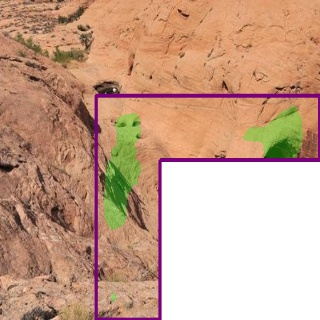}}
\end{subfigure}
\caption{
The failure cases of NPP classification. The regions inside the purple box are the initial non-periodic regions, and the regions highlighted in green refer to the final periodic regions. 
NPP-Net fails to correctly classify images in the first row (NPP images) and second row (non-NPP images) because they have homogenous contents or complicated backgrounds.
The more regions in green, the more likely the image is a non-NPP image. 
}
\label{classifcation_bad}
\end{figure}

Figure \ref{classifcation_good} shows the qualitative results of the NPP classification. NPP-Net correctly classifies the images even if the strong local variations are presented (row 1 column 4). 
Besides, we also show the failure cases in Figure \ref{classifcation_bad}, where NPP-Net fails for the images with homogenous contents or complicated backgrounds.

 \clearpage

\section{NPP Remapping}

\begin{figure}[!b]
    \captionsetup[subfigure]{labelformat=empty, font=small}
\begin{subfigure}{.3\linewidth}
  \centerline{\includegraphics[width=.57\textwidth]{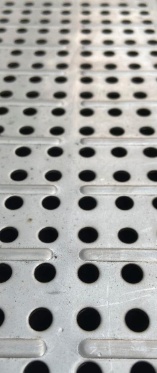}}
  \caption{(a) Perspective NPP}
\end{subfigure}
\hfill
\begin{subfigure}{.3\linewidth}
  \centerline{\includegraphics[width=.94\textwidth]{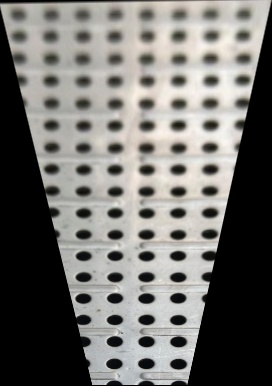}}
    \caption{(b) Rectified NPP}
\end{subfigure}
\hfill
\begin{subfigure}{.3\linewidth}
  \centerline{\includegraphics[width=.94\textwidth]{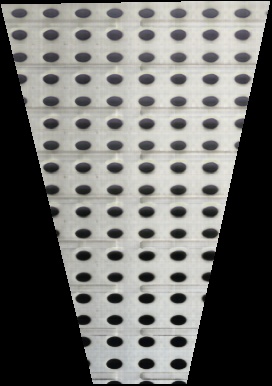}}
      \caption{(c) Output }
\end{subfigure}
\caption{
Illustration of an NPP Remapping scenario. 
}
\label{remapping_illu}
\end{figure}

\subsection{Method}

NPP textures are usually adopted in various applications such as rendering. A perspective camera can be used to obtain real-world NPP texture by rectifying the captured NPP perspective image.
However, the potential blurry texture issue prevents us to obtain high-quality texture.
Consider an NPP captured by a perspective camera in a tilted angle, as shown in Figure \ref{remapping_illu} (a). The far-away motifs are blurry because the regions are out of focus. This problem becomes severe after rectification due to the remapping issue, shown in Figure \ref{remapping_illu} (b).  
The goal of NPP remapping is to recover the blurry regions (caused by image remapping errors) in the NPP images. It outputs clear NPP images by preserving local variations and global structure, shown in Figure \ref{remapping_illu} (c). 

The key idea is to detect the blurry regions from the input and treat them the same as the unknown regions in the completion task with minor modification. 
Specifically, we first detect the blurry regions using~\cite{Su2011}, treated as unknown regions. This prevents the patch loss from sampling ground truth patches in blurry regions for supervision. Simply treating it as a completion problem cannot preserve the local variations (\eg lighting) in the blurry regions, thus we modify the pixel loss to account for this issue. Instead of computing this loss only in known (clear) regions, we also compute the loss on unknown (blurry) regions, given by:
\begin{equation}
    \mathcal{L}_{pixel}(\vect{x}) = M(\vect{x}) \mathcal{L}_{rob} (\hat{C} 
    (\vect{x}), C(\vect{x}))  +   \sigma (1 - M(\vect{x})) \mathcal{L}_{rob}(\hat{C} 
    (\vect{x}), C(\vect{x})),
\end{equation}
where $\sigma$ is a constant weight. $M(\vect{x})$ is 1 if $\vect{x}$ in clear regions, and 0 in blurry regions. We keep the remaining part the same as completion to optimize NPP-Net.  
For implementation details, we set $\sigma = 0.3$.  

\subsection{Comparison with Baselines}
We consider two baselines. MPRNet~\cite{Zamir2021MPRNet} and SelfDeblur~\cite{ren2020neural} are state-of-the-art deblurring methods trained on large datasets and a single image, respectively.

\begin{figure*}[h]
\vspace{10pt} 
    \centering
\begin{small}
\begin{flushleft}
\hspace{20pt} 
 \shortstack[c]{ Input   } 
\hspace{23pt} 
 \shortstack[c]{ Input (Zoom In)} 
\hspace{18pt} 
 \shortstack[c]{ MPRNet} 
\hspace{18pt}  
 \shortstack[c]{ SelfDeblur} 
\hspace{18pt} 
 \shortstack[c]{ NPP-Net } 
\end{flushleft}
\end{small}
\vspace*{-8pt}
    \begin{tabular}{cc}
        \includegraphics[width=0.99\textwidth]{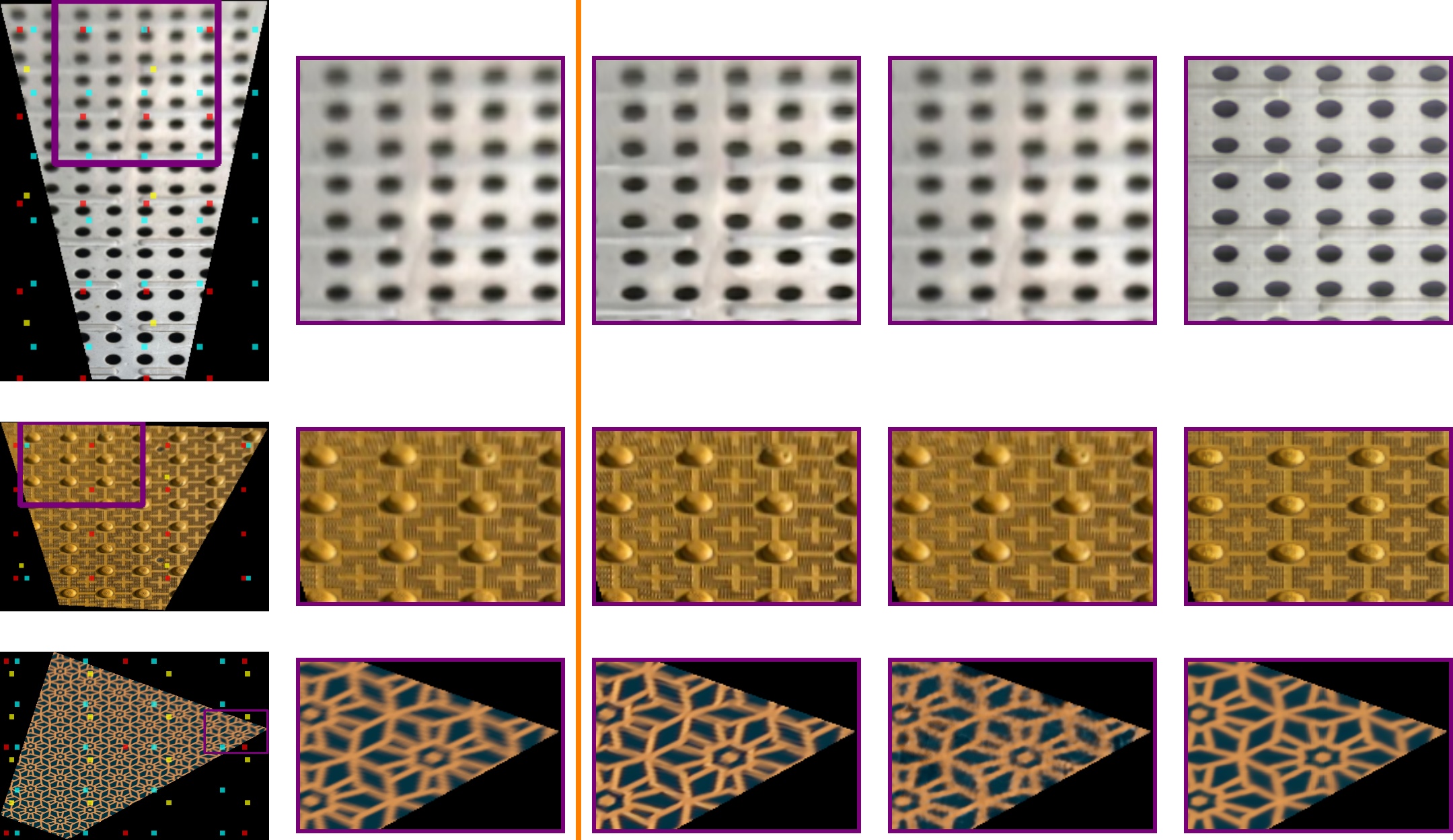}\\
    \end{tabular}
    \vspace{-4mm}
    \caption{ Qualitative results for image remapping. MPRNet generates pixelated result in row 3. NPP-Net outperforms all other baselines. 
    }
    \label{comparison_remapping2}
 \end{figure*}

\begin{figure*}[h]
\vspace{20pt} 
    \centering
\begin{small}
\begin{flushleft}
\hspace{20pt} 
 \shortstack[c]{ Input   } 
\hspace{23pt} 
 \shortstack[c]{ Input (Zoom In)} 
\hspace{18pt} 
 \shortstack[c]{ MPRNet} 
\hspace{18pt}  
 \shortstack[c]{ SelfDeblur} 
\hspace{18pt} 
 \shortstack[c]{ NPP-Net } 
\end{flushleft}
\end{small}
\vspace*{-8pt}
    \begin{tabular}{cc}
        \includegraphics[width=0.99\textwidth]{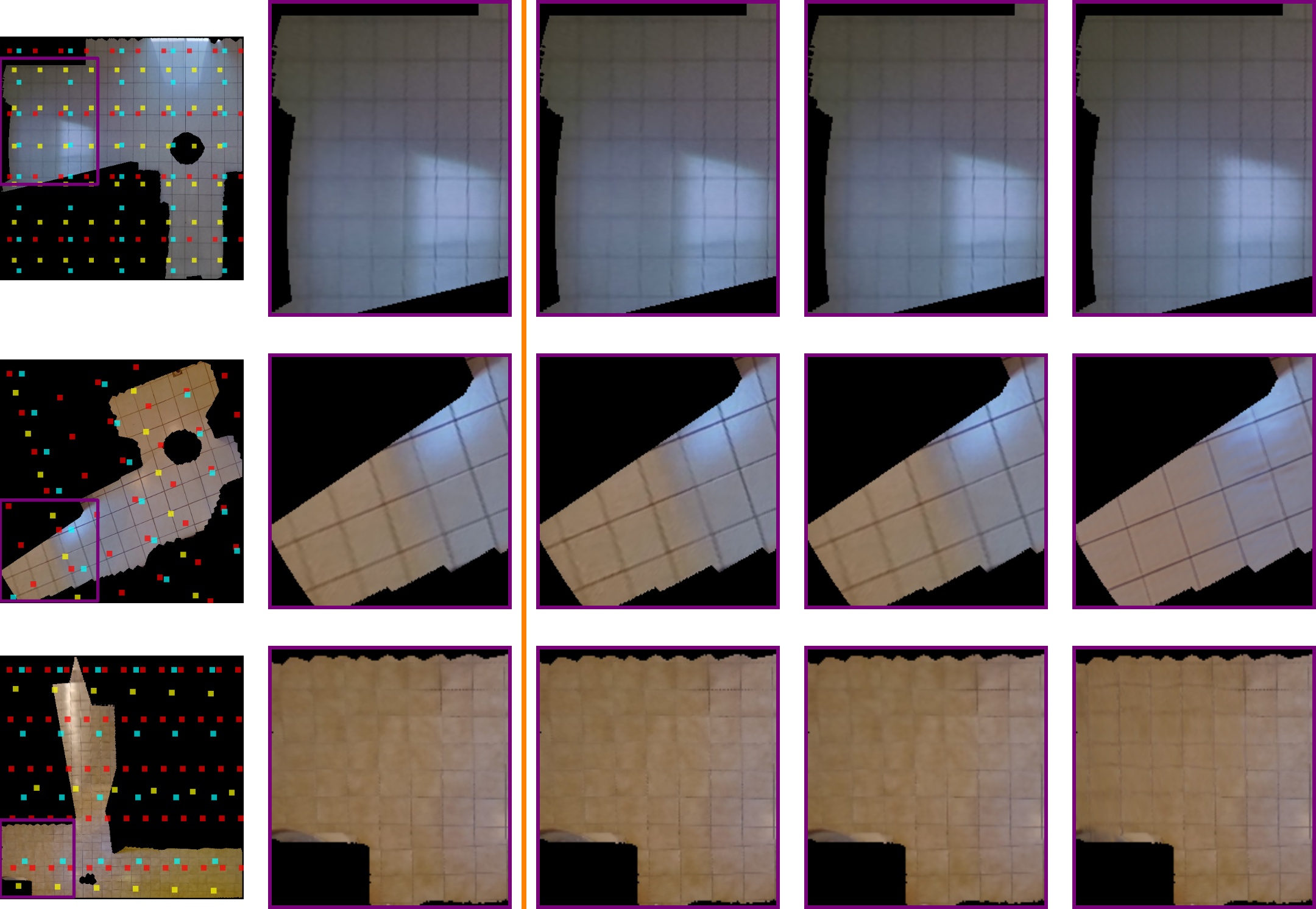}\\
    \end{tabular}
    \caption{ Qualitative results of remapped (orthographic) floor regions in indoor panorama for image remapping. The small black masks inside known regions is the tripod.   NPP-Net outperforms all other baselines. 
    }
    \label{comparison_remapping1}
 \end{figure*}

We show more qualitative results in Figure \ref{comparison_remapping2} and Figure \ref{comparison_remapping1}. Note that in Figure \ref{comparison_remapping2}, MPRNet produces pixelated pattern in row 3. 
NPP-Net outperforms all the baselines for the blurry region recovery. 

Also, Figure \ref{comparison_remapping1} demonstrates another application for image remapping~\cite{ZInD}: the indoor panorama may have an NPP floor with far-away blurry floor regions. We can use a pre-trained panorama segmentation method to segment the floor regions, and rectify them into orthographic view, shown as the inputs in Figure \ref{comparison_remapping1}. After recovering the blurry regions using NPP-Net, we can warp it back to the original panorama, resulting in a  panorama with clearer floor regions.

 \clearpage
 \section{Multi-Plane NPP Completion}

NPP-Net can be extended to multi-planar scenes. The key idea is to automatically segment and rectify each plane, apply NPP-Net on each plane for completion, and project the completed image back to the original image.

\begin{figure}[!h]
    \captionsetup[subfigure]{labelformat=empty, font=small}
\begin{subfigure}{.32\linewidth}
  \centerline{\includegraphics[width=.94\textwidth]{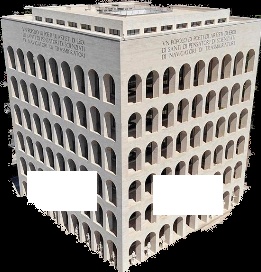}}
\caption{(a)}
\end{subfigure}
\hfill
\begin{subfigure}{.32\linewidth}
  \centerline{\includegraphics[width=.94\textwidth]{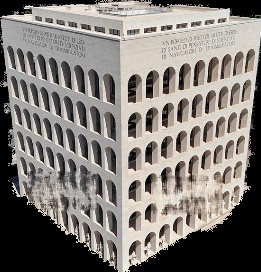}}
\caption{(b)}
\end{subfigure}
\hfill
\begin{subfigure}{.32\linewidth}
  \centerline{\includegraphics[width=.94\textwidth]{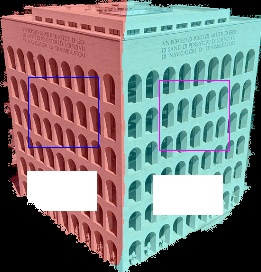}}
\caption{(c)}
\end{subfigure}
\\
\begin{subfigure}{.32\linewidth}
  \centerline{\includegraphics[width=.94\textwidth]{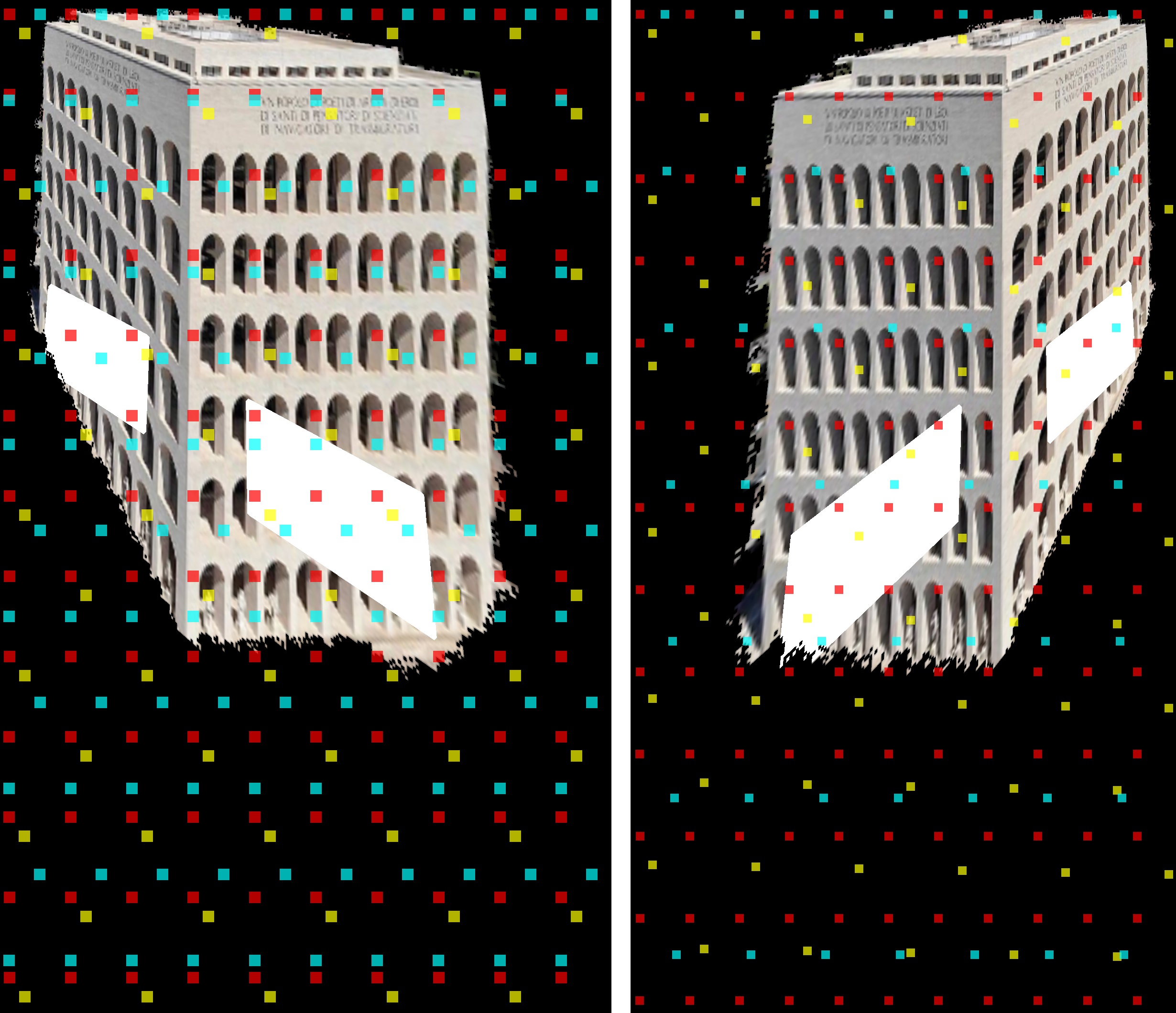}}
\caption{(d)}
\end{subfigure}
\hfill
\begin{subfigure}{.32\linewidth}
  \centerline{\includegraphics[width=.94\textwidth]{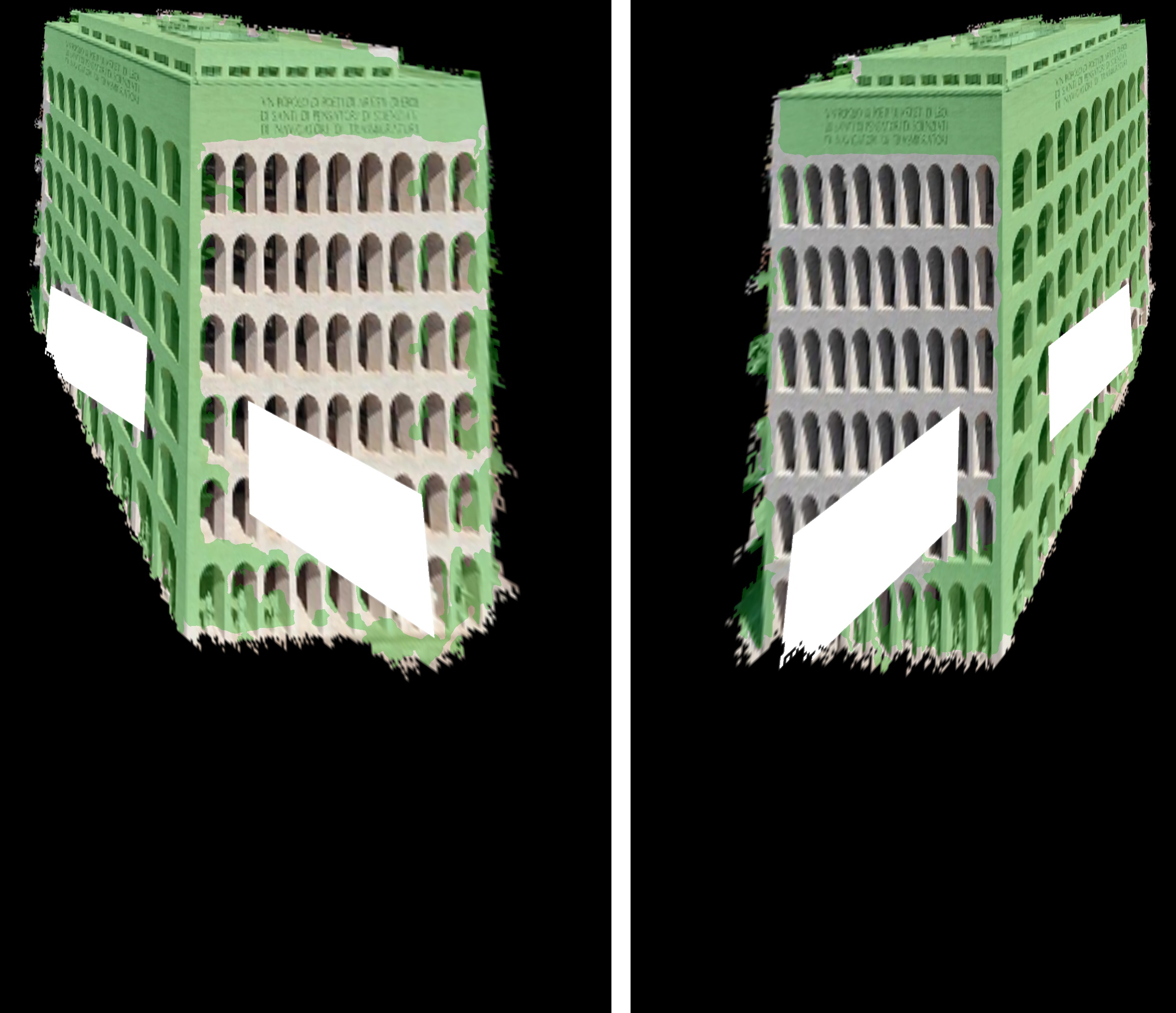}}
\caption{(e)}
\end{subfigure}
\hfill
\begin{subfigure}{.32\linewidth}
  \centerline{\includegraphics[width=.94\textwidth]{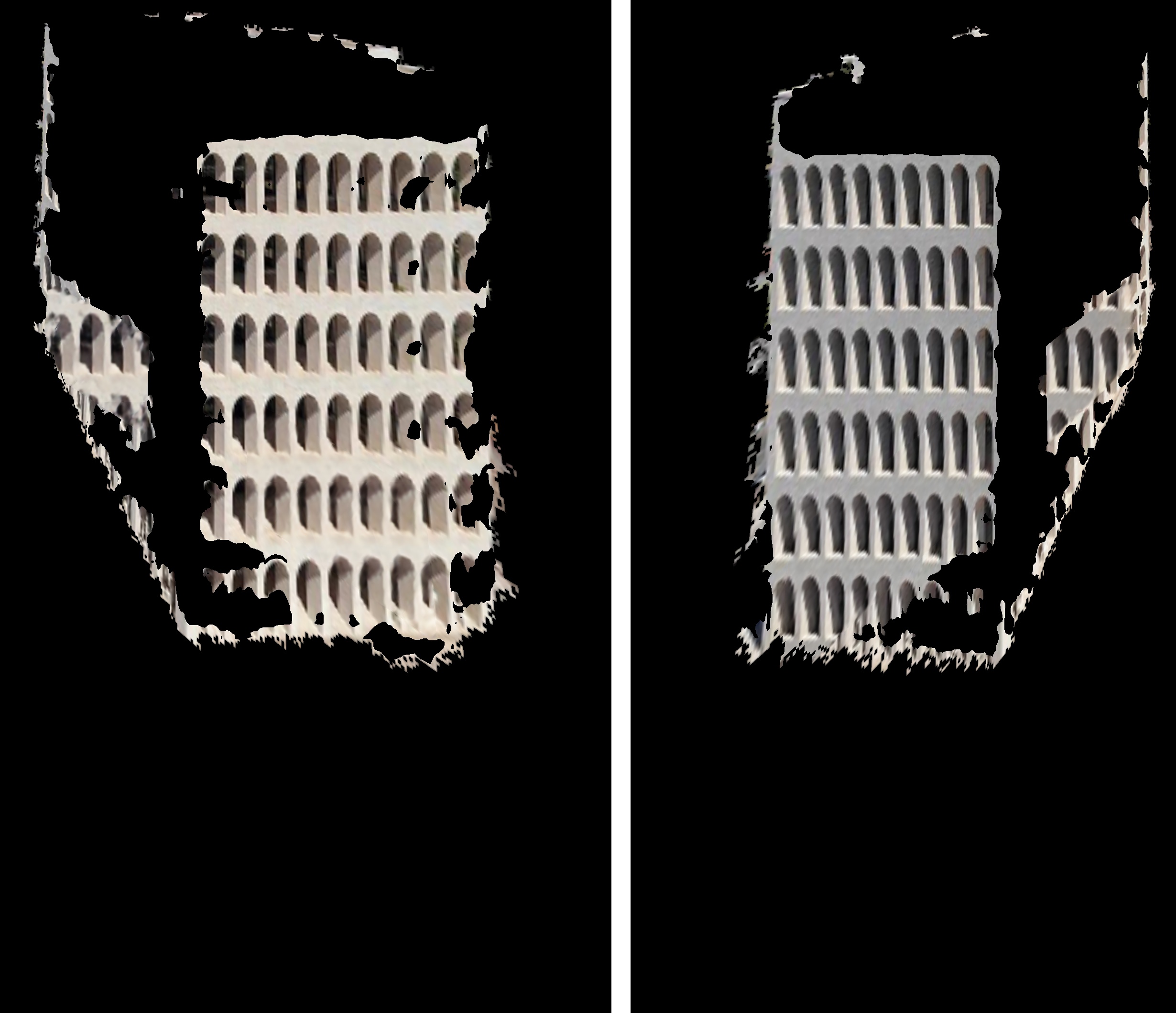}}
\caption{(f)}
\end{subfigure}
\\
\begin{subfigure}{.49\linewidth}
  \centerline{\includegraphics[width=.5\textwidth]{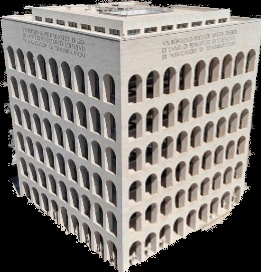}}
\caption{(g)}
\end{subfigure}
\begin{subfigure}{.49\linewidth}
  \centerline{\includegraphics[width=.5\textwidth]{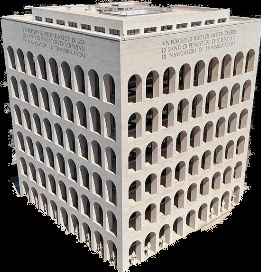}}
\caption{(h)}
\end{subfigure}
\caption{
Illustration of the multi-planar image completion extension for NPP-Net. }
\label{multi_plane_overview}
\end{figure}

Given a masked image (Figure \ref{multi_plane_overview} (a)) with different NPPs on different planes, we first adopt a pre-trained plane segmentation network~\cite{YuZLZG19} to obtain a coarse plane segmentation. Since this network is not trained on images with masks, we first use our \enquote{No Periodicity} variant to inpaint the masked (unknown) regions. The output is shown in Figure \ref{multi_plane_overview} (b). This process takes about 30 seconds. 
Then we can input this inpainted image in (b) to the pre-trained network to generate the coarse plane segmentation in Figure \ref{multi_plane_overview} (c). 

For each segmented plane, we select a bounding box using a similar strategy as the pseudo mask generation in periodicity searching (Sec \ref{ssec:implementation_details}). This bounding box is utilized as a reference to rectify the plane using TILT~\cite{zhang2012tilt}. 
Thus we do not require accurate segmentation since it is only used for bounding box selection.
Figure \ref{multi_plane_overview} (c) visualizes bounding boxes for two plane, and Figure \ref{multi_plane_overview} (d) shows the rectified planes. 

For each rectified plane, we detect the Top-3 periodicity (Figure \ref{multi_plane_overview} (d)) only using the rectified bounding box regions to remove the influence of other planes. Then we perform NPP segmentation to segment the non-periodic regions (mainly from other planes) and treat them as invalid pixels. In this case, the initial non-periodic regions are defined as all image regions excluding the bounding box regions. The segmentation results are shown in Figure \ref{multi_plane_overview} (e). After that, we run NPP completion on each plane, and the results are in Figure \ref{multi_plane_overview} (f). 

Finally, we transform all completed planes back to the original image coordinate system and recompose the image. The final inpainted and ground truth images are in Figure \ref{multi_plane_overview} (g) and (h) respectively. 

\begin{figure}[!h]
    \captionsetup[subfigure]{labelformat=empty, font=normalsize}
\begin{subfigure}{.49\linewidth}
  \centerline{\includegraphics[width=.94\textwidth]{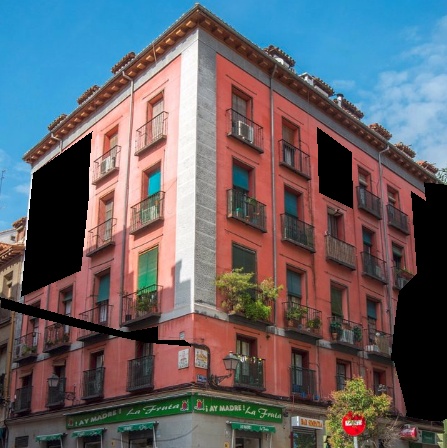}}
\caption{(a) Input}
\end{subfigure}
\begin{subfigure}{.49\linewidth}
  \centerline{\includegraphics[width=.94\textwidth]{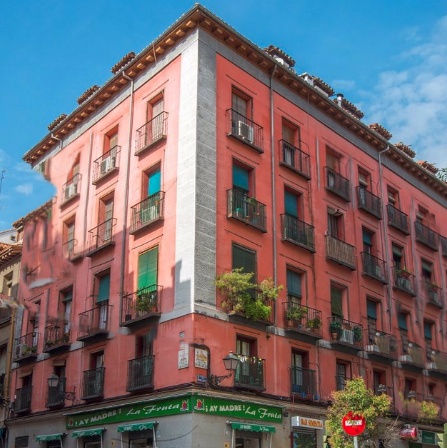}}
\caption{(b) Huang \etal}
\end{subfigure}
\\
\begin{subfigure}{.49\linewidth}
  \centerline{\includegraphics[width=.94\textwidth]{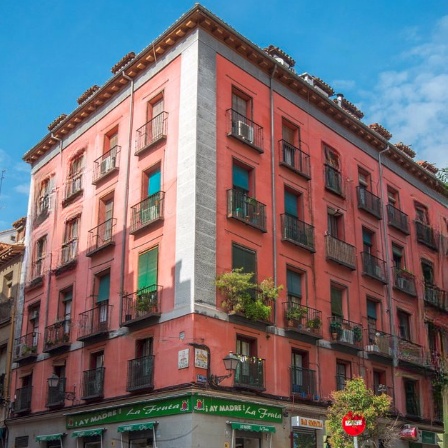}}
\caption{(c) Lama}
\end{subfigure}
\begin{subfigure}{.49\linewidth}
  \centerline{\includegraphics[width=.94\textwidth]{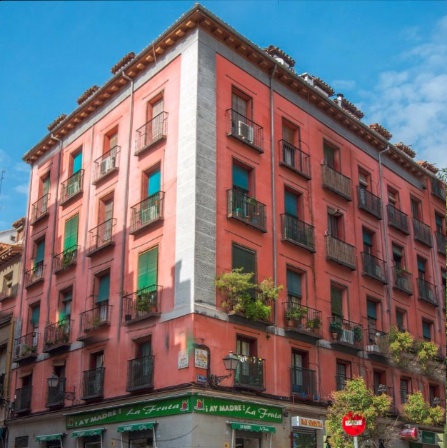}}
\caption{(d) BPI}
\end{subfigure}
\\
\begin{subfigure}{.49\linewidth}
  \centerline{\includegraphics[width=.94\textwidth]{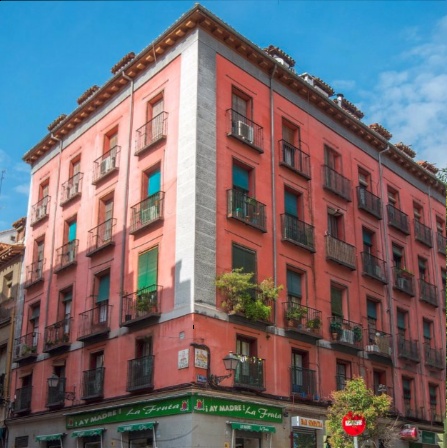}}
\caption{(e) NPP-Net}
\end{subfigure}
\begin{subfigure}{.49\linewidth}
  \centerline{\includegraphics[width=.94\textwidth]{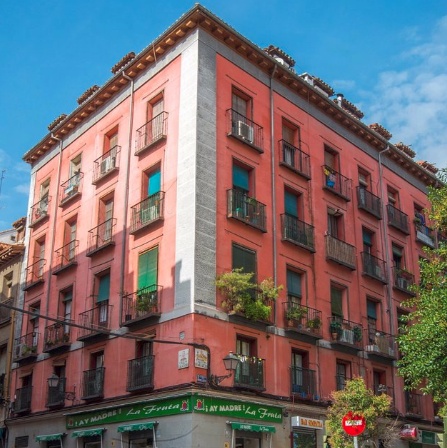}}
\caption{(f) GT}
\end{subfigure}
\caption{
Qualitative results of a multi-planar NPP scene.
}
\label{multi_plane1}
\end{figure}

\begin{figure}[!h]
    \captionsetup[subfigure]{labelformat=empty, font=normalsize}
\begin{subfigure}{.49\linewidth}
  \centerline{\includegraphics[width=.94\textwidth]{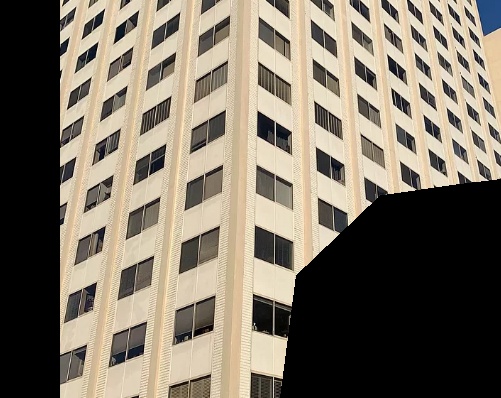}}
\caption{(a) Input}
\end{subfigure}
\begin{subfigure}{.49\linewidth}
  \centerline{\includegraphics[width=.94\textwidth]{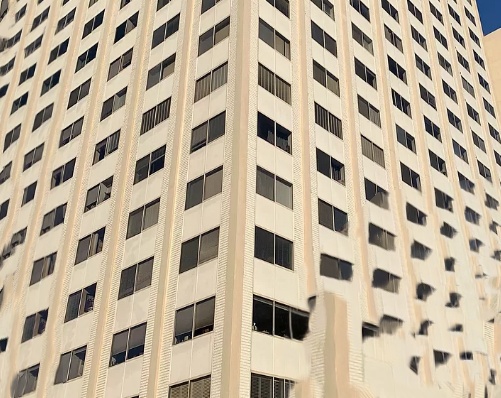}}
\caption{(b) Huang \etal}
\end{subfigure}
\\
\begin{subfigure}{.49\linewidth}
  \centerline{\includegraphics[width=.94\textwidth]{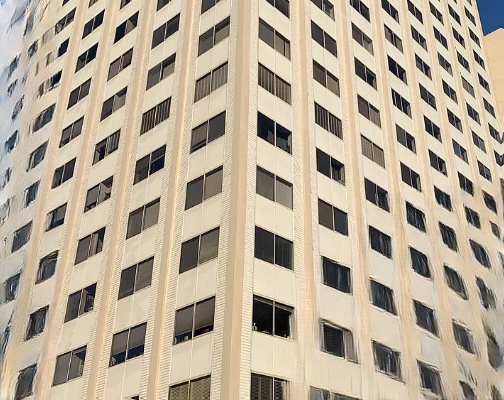}}
\caption{(c) Lama}
\end{subfigure}
\begin{subfigure}{.49\linewidth}
  \centerline{\includegraphics[width=.94\textwidth]{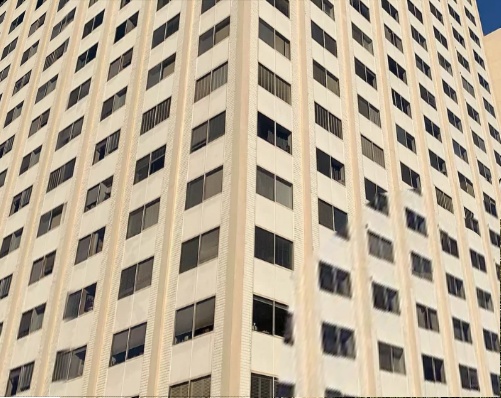}}
\caption{(d) BPI}
\end{subfigure}
\\
\begin{subfigure}{.49\linewidth}
  \centerline{\includegraphics[width=.94\textwidth]{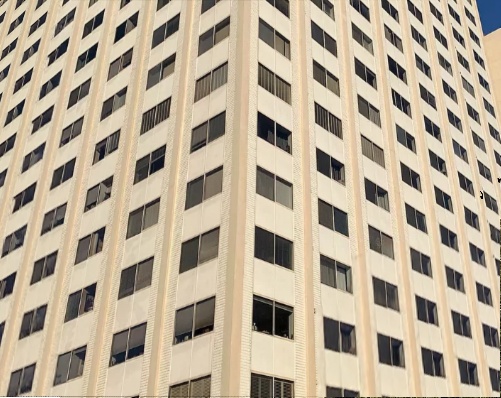}}
\caption{(e) NPP-Net}
\end{subfigure}
\begin{subfigure}{.49\linewidth}
  \centerline{\includegraphics[width=.94\textwidth]{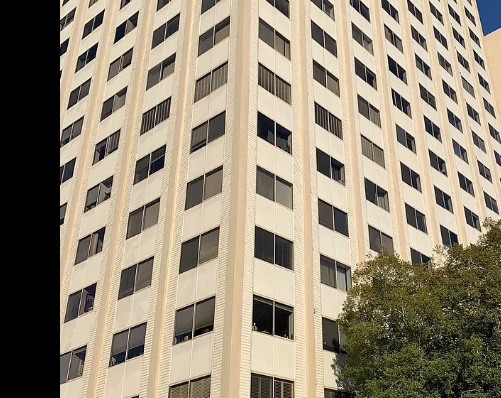}}
\caption{(f) GT}
\end{subfigure}
\caption{
Qualitative results of a multi-planar NPP scene.
}
\label{multi_plane2}
\end{figure}

\begin{figure}[!h]
    \captionsetup[subfigure]{labelformat=empty, font=normalsize}
\begin{subfigure}{.49\linewidth}
  \centerline{\includegraphics[width=.94\textwidth]{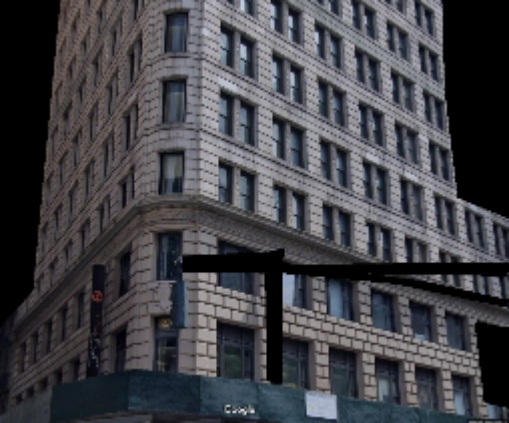}}
\caption{(a) Input}
\end{subfigure}
\begin{subfigure}{.49\linewidth}
  \centerline{\includegraphics[width=.94\textwidth]{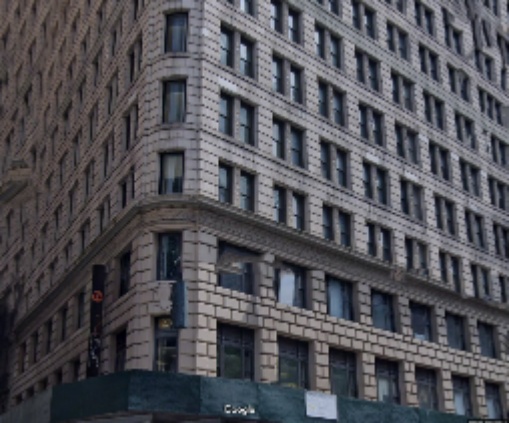}}
\caption{(b) Huang \etal}
\end{subfigure}
\\
\begin{subfigure}{.49\linewidth}
  \centerline{\includegraphics[width=.94\textwidth]{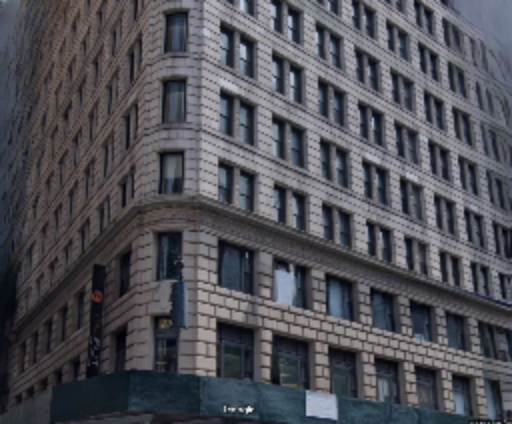}}
\caption{(c) Lama}
\end{subfigure}
\begin{subfigure}{.49\linewidth}
  \centerline{\includegraphics[width=.94\textwidth]{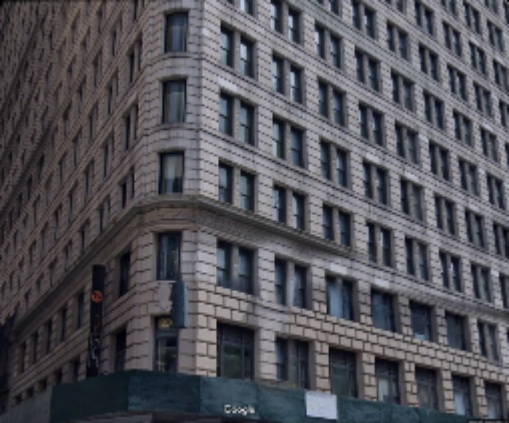}}
\caption{(d) BPI}
\end{subfigure}
\\
\begin{subfigure}{.49\linewidth}
  \centerline{\includegraphics[width=.94\textwidth]{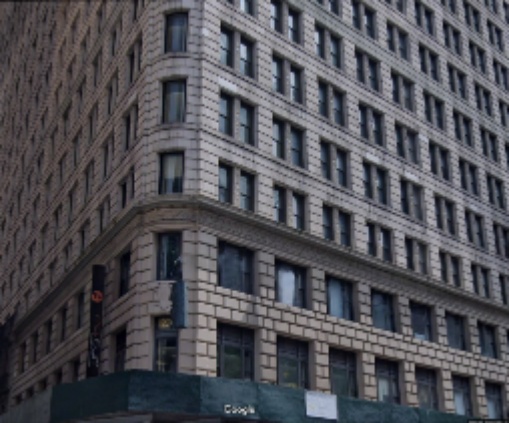}}
\caption{(e) NPP-Net}
\end{subfigure}
\begin{subfigure}{.49\linewidth}
  \centerline{\includegraphics[width=.94\textwidth]{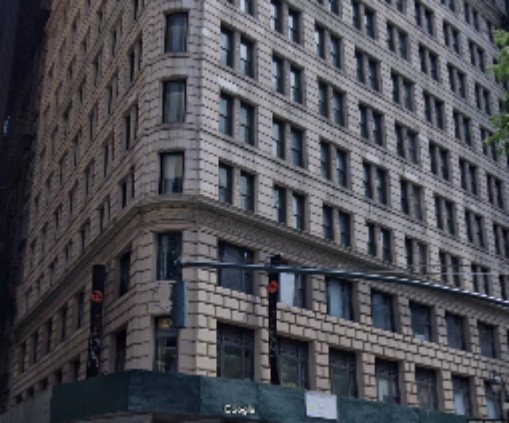}}
\caption{(f) GT}
\end{subfigure}
\caption{
Qualitative results of a multi-planar NPP scene. All images are resized to half of the original size to reduce file size.
}
\label{multi_plane3}
\end{figure}

\begin{figure}[!h]
    \captionsetup[subfigure]{labelformat=empty, font=normalsize}
\begin{subfigure}{.49\linewidth}
  \centerline{\includegraphics[width=.75\textwidth]{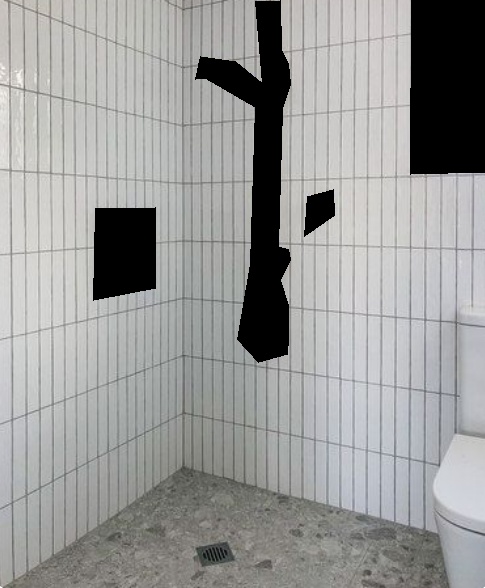}}
\caption{(a) Input}
\end{subfigure}
\begin{subfigure}{.49\linewidth}
  \centerline{\includegraphics[width=.75\textwidth]{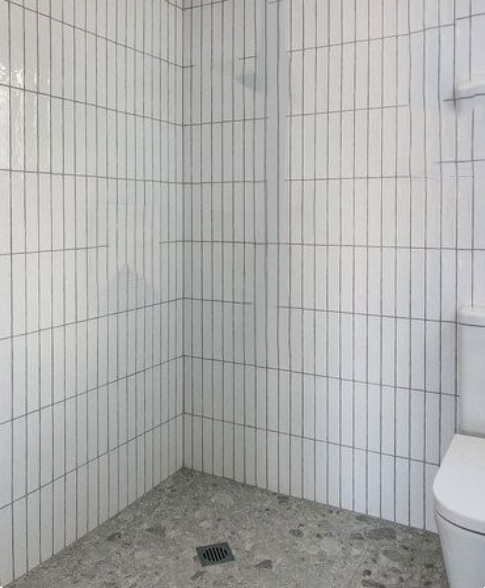}}
\caption{(b) Huang \etal}
\end{subfigure}
\\
\begin{subfigure}{.49\linewidth}
  \centerline{\includegraphics[width=.75\textwidth]{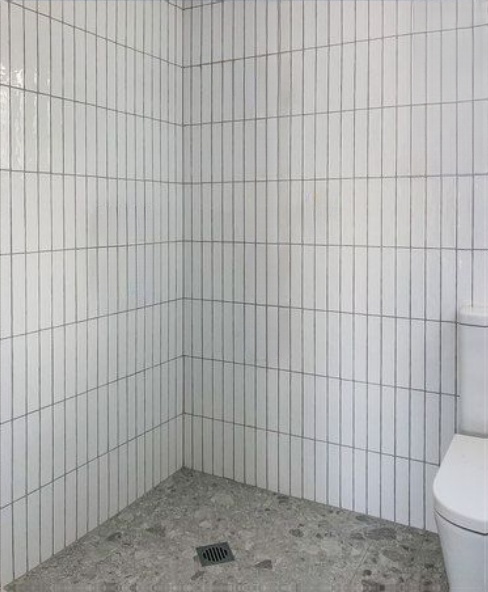}}
\caption{(c) Lama}
\end{subfigure}
\begin{subfigure}{.49\linewidth}
  \centerline{\includegraphics[width=.75\textwidth]{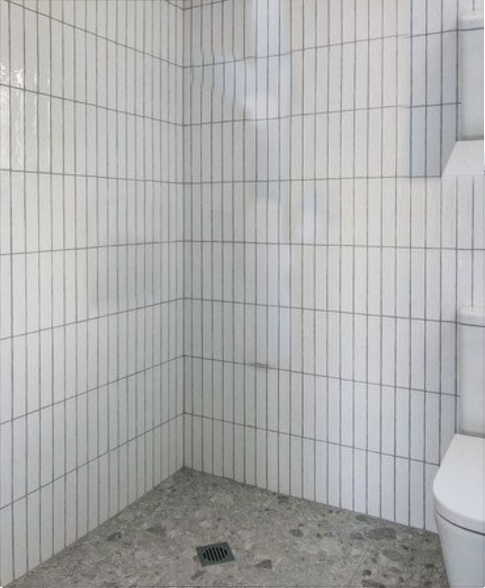}}
\caption{(d) BPI}
\end{subfigure}
\\
\begin{subfigure}{.49\linewidth}
  \centerline{\includegraphics[width=.75\textwidth]{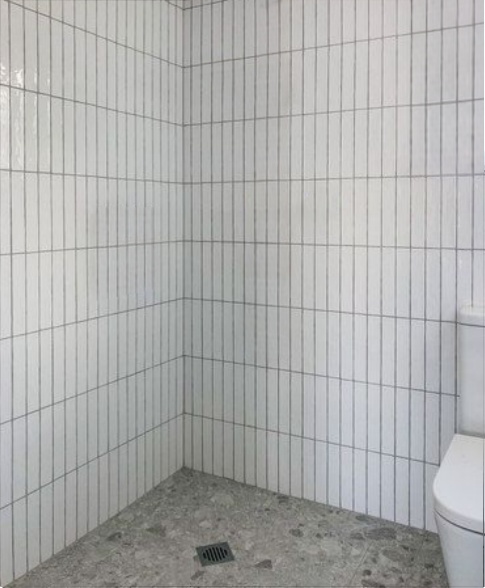}}
\caption{(e) NPP-Net}
\end{subfigure}
\begin{subfigure}{.49\linewidth}
  \centerline{\includegraphics[width=.75\textwidth]{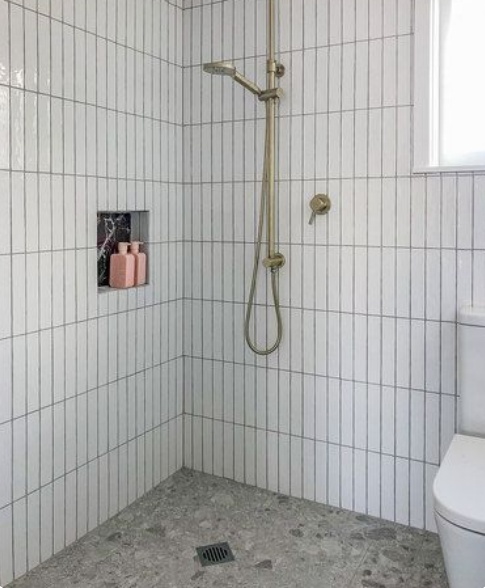}}
\caption{(f) GT}
\end{subfigure}
\caption{
Qualitative results of a multi-planar NPP scene. 
}
\label{multi_plane4}
\end{figure}

The qualitative comparisons with baselines are shown in Figure \ref{multi_plane1} to Figure \ref{multi_plane4}.


\clearpage

\section{Failure Case and Future Work}

\begin{figure}
    \centering
    \includegraphics[scale=0.5]{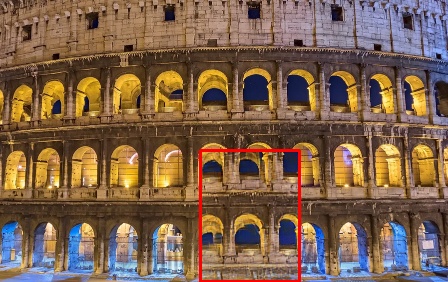}
    \caption{Failure case of NPP-Net for non-planar scene, where region inside red box is inpainted.}
    \label{failure}
\end{figure}

Our method fails in the case of non-planar NPP scene. Figure \ref{failure} shows a failure case for non-planar scene.

For future work, We plan to explore:
(1) NPP scenes with more complicated geometry (\eg non-planar scenes).
(2) a few-shot learning strategy for NPP-Net that can incorporate prior knowledge from only a few NPP images.

\section{Use of Existing Assets}

\textbf{CodeBase}: 
We implement NPP-Net based on the existing assets~\cite{yenchenlin,richzhang,S-aiueo32,jonbarron,fled,borda}. For metrics, we adopt the following implementations: LPIPS~\cite{zhang2018perceptual}, FID~\cite{parmar2021cleanfid}, RMSE~\cite{sklearn_api}, SSIM~\cite{van2014scikit}, and PSNR~\cite{opencv_library}.
Also, we adopt the following implementations for baselines: Image Qualiting~\cite{axu2}, PatchMatch~\cite{vacancy}, DIP~\cite{DmitryUlyanov}, Siren~\cite{vsitzmann}, ProFill~\cite{zeng}, Huang \etal~\cite{jbhuang0604,SunskyF}, and BPI~\cite{vacancy}.

\clearpage
%
%
\bibliographystyle{splncs04}
\bibliography{egbib}